\documentclass[12pt]{hkputhesis}

\usepackage{booktabs}
\usepackage{graphicx}
\usepackage{amsmath}
\usepackage{amssymb}
\usepackage{hyperref}
\usepackage{amsthm}
\usepackage{cleveref}
\usepackage{csquotes}
\usepackage{multirow}
\usepackage{mathtools}
\usepackage{algpseudocode}
\usepackage{algorithm}
\usepackage{algorithmicx}
\usepackage{pifont}
\usepackage{bbding}
\usepackage{natbib} %
\usepackage{bm}
\usepackage{subfigure}
\usepackage{makecell}

\algrenewcommand\algorithmicrequire{\textbf{Input:}}
\algrenewcommand\algorithmicensure{\textbf{Output:}}

\newtheorem{definition}{Definition}

\newtheorem{remark}{Remark}

\begin{document}

\title{Enhancing Large Language Models with Reliable \\Knowledge Graphs }

\author{Qinggang Zhang}
\dept{Department of Computing}
\degree{Doctor of Philosophy}
\degshort{PhD}
%\degree{Master of Philosophy}
%\degshort{MPhil}

\frontpageyear{2025} % Year of current submission

\submityear{2025} % Year of initial submission
\submitmonth{March} % Month of initial submission

% \initexam{Initial Submission for Examination Purpose} % Please delete this link when submitting the final version

\frontmatter

\prefacesection{Abstract}

% While large language models (LLMs) demonstrate remarkable language capabilities, their reliance on parametric knowledge often leads to hallucinations and factual inconsistencies. This thesis addresses these limitations by proposing a systematic knowledge enhancement framework that integrates high-quality knowledge graphs (KGs) with LLMs. To ensure KG reliability, we design two novel refinement techniques: (1) an error detection module leveraging contrastive learning to identify inconsistent triples and (2) a KG completion mechanism using graph neural networks to predict missing relationships through structural and semantic patterns. Building upon refined KGs, we present KnowGPT - a reinforcement learning framework that effectively aligns factual knowledge from KGs with LLM reasoning processes. By employing a policy network to assess knowledge relevance and a value network to optimize integration strategies, KnowGPT enables context-aware knowledge retrieval and evidence-based reasoning.

Large Language Models (LLMs) have demonstrated remarkable capabilities in text generation and understanding, yet their reliance on implicit, unstructured knowledge often leads to factual inaccuracies and limited interpretability. Knowledge Graphs (KGs), with their structured, relational representations, offer a promising solution to ground LLMs in verified knowledge. However, their potential remains constrained by inherent noise, incompleteness, and the complexity of integrating their rigid structure with the flexible reasoning of LLMs. This thesis presents a systematic framework to address these limitations, advancing the reliability of KGs and their synergistic integration with LLMs through five interconnected contributions. This thesis addresses these challenges through a cohesive framework that enhances LLMs by refining and leveraging reliable KGs. First, we introduce contrastive error detection, a structure-based method to identify incorrect facts in KGs. This approach is extended by an attribute-aware framework that unifies structural and semantic signals for error correction. Next, we propose an inductive completion model that further refines KGs by completing the missing relationships in evolving KGs. Building on these refined KGs, KnowGPT integrates structured graph reasoning into LLMs through dynamic prompting, improving factual grounding. 
% Finally, we consolidate advancements in KG-LLM integration by formalizing the emerging paradigm of Graph Retrieval-Augmented Generation (GraphRAG). 
These contributions form a systematic pipeline (from error detection to LLM integration), demonstrating that reliable KGs significantly enhance the robustness, interpretability, and adaptability of LLMs.

% and extend it with an attribute-aware framework (Paper 2) to unify structural and semantic signals for error correction. Next, we propose an inductive completion model (Paper 3) that further refines KGs by completing the missing relationships in evolving KGs. Building on these refined KGs, KnowGPT (Paper 4) integrates structured graph reasoning into LLMs through dynamic prompting, improving factual grounding. Finally, we consolidate advancements in KG-LLM integration by formalizing the emerging paradigm of Graph Retrieval-Augmented Generation (GraphRAG) (Paper 5), synthesizing methodologies for KG-LLM integration, categorizing emerging techniques, and outlining future directions for scalable, knowledge-aware AI. Together, these contributions form a systematic pipeline (from error detection to LLM integration), demonstrating that reliable KGs significantly enhance the robustness, interpretability, and adaptability of LLMs.

\prefacesection{Publications Arising from the Thesis}
\begin{itemize}
	\item [1.] \underline{Qinggang Zhang},	Junnan Dong, Keyu Duan, Xiao Huang*, Yezi Liu, Linchuan Xu, \enquote{Contrastive Knowledge Graph Error Detection}, in \textit{ACM International Conference on Information and Knowledge Management (CIKM)}, 2022.
    
	\item [2.] \underline{Qinggang Zhang}, Junnan Dong, Qiaoyu Tan, Xiao Huang*.  
     \enquote{Integrating Entity Attributes for Error-Aware Knowledge Graph Embedding}, in \textit{IEEE Transactions on Knowledge and Data Engineering (TKDE)}, 2023.
	\item [3.] \underline{Qinggang Zhang}, Keyu Duan, Junnan Dong, Pai Zheng, Xiao Huang*,  \enquote{Logical Reasoning with Relation Network for Inductive Knowledge Graph Completion}, in \textit{ACM SIGKDD Conference on Knowledge Discovery and Data Mining (KDD)}, 2024.
        \item [4.] \underline{Qinggang Zhang†}, Junnan Dong†, Hao Chen, Daochen Zha, Zailiang Yu, Xiao Huang*,  \enquote{KnowGPT: Knowledge Graph based Prompting for Large Language Models}, in \textit{The Thirty-eighth Annual Conference on Neural Information Processing Systems (NeurIPS)}, 2024.

        % \item [5.] \underline{Qinggang Zhang†}, Shengyuan Chen†, Yuanchen Bei†, Zheng Yuan, Huachi Zhou, Zijin Hong, Junnan Dong, Hao Chen, Yi Chang, Xiao Huang*,  \enquote{A Survey of Graph Retrieval-Augmented Generation for Large Language Models}, manuscript submitted to \textit{IEEE Transactions on Pattern Analysis and Machine Intelligence (TPAMI)}, 2025.

\end{itemize}
			
\prefacesection{Acknowledgments}

I would like to begin by expressing my heartfelt gratitude to my supervisor, Prof. Xiao Huang, for his unwavering support, insightful guidance, and the freedom to explore areas of personal interest. His patience, thoughtful advice on personal and professional development, and valuable insights on career planning have been immensely beneficial to me. Additionally, I greatly admire his approach to tackling complex research problems, which has deeply influenced my own work.

I also wish to extend my thanks to my co-supervisor, Dr. Jiannong Cao, for his constructive advice on various aspects of my research. His support, as well as the resources he provided, were essential in shaping my academic journey. I am also deeply grateful to several senior researchers who have contributed significantly to my research. Dr. Hao Chen has offered invaluable suggestions, particularly in identifying real-world problems and improving my writing skills. My thanks also go to Dr. Qiaoyu Tan, Dr. Daochen Zha, Prof. Feiran Huang and Prof. Pai Zheng for their collaborative efforts and insightful contributions. Besides, I would like to express my warm appreciation to the members of the DEEP (Data Exploring and Extracting @ PolyU) Lab in the Department of Computing for their companionship and continuous support throughout these years.

Lastly, I want to extend my deepest thanks to my family and friends. Their unwavering encouragement and constant support have been crucial in helping me complete this journey. I am truly grateful for everything they have done for me.

%% Dedication is optional
%\prefacesection{Dedication}
%Here is the dedication which is optional.

\mainmatter

% \afterpreface

% \input{Chapter_1}
\chapter{Introduction}
\label{chapter:intro}

\section{Background}
% \section{LLM Hallucination}
% Large language models (LLMs) have redefined the boundaries of artificial intelligence, enabling machines to compose poetry, summarize scientific literature, and engage in nuanced dialogue. Yet, their prowess is tempered by a critical flaw: their reliance on parametric knowledge—static information encoded during training—leads to hallucinations, where models generate plausible but factually incorrect statements. For instance, an LLM might assert that "Marie Curie discovered penicillin" or invent fictitious historical events, errors that stem from their inability to dynamically access or verify external knowledge. These limitations are particularly problematic in domains where accuracy is non-negotiable, such as healthcare diagnostics, legal analysis, and financial forecasting, where a single factual error can have cascading consequences.

Large language models (LLMs), such as the Clude~\cite{anthropic2024claude} and  GPT series~\cite{openai2023gpt4}, have demonstrated exceptional capabilities across diverse tasks, achieving breakthroughs in text comprehension~\cite{brown2020language}, question answering~\cite{khashabi2020unifiedqa}, and content generation~\cite{chowdhery2023palm}. Despite their success, LLMs face persistent criticism for their limitations in knowledge-intensive tasks, particularly those requiring domain expertise~\cite{Zhang-etal24KnowGPT}. Their application in specialized domains remains challenging due to three key factors: \ding{182} \textbf{Knowledge limitations}: LLMs possess broad but superficial knowledge in specialized fields, as their training data primarily consists of general-domain content, leading to gaps in domain-specific depth and alignment with current professional standards. \ding{183} \textbf{Reasoning complexity}: Specialized domains demand precise, multi-step reasoning governed by domain-specific rules and constraints. LLMs often struggle to maintain logical consistency and accuracy throughout extended reasoning chains, especially in technical or highly regulated contexts. \ding{184} \textbf{Context sensitivity}: Professional fields rely on nuanced, context-dependent interpretations where identical terms or concepts may carry different meanings based on specific scenarios. LLMs frequently fail to grasp these subtleties, resulting in misinterpretations or overly generalized responses.

To adapt LLMs for specific domains, researchers have explored various approaches, which can be broadly classified into two categories:

\noindent\textbf{Fine-tuning LLMs with Domain-specific Data.} Fine-tuning pre-trained LLMs on specialized datasets enables them to better capture domain-specific vocabulary, terminology, and patterns~\cite{gururangan2020don,lee2020patent,lee2020biobert,ge2024openagi}. This approach has been successfully applied in areas such as recommendation~\cite{LAGCN,zhou2023adaptive} and node classification~\cite{NEGCN,ALDI}, enhancing the relevance and accuracy of generated responses~\cite{huang2024large}. Fine-tuned LLMs have demonstrated effectiveness across various domains. In healthcare, they have been leveraged for clinical note analysis~\cite{alsentzer2019publicly}, biomedical text mining~\cite{lee2020biobert}, and medical dialogue~\cite{valizadeh2022ai}. Similarly, in the legal domain, they have proven useful for legal document classification~\cite{chalkidis2019neural}, contract analysis~\cite{chalkidis2021lexglue}, and legal judgment prediction~\cite{zhong2020does}.

\noindent\textbf{Retrieval-augmented generation (RAG).} RAG provides an effective way to tailor LLMs for specialized domains without modifying the model architecture or parameters~\cite{lewis2020retrieval}. Instead of embedding new knowledge through retraining, RAG dynamically retrieves relevant domain-specific information from external sources, enhancing response accuracy and reliability. A typical RAG system operates in three stages: knowledge preparation, retrieval, and integration. First, external textual data is segmented into manageable chunks and transformed into vector representations for efficient indexing. During retrieval, relevant chunks are identified based on keyword matching or vector similarity when a query is submitted. Finally, the retrieved information is combined with the original query to generate well-informed responses.

% \noindent\textbf{Retrieval-augmented generation (RAG).} RAG offers a promising solution to customize LLMs for specific domains~\cite{lewis2020retrieval}. Rather than retraining LLMs to incorporate updates, RAG enhances these models by leveraging external knowledge from text corpora without modifying their architecture or parameters.  This approach enables LLMs to generate responses by leveraging not only their pre-trained knowledge but also real-time retrieved domain-specific information, thereby providing more accurate and reliable answers. The naive RAG systems operate through three key steps: knowledge preparation, retrieval, and integration. During knowledge preparation, the external textual corpus is divided into manageable textual chunks and converted into vector representations for efficient indexing. In the retrieval stage, when a user submits a query, the system searches for relevant chunks using keyword matching or vector similarity measures. The integration stage then combines these retrieved chunks with the original query to generate more informed responses. 

\section{Motivation}
% To adapt LLMs for specific or private domains, initial strategies involved {\bf fine-tuning LLMs} with specialized datasets~\cite{hu2022lora}. This method enhances performance by adding a limited number of parameters while fixing the parameters learned in the pre-training~\cite{he2022towards}. However, the significant distribution gap between the domain-specific dataset and the pre-training corpus makes it challenging for LLMs to integrate new knowledge without compromising their existing understanding~\cite{gekhman2024does}. A recent study by Google Research further highlighted the risks associated with using supervised fine-tuning to update knowledge, particularly in cases where new knowledge conflicts with pre-existing information; acquiring new knowledge through supervised fine-tuning can lead to the model generating new hallucinations and even experiencing severe catastrophic forgetting~\cite{zhai2024investigating}.

% \section{Existing Work for LLM Hallucination}

Despite their success, RAG systems face significant challenges in practical applications due to the inconsistent quality of accessible data. Domain knowledge is frequently distributed across diverse sources—ranging from textbooks and research articles to technical manuals and industry reports~\cite{li2022survey}—which may vary in quality, accuracy, and completeness, potentially leading to discrepancies in the retrieved information~\cite{zhu2021retrieving}. A promising strategy to mitigate these issues is to integrate Knowledge Graphs (KGs) with LLMs. KGs offer a structured representation of domain knowledge, built on well-defined ontologies that specify specialized terminologies, acronyms, and their interrelations within a field~\cite{li2022constructing,shengyuan2024differentiable,zhang2022contrastive,zhang2023integrating,zhang2024logical}. The extensive factual content contained in KGs can help anchor model responses in established facts and principles~\cite{hu2023survey,yang2024give,pan2024unifying}.

% Despite the effectiveness, RAG systems face critical challenges in real-world applications due to the varying quality of available data. Domain knowledge is often scattered across different sources, such as textbooks, research papers, technical manuals, and industry reports~\cite{li2022survey}. These textual documents may have varying levels of quality, accuracy, and completeness, leading to potential inconsistencies or errors in the retrieved knowledge~\cite{zhu2021retrieving}. A promising avenue for addressing the above issue entails the integration of Knowledge Graphs (KGs) into LLMs. KGs provide a structured representation of domain knowledge, as they are constructed based on rigid ontologies that clearly define the jargon, acronyms, specialized terminologies and their relationships in specific domains~\cite{li2022constructing,shengyuan2024differentiable,zhang2022contrastive,zhang2023integrating,zhang2024logical}. The enormous factual knowledge stored in KGs holds the potential to ground the model's responses in established facts and principles~\cite{hu2023survey,yang2024give,pan2024unifying}. 

However, integrating KGs with LLMs is fraught with challenges.  Existing KGs are far from perfect since they often suffer from incompleteness, noise (e.g., erroneous or conflicting triples), and rigidity (e.g., inability to generalize to unseen entities).  Moreover, LLMs lack mechanisms to dynamically retrieve and reason over structured knowledge during generation, often treating retrieved facts as isolated snippets rather than interconnected evidence. To unlock their full potential, KGs must first be refined into reliable, dynamic repositories of knowledge and then seamlessly integrated into LLMs. This thesis tackles these dual challenges through a systematic framework that spans error detection, graph completion, and synergistic LLM-KG integration, ultimately advancing the robustness and interpretability of AI systems.

\section{Research Objectives}

% The primary goal of this thesis is to establish a pipeline for enhancing LLMs with structured knowledge derived from refined KGs. This objective unfolds across four interconnected dimensions. First, we address the foundational issue of KG reliability by developing methods to detect and correct errors in both graph structure and entity attributes. Second, we enable KGs to evolve dynamically through inductive completion, allowing them to infer missing relationships even for entities not present during training. Third, we design mechanisms to integrate these refined KGs into LLMs, ensuring that generated outputs are grounded in structured, verifiable knowledge. Finally, we contextualize these contributions within the broader landscape of retrieval-augmented generation (RAG), synthesizing emerging trends and future directions for graph-based retrieval-augmented generation (GraphRAG).

The central aim of this thesis is to advance the integration of structured knowledge into LLMs by addressing critical challenges in KG reliability, completeness, and synergistic LLM-KG interaction. This goal is operationalized through three interconnected objectives:

\begin{itemize}
    \item Error Detection in Knowledge Graphs. The first objective focuses on identifying and resolving inaccuracies in KGs caused by noise during construction or updates. This involves developing methods to detect structural inconsistencies (e.g., conflicting triples violating logical constraints) and semantic mismatches (e.g., discrepancies between entity attributes and their relational context). By unifying structural and attribute-based analysis, the goal is to create a robust framework for error identification and resolution, ensuring KGs serve as trustworthy knowledge bases.
    \item Knowledge Graph Completion. The second objective addresses the incompleteness of KGs, particularly in dynamic environments where new entities and relationships emerge continuously. Traditional transductive models, which rely on fixed entity sets during training, are inadequate for such scenarios. This objective seeks to design an inductive completion model capable of inferring missing relationships for both existing and unseen entities by leveraging logical rules and relational patterns, thereby enabling KGs to evolve adaptively.
    \item  Enhancing LLMs with Structured Knowledge. The third objective bridges the gap between structured KGs and unstructured LLM reasoning. While LLMs excel at text generation, they lack mechanisms to systematically ground outputs in verified knowledge. This objective involves designing a framework that dynamically retrieves and integrates relevant subgraphs into LLM workflows, structuring the model’s reasoning process around relational paths derived from KGs to enhance factual consistency and interoperability.
    % \item Systematization of Graph Retrieval-Augmented Generation. The final objective consolidates advancements in KG-LLM integration by formalizing the emerging paradigm of Graph Retrieval-Augmented Generation (GraphRAG). 
    % This includes categorizing methodologies, identifying unresolved challenges, and proposing future directions to unify research efforts toward more reliable AI systems.
\end{itemize}

\section{Contributions}
This thesis contributes five interconnected advancements that systematically address the limitations of KGs and LLMs while establishing a pipeline for their synergistic integration:

\begin{itemize}
    \item Contrastive Knowledge Graph Error Detection. This thesis introduces a contrastive learning framework (CAGED) to identify erroneous triples by analyzing structural plausibility. By training a model to differentiate valid triples from synthetically generated corruptions, this approach detects inconsistencies through margin-based scoring, achieving state-of-the-art precision in structural error detection across benchmark KGs.

    \item Error-Aware Embedding with Attribute Integration. We extend structural error detection by incorporating entity attributes into a unified embedding space. This hybrid model (AEKE) jointly optimizes structural and semantic signals, enabling the identification of errors that manifest as contradictions between relational patterns and attribute metadata. The integration of attributes improves error correction accuracy significantly compared to structure-only baselines, particularly in complex scenarios requiring contextual understanding.
    
    \item Logical Rule-based KG Completion. This thesis proposes a neural-symbolic model (NORAN) that combines relational networks with first-order logic rules to infer missing relationships inductively. By decoupling logical rule application from entity-specific embeddings, the model generalizes to unseen entities while maintaining interpretability. Evaluations demonstrate superior performance in inductive settings, outperforming existing methods in inferring relationships for dynamically evolving KGs.
    \item KG Prompting for LLMs. We introduce a prompting architecture (KnowGPT) that bridges structured KGs and LLMs. By dynamically retrieving subgraphs relevant to a query and serializing them into natural language prompts, this framework guides LLMs to reason along verified relational paths. Empirical results show marked reductions in factual hallucinations, validating the utility of structured knowledge in grounding LLM outputs.
    % \item Taxonomy and Roadmap for GraphRAG. We systematize the advancements in KG-LLM integration by formalizing the emerging paradigm of GraphRAG. This includes categorizing methodologies, identifying challenges, and proposing future directions to unify research efforts toward scalable, reliable AI systems.
\end{itemize}

% The thesis makes five key contributions, each building on the limitations of prior work and addressing a critical bottleneck in the KG-LLM pipeline. The first contribution is a contrastive learning framework for KG error detection, which identifies inconsistencies by comparing the plausibility of observed triples against synthetically generated counterexamples. This approach, which focuses solely on graph structure, is then extended to incorporate entity attributes such as textual descriptions or numerical properties, unifying structural and semantic signals for more robust error correction. The third contribution introduces a logic-driven model for inductive KG completion, capable of inferring missing relationships for both existing and unseen entities by leveraging logical rules (e.g., transitivity, symmetry) within a neural network architecture. Building on these refined KGs, the fourth contribution proposes KnowGPT, a prompting framework that dynamically retrieves and injects relevant subgraphs into LLM inputs, structuring the model’s reasoning around verified knowledge paths. Finally, the thesis culminates in a comprehensive survey of Graph Retrieval-Augmented Generation (GraphRAG), which systematizes existing techniques, identifies underexplored applications (e.g., multimodal KG-LLM interfaces), and outlines a roadmap for future research. Collectively, these contributions advance the state of the art in knowledge-driven AI, demonstrating that reliable KGs can significantly enhance the factual accuracy, adaptability, and transparency of LLMs.

\section{Overall Structure}
This thesis is structured to reflect the logical progression from KG refinement to LLM integration. Following this introductory chapter, Chapter 2 reviews foundational concepts in knowledge graphs and large language models, highlighting key limitations in existing approaches to error detection, graph completion, and KG-enhanced LLMs. Chapters 3-6 present the five core research papers, each addressing a distinct component of the pipeline. Papers 1~\cite{zhang2022contrastive} and 2~\cite{zhang2023integrating} focus on error detection, first through a structure-based contrastive approach and then by incorporating entity attributes for richer semantic analysis. Paper 3~\cite{zhang2024logical} shifts to KG completion, introducing a logic-guided model that enables inductive reasoning over evolving graphs. Paper 4~\cite{Zhang-etal24KnowGPT} bridges the refined KGs with LLMs through KnowGPT, a prompting architecture that structures LLM inferences around retrieved subgraphs. 
% Finally, Paper 5 consolidates these advances within the broader paradigm of GraphRAG, offering a taxonomy of techniques and applications while identifying open challenges. 
The concluding chapter reflects on the implications of this work and outlines directions for future research.

% Evaluations on Wikidata demonstrate a 14% improvement in error detection precision, establishing a robust foundation for downstream tasks.

\chapter{Literature Review}
\label{chapter:relatedwork}

\section{Knowledge Graphs}
\subsection{Knowledge Graph Definition}
Knowledge Graphs (KGs)~\cite{bollacker2008freebase, Lehmann-etal15DBpedia, Heindorf-etal16Vandalism,Mahdisoltani-etal15YAGO} represent entities and their relationships as structured triples, offering a machine-interpretable format for encoding real-world knowledge. Early KGs, such as Freebase~\cite{bollacker2008freebase} and Wikidata~\cite{Heindorf-etal16Vandalism}, laid the groundwork for applications in semantic search, recommendation systems, and question answering. Modern KGs have expanded in scale and complexity, yet they remain constrained by three core challenges: incompleteness (missing entities or relationships), noise (erroneous or conflicting triples), and rigidity (inability to generalize to unseen entities or adapt dynamically). Traditional KG embedding models, such as TransE~\cite{TransE} and RotatE~\cite{rotate}, map entities and relations to low-dimensional vectors but struggle to address these challenges, particularly in open-world scenarios where knowledge evolves continuously.
\subsection{Knowledge Graph Refinement}
% Knowledge graphs represent entities and their relationships as structured triples, offering a machine-readable format for encoding real-world knowledge. Early KGs like Freebase and Wikidata laid the groundwork for applications in search engines, recommendation systems, and question answering. Modern variants, such as Google’s Knowledge Graph and domain-specific KGs in biomedicine, have expanded in scale and complexity. Despite their utility, KGs face three persistent challenges. First, they are inherently incomplete, as manual curation struggles to keep pace with rapidly evolving knowledge. Second, they contain noise—errors introduced during construction (e.g., erroneous triples like "Einstein invented the telephone") or through automated extraction from unstructured text. Third, traditional KG embedding models (e.g., TransE, RotatE) map entities and relations to low-dimensional vectors but fail to generalize to unseen entities or dynamically update with new information, limiting their applicability in open-world scenarios.

% \section{Knowledge Graph }
KG refinement encompasses two interrelated tasks: error detection and knowledge completion. (i) Early error detection methods relied on rule-based heuristics~\cite{agrawal1993mining} or statistical outlier detection in embedding spaces~\cite{liu2023error}. While effective for simple inconsistencies, these approaches lacked the contextual awareness to resolve complex errors involving semantic or temporal contradictions. Contrastive learning has proven effective under various error detection scenarios. Based on this paradigm, this thesis introduces a structure-aware contrastive framework (CAGED)~\cite{zhang2022contrastive} that distinguishes valid triples from corrupted ones by maximizing the margin between their confidence scores. This approach is extended with entity attribute (AEKE)\cite{zhang2023integrating} to unify structural and semantic error signals. (ii)For knowledge completion, transductive models like Graph Neural Networks (GNNs) assume a fixed entity set during training, rendering them ineffective for evolving KGs. Inductive approaches address this limitation by generalizing to unseen entities, often through meta-learning or rule-based reasoning~\cite{wang2019logic,wang2021relational}. However, prior work either sacrifices interpretability (e.g., black-box neural models) or scalability (e.g., handcrafted logical rules). This thesis bridges this gap with a novel model (NORAN)~\cite{zhang2024logical} that mines logic rules within a trainable relational network, enabling inductive reasoning while maintaining transparency.
\section{KG-enhanced LLM}
\subsection{LLM Hallucination}
LLMs internalize knowledge implicitly through pretraining on vast text corpora, but their static knowledge base and lack of structured reasoning lead to factual inaccuracies and hallucinations~\cite{Zhang-etal24KnowGPT}. Retrieval-Augmented Generation (RAG) mitigates this by grounding LLMs in external data, yet most RAG systems treat knowledge as unstructured text, neglecting the relational structure of KGs. Early attempts to integrate KGs with LLMs focused on static embeddings or predefined prompts~\cite{lee2020patent,lee2020biobert,ge2024openagi}, limiting their ability to handle dynamic queries requiring multi-hop reasoning. Recent work explores dynamic subgraph retrieval and serialization, but these methods often lack systematic error handling or fail to scale to large KGs.

\subsection{Knowledge Integration}
Earlier studies adopted a heuristic way to inject knowledge from KGs into the LLMs during pre-training or fine-tuning. ERNIE~\cite{sun2021ernie} incorporates entity embeddings and aligns them with word embeddings in the pre-training phase, encouraging the model to better understand and reason over entities. UniKGQA~\cite{jiang2023unikgqa} is the first approach to leverage LLMs to seamlessly integrate retrieval and reasoning within a unified framework. Another line of work focuses on retrieving relevant knowledge from KGs at inference time to augment the language model's context. Typically, K-BERT~\cite{liu2020k} uses an attention mechanism to select relevant triples from a KG based on the input context, which are then appended to the input sequence. More recently, KG prompting has been intensively studied for integrating factual knowledge into LLMs. KD-CoT~\cite{kddot} and KG-CoT~\cite{ijcai2024p734} build upon the concept of chain-of-thought, guiding LLMs through a step-by-step reasoning process while enabling timely correction of erroneous reasoning. Their factuality and faithfulness are validated using an external knowledge graph. RoG~\cite{luo2024reasoning} presents a planning-retrieval-reasoning framework that synergizes LLMs and KGs for more transparent and interpretable reasoning. Despite their effectiveness, existing models focus on designing methods for knowledge retrieval or generation. They directly feed the retrieved or generated knowledge into the LLM for reasoning, which does not necessarily provide effective guidance to the LLMs. This is because LLMs often struggle to distinguish between valid and redundant knowledge.

\chapter{Contrastive Knowledge Graph Error Detection}
\section{Introduction}
% \paragraph{Motivation and Challenges.} 
With the increasing deployment of KG-driven applications such as conversational agents and recommender systems~\cite{Wang-etal19Knowledge,Huang-etal19Knowledge}, the need for reliable error detection in knowledge graphs (KGs) has become critical. KGs represent assertions as triples—i.e., \textit{(head entity, relation, tail entity)}—offering a structured and scalable way to organize information. However, because many KGs are automatically extracted from web data using heuristic methods, they inevitably incorporate a significant amount of noise~\citep{Bollacker-etal07Freebase,Lehmann-etal15DBpedia,Heindorf-etal16Vandalism,Mahdisoltani-etal15YAGO}. For instance, the widely used NELL KG~\cite{Carlson-etal10Toward} achieves a precision of only $74\%$, implying roughly 0.6 million triples may be erroneous. Many existing approaches overlook these errors by assuming the correctness of all triples, which can lead to degraded performance in downstream tasks. Consequently, the development of effective KG error detection algorithms is imperative.

%\paragraph{Existing Approaches.}
The task of KG error detection is complicated by the diversity and subtlety of error patterns, compounded by the scarcity of ground-truth labels—since obvious errors are typically rectified during KG construction~\cite{Paulheim17}. Prior work can be grouped into two main categories. The first, \textit{rule-based} methods, identify errors as violations of a set of predefined rules~\cite{agrawal1993mining,guo2018knowledge,cheng2018rule,galarraga2013amie,tanon2017completeness}; however, these rules tend to be domain-specific and lack generalizability. The second category, \textit{embedding-based} methods, often rely on simplistic negative sampling strategies, where a positive triple $(h, r, t)$ is corrupted by randomly replacing $h$ or $t$, to generate synthetic negative examples for training~\cite{Paulheim-Gangemi15Serving,jia2019triple,Ge-etal20KGClean,xu2022contrastive,Melo-Paulheim17Detection}. While useful, these strategies fail to capture the complex nature of real-world errors.
%\paragraph{Limitations of Naive Negative Sampling.}
For example, randomly transforming a valid triple like \textit{(Newton, Nationality, England)} into \textit{(Newton, Nationality, Google)} does not reflect the nuanced errors encountered in practice. Real errors may involve subtle mismatches—such as \textit{(Bruce\_Lee, place\_of\_birth, China)}—where the entities are contextually related yet incorrect. This discrepancy highlights the need for more advanced error simulation and detection mechanisms.

%\paragraph{Contrastive Learning for KG Error Detection.}
Contrastive learning has emerged as a powerful self-supervised technique in various tasks such as classification~\cite{ZengX21}, link prediction~\cite{WangMCW21}, and recommendation~\cite{0013YYWC021}. By generating different views of the same data through transformations and then maximizing the agreement between these views~\cite{chen2020simple,henaff2020dataeff,hjelm2019learning,tack2020csi,chen2021novelty}, models can learn discriminative features without relying on manual labels. This approach has proven effective in computer vision for error detection, suggesting its potential applicability to KGs.
%\paragraph{Challenges in Adapting Contrastive Learning to KGs.}
However, applying contrastive learning to KG error detection introduces two primary challenges. First, constructing meaningful views for KGs is nontrivial; traditional graph augmentation methods (e.g., node dropping, edge perturbation, subgraph sampling)~\cite{you2020graph,hassani2020contrastive} are tailored for network embeddings and may disrupt the delicate structure of erroneous triples. Since a KG is essentially a collection of triples, effective error detection demands augmentation strategies that preserve the inherent error patterns. Second, existing graph encoders do not differentiate between reliable and noisy triples during message passing, potentially diluting the learned representations. Thus, an encoder that is aware of and can mitigate the influence of erroneous triples is needed.

%\paragraph{Proposed Framework.}
To address these issues, we formally define the problem of KG error detection and explore two central questions: \ding{182} How can we generate meaningful, distinct views of a KG to facilitate effective contrastive learning? \ding{183} How can we design an error-aware KG encoder that enhances the detection of noisy triples?
In response, we propose the ContrAstive knowledge Graph Error Detection (CAGED) framework. Our key contributions include:
\begin{itemize}
    \item Introducing CAGED, which integrates contrastive learning with KG embeddings to improve error detection.
    \item Developing a novel KG augmentation technique that produces triple-level views while preserving the structure of potential errors.
    \item Proposing an error-aware graph encoder, EaGNN, that employs a gated attention mechanism to suppress the influence of noisy triples during representation learning.
    \item Demonstrating the effectiveness of CAGED through extensive experiments on three real-world KGs, where it outperforms existing error detection methods.
\end{itemize}

\begin{figure*}
\centering
	\includegraphics[scale=0.2485]{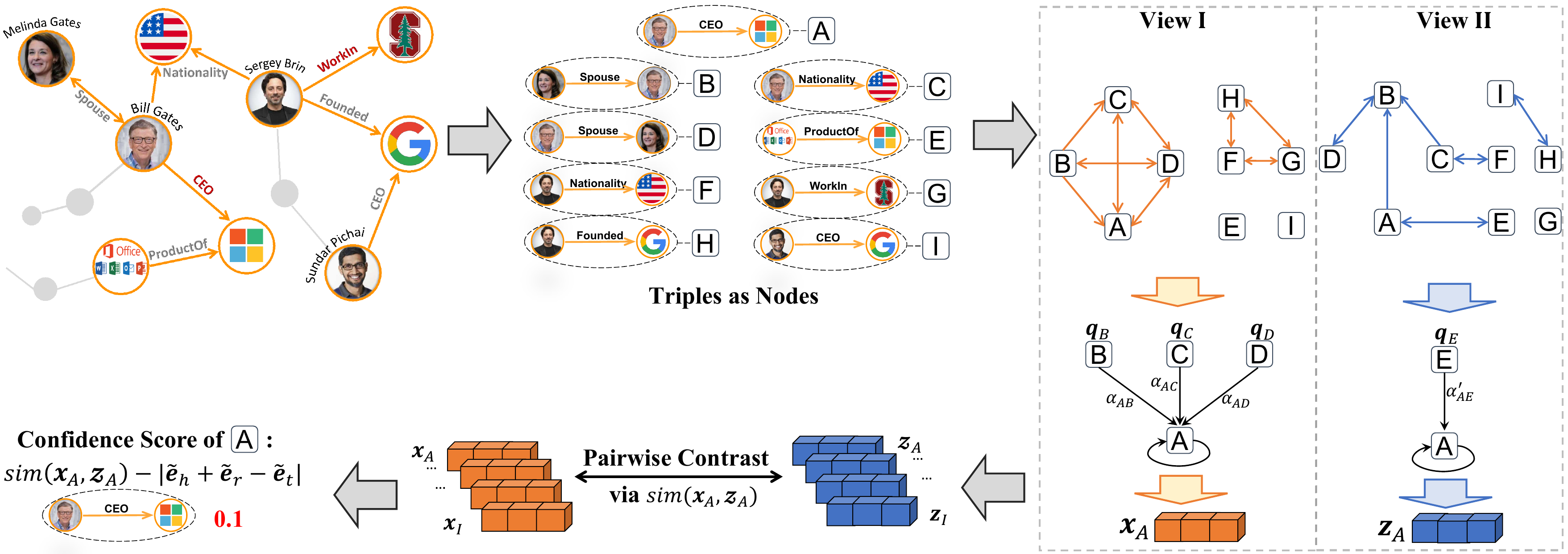}
	\vspace{-0.2cm}
	\caption{Two separate augmentation operators are applied to the original KG, generating two triple graphs as congruent views, i.e., View I and View II. After training, we estimate the confidence score by measuring the consistency of triple representations across multi-views, i.e., $\bm{x}_A$ and $\bm{z}_A$, and the self-consistency within the triple, i.e., $(\tilde{\bm{e}}_h, \tilde{\bm{e}}_r, \tilde{\bm{e}}_t)$.}
	\label{fig:caged}\vspace{-0.4cm}
\end{figure*}
\section{Contrastive Knowledge Graph Error Detection (CAGED) }
In conventional KG representation learning, a knowledge graph is typically treated as a heterogeneous graph where entities serve as nodes and relations as semantic links. However, this formulation often struggles to capture the intricate interdependencies among triples. In this work, we introduce ContrAstive knowledge Graph Error Detection (CAGED), a framework designed to pinpoint inaccuracies in large-scale KGs. The key idea is to recast the KG into multiple hyper-views—specifically, triple graphs—by regarding each relational triple as a node, and then to evaluate the reliability of each triple by comparing its representations across these views. The intuition is that contrastive learning can extract rich semantic features for normal triples, whereas anomalous triples, lacking these features, will have representations that diverge in the latent space. Thus, the degree to which a triple’s representations from different views converge serves as a reliable signal for error detection.

As depicted in Figure~\ref{fig:caged}, CAGED comprises three main components: KG augmentation, an error-aware encoder, and joint confidence estimation. First, we propose an innovative KG augmentation technique that produces two triple-level graphs by treating each relational triple as a node, thereby creating two congruent views (View I and View II). Next, we introduce a tailored error-aware knowledge graph neural network (EaGNN) that suppresses the influence of erroneous triples during representation learning. Finally, to obtain discriminative multi-view representations, we jointly optimize a translation-based KG embedding loss and a contrastive learning loss, and estimate each triple’s confidence by leveraging both the consistency across views and its internal coherence.

\subsection{Knowledge Graph Augmentation}\label{sec:KG_augmentation}
Data augmentation plays a pivotal role in contrastive learning~\cite{kipf2020contrastive}. Unlike image data—where standard techniques such as rotation, cropping, or distortion can easily yield diverse views~\citep{zhu2020deep,hassani2020contrastive}—constructing views for KGs is challenging due to their unique structure. Traditional graph contrastive learning methods typically use augmentations like node dropping, edge perturbation, subgraph sampling, or matrix diffusion~\cite{you2020graph}, which have shown promising results in tasks such as node classification, graph classification, and link prediction.

However, these methods are not ideal for KG error detection since a KG is essentially a collection of triples, and error detection entails identifying spurious triples. Approaches that emphasize entity or graph-level contrasts do not adequately capture the nuances required at the triple level. Moreover, while these augmentations can improve the robustness of representations, they may inadvertently distort the underlying error distribution or even introduce additional noise, thereby complicating the detection task.

\begin{definition}\label{definition: linking patterns}
	\noindent\textbf{Linking Pattern.} Consider two triples, $T_1=(h_1, r_1, t_1)$ and $T_2=(h_2, r_2, t_2)$, that share entities. They can be linked in two distinct ways: $(i)$ by sharing a head entity (i.e., $h_1 = h_2$ or $h_1 = t_2$), and $(ii)$ by sharing a tail entity (i.e., $t_1 = h_2$ or $t_1 = t_2$). Based on this observation, we construct two triple graphs using these linking patterns, as they capture different semantic relationships.
\end{definition}

Our augmentation strategy generates two triple graphs, denoted by $\mathcal{T}$ and $\mathcal{T'}$, by treating each relational triple as a node and connecting them according to the linking patterns in Definition~\ref{definition: linking patterns}. Concretely, $\mathcal{T}$ reflects connections based on shared head entities, whereas $\mathcal{T'}$ is built on shared tail entities. Given that triples sharing an entity are generally semantically related, normal triples typically have ample neighbors in $\mathcal{T}$ (View I) that can mirror the semantics captured in $\mathcal{T'}$ (View II). Thus, assessing the consistency between a triple's representations in these two views provides a reliable measure of its trustworthiness.

\subsection{Error-aware Encoder -- EaGNN}
Erroneous triples can severely impair the quality of learned representations by propagating misleading information during message passing. To counteract this, we propose a custom error-aware graph neural network (EaGNN) that captures both the semantic and structural properties of triples while mitigating the influence of noise. EaGNN integrates a local information modeling layer with a global error-aware attention mechanism to generate robust KG embeddings.

\paragraph{Local Information Modeling Layer.}
Transforming the original KG into a triple graph may result in a loss of the inherent sequential structure (i.e., the pattern $h \rightarrow r \rightarrow t$). To preserve this local information, we initialize the embeddings of entities and relations randomly and employ a set of Bi-LSTM units to learn the internal structure of each triple. For a given triple $(h, r, t)$, the process is formalized as:
\begin{align}
\label{eq:qi_revised}
    \tilde{\bf e}_h, \tilde{\bf e}_r, \tilde{\bf e}_t &= \text{Bi-LSTM}({\bf e}_h, {\bf e}_r, {\bf e}_t),\\
    {\bf q}_i &= [\tilde{\bf e}_h; \tilde{\bf e}_r; \tilde{\bf e}_t].
\end{align}
The derived triple embedding ${\bf q}_i$ effectively encapsulates the relational structure within the triple and serves as the initial representation in the constructed triple graphs.

\paragraph{Global Error-aware Attention Layer.}
In addition to local features, the global context provided by neighboring triples is essential for evaluating a triple's reliability. Standard KG neural networks often assign uniform attention to all neighbors, which can cause noisy information from erroneous triples to contaminate the final representation. To address this, we propose a novel attention mechanism that selectively attenuates the contribution of dubious neighbors.

For an anchor triple ${\bf q}_i \in \mathbb{R}^{d}$ in View I, we update its representation by aggregating information from its neighboring triples $\{{\bf q}_1, {\bf q}_2, \ldots, {\bf q}_m\}$. The attention weight between the anchor triple and a neighbor $j$ is computed as:
\begin{equation} %\small
\hat{\alpha}_{ij} = \mathcal{A}(\mathbf{W}\mathbf{q}_i, \mathbf{W}\mathbf{q}_j),
\label{eq:attention_revised}
\end{equation}
where $\mathbf{W} \in \mathbb{R}^{n \times d}$ is a learnable projection matrix, and $\mathcal{A}$ denotes an attention function. We then normalize these weights using softmax:
\begin{equation} %\small
\overline{\alpha}_{ij} = \frac{\exp(\hat{\alpha}_{ij})}{\sum_{k=1}^{m}\exp(\hat{\alpha}_{ik})}.
\label{eq:softmax_revised}
\end{equation}
To suppress the effect of potential outliers, we introduce a threshold $\mu \in \mathbb{R}$:
\begin{equation} %\small
\label{eq:mu_revised}
\alpha_{ij} =
\begin{cases}
\overline{\alpha}_{ij}, & \text{if } \overline{\alpha}_{ij} > \mu,\\[5mm]
0, & \text{otherwise}.
\end{cases}
\end{equation}
The final representation for the anchor triple in View I is then obtained by:
\begin{equation} %\small
\mathbf{x}_i = \sigma\Bigg(\sum_{j=1}^{m} \alpha_{ij}\, \mathbf{W}\, {\bf q}_j\Bigg),
\label{eq:xi_revised}
\end{equation}
and similarly, for an anchor triple in View II:
\begin{equation} %\small
\mathbf{z}_i = \sigma\Bigg(\sum_{j=1}^{m} \alpha_{ij}'\, \mathbf{W}\, {\bf q}_j\Bigg).
\label{eq:final_x_revised}
\end{equation}

\subsection{Joint Confidence Estimation}
To ensure that the model learns rich and discriminative representations across views, we jointly optimize two loss functions: a KG embedding loss and a contrastive learning loss.

\paragraph{KG Embedding Loss.}
At the level of individual triples, we leverage the translation principle—i.e., $\mathbf{h} + \mathbf{r} \approx \mathbf{t}$—to assess triple plausibility. We use the squared Euclidean distance as an energy function:
\begin{equation} %\small
E(h, r, t) = \left\|e_h + e_r - e_t\right\|_{2},
\end{equation}
and define the KG embedding loss as:
\begin{equation} %\small
\mathcal{L}_{kge} = \sum_{(h,r,t) \in \mathcal{G}} \sum_{(\hat{h},\hat{r},\hat{t}) \in \mathcal{\hat{G}}} \max\left(0, \gamma + E(h, r, t) - E(\hat{h}, \hat{r}, \hat{t})\right),
\label{eq:trans_loss_revised}
\end{equation}
where $\gamma > 0$ is a margin hyperparameter, $\mathcal{G}$ denotes the set of positive triples, and $\mathcal{\hat{G}}$ is constructed by corrupting the head or tail of each triple:
\begin{equation} %\small
\mathcal{\hat{G}} = \{(\hat{h}, r, t) \mid \hat{h} \in \mathcal{G}\} \cup \{(h, r, \hat{t}) \mid \hat{t} \in \mathcal{G}\}.
\end{equation}

\paragraph{Contrastive Loss.}
While the KG embedding loss focuses on local structural details, contrastive learning enables the model to capture global semantic consistency across views. We randomly sample a minibatch of $n$ triples from the KG, resulting in $2n$ representations: $\{{\bf x}_1, {\bf x}_2, \ldots, {\bf x}_n\}$ from View I and $\{{\bf z}_1, {\bf z}_2, \ldots, {\bf z}_n\}$ from View II. For each triple $i$, the pair $({\bf x}_i, {\bf z}_i)$ is treated as positive, while the remaining representations serve as negatives. The contrastive loss is formulated as:
\begin{equation} %\small
\mathcal{L}_{con}({\bf x}_i, {\bf z}_i) = -\log \frac{\exp\left(\operatorname{sim}({\bf x}_i, {\bf z}_i)/\tau\right)}{\sum_{j \in \{1,2,\dots,n\} \setminus \{i\}} \exp\left(\operatorname{sim}({\bf x}_i, {\bf z}_j)/\tau\right)},
\label{con_loss_revised}
\end{equation}
where $\operatorname{sim}({\bf x}_i, {\bf z}_i)$ denotes the cosine similarity between the two representations and $\tau$ is a temperature parameter. The overall contrastive loss is computed over all triples in the minibatch.

\begin{table}[t]
  \caption{The statistical information of the datasets.}
  \label{tab:DatasetStatistics}
\centering
\vspace{4mm}
\setlength{\tabcolsep}{6.5pt}
% \resizebox{230pt}{30pt}{
\begin{tabular}{lcccc}
\toprule
     Dataset &Entities   &Relations  &Triples    &Mean in-degree \\ \midrule
FB15K &$14$,$541$  &$237$    &$310$,$116$  &$18.71$     \\ 
 WN18RR  &$40$,$943$  &$11$    &$93$,$003$   &$2.12$     \\ 
 NELL-995 &$75$,$492$  &$200$    &$154$,$213$  &$1.98$     \\ \bottomrule
\end{tabular}
% }
% \vspace{-4mm}
\end{table}

\section{Experiments}
To assess the performance of our proposed CAGED framework, we conduct extensive experiments on three real-world knowledge graphs. Our evaluation is structured around the following research questions:
\begin{itemize}
    \item \textbf{Q1 (Effectiveness):} How does CAGED compare to state-of-the-art methods for KG error detection?
    \item \textbf{Q2 (Ablation Study):} What are the contributions of each individual component within CAGED?
    \item \textbf{Q3 (Parameter Analysis):} How do variations in hyperparameters impact the performance of CAGED?
\end{itemize}

\subsection{Experimental Settings}
This section details the experimental configuration, including datasets, baseline methods, evaluation metrics, and implementation specifics.

\paragraph{Datasets.}
Following previous work~\cite{xie2018does, jia2019triple}, we evaluate our approach on three datasets derived from established benchmarks: FB15k, WN18RR, and NELL-995. Each dataset is constructed by injecting noise at levels of 5\%, 10\%, and 15\% of the total triples. Table~\ref{tab:DatasetStatistics} summarizes their  statistics. In this paper, we detect errors from three categories, including factual errors that include incorrect entity relationships, coverage errors that arise from missing entities or relations, and logical inconsistencies that violate formal rules, like circular dependencies or conflicting facts. 
% Additionally, provenance errors stem from unreliable data sources or extraction inaccuracies during KG construction. These errors propagate downstream tasks like reasoning or question answering, emphas

\noindent\textbf{FB15K} originates from the Freebase Knowledge Base and is augmented with textual relations extracted from ClueWeb12 and annotated in Freebase.

\noindent\textbf{WN18RR} is a refined subset of WordNet designed to eliminate test leakage by removing inverse relations; it comprises 11 relation types and features a relatively simpler structure.

\noindent\textbf{NELL-995} is obtained from the 995th iteration of the NELL system, a large-scale and evolving knowledge base. Triples lacking reasoning values are removed prior to selecting the top-200 unique relations.

\paragraph{Baseline Methods.}
We compare CAGED with two groups of baseline approaches. The first group consists of KG embedding techniques, including \textbf{TransE}~\cite{bordes2013translating}, \textbf{DistMult}~\cite{yang2015embedding}, and \textbf{ComplEx}~\cite{trouillon2016complex}. In these methods, the confidence score for each triple is computed from the embedding-based score (e.g., $\|{\bm e}_h+{\bm e}_r - {\bm e}_t\|_2$ in TransE). The second group comprises specialized KG error detection methods such as \textbf{CKRL}~\cite{xie2018does}, \textbf{KGTtm}~\cite{jia2019triple}, and \textbf{KGIst}~\cite{BelthZVK20}. CKRL extends TransE by incorporating all paths between entities, KGTtm further integrates the overall graph structure, and KGIst employs an unsupervised strategy to learn soft rules for error identification.

\paragraph{Evaluation Metrics.}
We employ ranking-based metrics to assess performance. Triples are sorted in ascending order according to their confidence scores, with lower scores indicating a higher likelihood of error. Specifically, we use:
\begin{itemize}
    \item \textbf{Precision@K}: The proportion of false triples within the top K triples.
    \item \textbf{Recall@K}: The fraction of all erroneous triples identified in the top K.
\end{itemize}

\begin{table*}[t]
\centering%\smalll
% \vspace{2pt}
\caption{Error detection results of Precision@K and Recall@K based on the three datasets with anomaly ratio $= 5\%$.}
\resizebox{1\textwidth}{!}{
\vspace{0.2cm}
%\resizebox{490pt}{100pt}{
\begin{tabular}{llccccccccccccccc}
\toprule
& &\multicolumn{5}{c}{FB15K} &\multicolumn{5}{c}{WN18RR} &\multicolumn{5}{c}{NELL-995} \cr \cmidrule(lr){3-7} \cmidrule(lr){8-12} \cmidrule(lr){13-17} 
& &$K$=$1\%$  &$K$=$2$\%  &$K$=$3$\%  &$K$=$4$\%  &$K$=$5$\%$^a$ & $K$=$1\%$  &$K$=$2$\%  &$K$=$3$\%  &$K$=$4$\%  &$K$=$5$\%$^a$ & $K$=$1\%$  &$K$=$2$\%  &$K$=$3$\%  &$K$=$4$\%  &$K$=$5$\%$^a$ \\ \cmidrule(lr){3-7} \cmidrule(lr){8-12} \cmidrule(lr){13-17} 
% \multirow{9}{*}{$Precision@K$}
\parbox[t]{5mm}{\multirow{7}{*}{\rotatebox[origin=c]{90}{$Precision@K$}}}
& TransE
 &0.756 &0.674 &0.605 &0.546 &0.488 
 &0.581 &0.488 &0.371 &0.345 &0.331
 &0.659 &0.550 &0.476 &0.423 &0.383 \\ 
& ComplEx
 &0.718 &0.651 &0.590 &0.534 &0.485 
 &0.518 &0.444 &0.382 &0.341 &0.307
 &0.627 &0.538 &0.472 &0.427 &0.378  \\ 
& DistMult
 &0.709 &0.646 &0.582 &0.529 &0.483
 &0.574 &0.451 &0.390 &0.349 &0.322
 &0.630 &0.553 &0.493 &0.446 &0.408 \\
& CKRL
 &0.789 &0.736 &0.684 &\underline{0.630} &0.574
 &0.675 &0.526 &0.436 &0.389 &0.349 
 &0.735 &0.642 &0.559 &0.498 &0.450 \\
& KGTtm
 &0.815 &\underline{0.767} &\underline{0.713} &0.612 &\underline{0.579} 
 &\underline{0.770} &\underline{0.628} &\underline{0.516} &\underline{0.444} &\underline{0.396}
 &\underline{0.808} &\underline{0.691} &\underline{0.602} &\underline{0.535} &0.481 \\
& KGIst
 &\underline{0.825} &0.754 &0.703 &0.617 &0.569
 &0.747 &0.599 &0.476 &0.407 &0.379
 &0.782 &0.678 &0.584 &0.528 &\underline{0.485} \\
& CAGED 
 &\textbf{0.852} &\textbf{0.796} &\textbf{0.735} &\textbf{0.665} &\textbf{0.595} 
 &\textbf{0.826} &\textbf{0.726} &\textbf{0.632} &\textbf{0.541} &\textbf{0.469} 
 &\textbf{0.850} &\textbf{0.736} &\textbf{0.644} &\textbf{0.573} &\textbf{0.516} \\ \cmidrule(lr){3-7} \cmidrule(lr){8-12} \cmidrule(lr){13-17} 
\parbox[t]{5mm}{\multirow{7}{*}{\rotatebox[origin=c]{90}{$Recall@K$}}}
&TransE%~\cite{bordes2013translating}
 &0.151 &0.270 &0.363 &0.437 &0.488 
&0.116 &0.195 &0.223 &0.276  &0.331 
&0.132 &0.220 &0.285 &0.338  &0.383  \\
&ComplEx%~\cite{trouillon2016complex} 
&0.143 &0.260 &0.354 &0.427 &0.485 
&0.103 &0.177 &0.229 &0.273 &0.307  
&0.125 &0.215 &0.283 &0.341 &0.378 \\
&DistMult%~\cite{yang2015embedding} 
&0.141 &0.258 &0.349 &0.423 &0.483 
&0.114 &0.180 &0.234 &0.279 &0.322  
&0.126 &0.221 &0.295 &0.357 &0.408 \\
&CKRL%~\cite{xie2018does}  
&0.158 &0.294 &0.410 &\underline{0.504} &0.574 
&0.135 &0.210 &0.261 &0.311 &0.349  
&0.147 &0.256 &0.335 &0.398 &0.450 \\ 
&KGTtm%~\cite{jia2019triple} 
&0.163 &\underline{0.307} &\underline{0.428} &0.490 &\underline{0.579} 
&\underline{0.154} &\underline{0.251} &\underline{0.309} &\underline{0.355} &\underline{0.396}  
&\underline{0.161} &\underline{0.276} &\underline{0.361} &\underline{0.428} &0.481 \\
&KGIst%~\cite{BelthZVK20}
 &\underline{0.165} &0.302 &0.422 &0.494 &0.569
 &0.149 &0.240 &0.285 &0.325 &0.379
 &0.156 &0.271 &0.350 &0.422 &\underline{0.485} \\
&CAGED 
&\textbf{0.171} &\textbf{0.318} &\textbf{0.441} &\textbf{0.532} &\textbf{0.595} 
&\textbf{0.165} &\textbf{0.290} &\textbf{0.379} &\textbf{0.433} &\textbf{0.469} 
&\textbf{0.170} &\textbf{0.294} &\textbf{0.386} &\textbf{0.458} &\textbf{0.516} \\\bottomrule
\end{tabular}
}
% \tabnote{$^{\rm a}$This footnote shows what footnote symbols to use.}
% \footnotesize{$^a$ Please note that according to Eq.\eqref{equ:metric}, when $K$ equals the anomaly ratio (i.e. $5\%$), $Precision@K$ and $Recall@K$ are equal.}
% \footnotesize{$^a$ It is noted that results at $K=5\%$ share the same $Precision@K$ and $Recall@K$ since the percentage of errors is also $5\%$. }\\
%}
\label{tab:performance1}
% \vspace{-2mm}
\end{table*}

\paragraph{Implementation Details.}
All methods are implemented using PyTorch, and we utilize publicly available code for baseline models. Experiments are performed on an Nvidia RTX 3090 GPU. We use the Adam optimizer with a fixed batch size of $256$, an initial learning rate of $0.01$, and an embedding dimension of $100$ for all models. A grid search is conducted to tune hyperparameters: the attention threshold $\mu$ (ranging from $0.001$ to $0.2$), the margin parameter $\gamma$ (ranging from $0$ to $1$), and the trade-off coefficient $\lambda$ (ranging from $0.001$ to $1000$). The number of neighbors is determined by averaging the neighbor count across all triples in each dataset. To reduce variability, results are averaged over ten runs using a fixed random seed.

\begin{table*}[t]
\centering%\small
% \vspace{2pt}
\caption{Error detection results on NELL-995 with different anomaly ratios.} \label{tab:ratio}
\resizebox{1\textwidth}{!}{
\vspace{-0.2cm}
\begin{tabular}{llccccccccc}
\toprule
&Ratio &\multicolumn{3}{c}{5\%} &\multicolumn{3}{c}{10\%} &\multicolumn{3}{c}{15\%} \cr \cmidrule(lr){3-5} \cmidrule(lr){6-8} \cmidrule(lr){9-11} 
&$K$ &$K$=$5$\%$^a$  &$K$=$10$\%  &$K$=$15$\% &$K$=$5$\%  &$K$=$10$\%$^a$  &$K$=$15$\%  &$K$=$5$\%  &$K$=$10$\%  &$K$=$15$\%$^a$ \\\cmidrule(lr){3-5} \cmidrule(lr){6-8} \cmidrule(lr){9-11} 
% \multirow{7}{*}{$Precision@K$}
\parbox[t]{5mm}{\multirow{7}{*}{\rotatebox[origin=c]{90}{$Precision@K$}}}
&TransE
 &0.383 &0.285 &0.225
 &0.626 &0.499 &0.407
 &0.702 &0.621 &0.535 \\ 
&ComplEx
 &0.378 &0.289 &0.231
 &0.614 &0.507 &0.402
 &0.696 &0.589 &0.528  \\ 
&DistMult
 &0.408 &0.298 &0.227
 &0.633 &0.510 &0.414
 &0.718 &0.618 &0.548\\
&CKRL
 &0.450 &0.306 &0.236
 &0.679 &0.524 &0.421
 &0.745 &0.646 &0.560\\
&KGTtm
 &0.481 &\underline{0.320} &0.242
 &0.713 &0.527 &0.437
 &0.788 &\underline{0.673} &\underline{0.576}\\
 &KGIst
 &\underline{0.485} &0.317 &\underline{0.244}
 &\underline{0.748} &\underline{0.552} &\underline{0.440}
 &\underline{0.791} &0.663 &0.569\\
&CAGED 
 &\textbf{0.516} &\textbf{0.325} &\textbf{0.251}
 &\textbf{0.799} &\textbf{0.585} &\textbf{0.458}
 &\textbf{0.823} &\textbf{0.729} &\textbf{0.599}\\ \cmidrule(lr){3-5} \cmidrule(lr){6-8} \cmidrule(lr){9-11} 
\parbox[t]{5mm}{\multirow{7}{*}{\rotatebox[origin=c]{90}{$Recall@K$}}}
&TransE
&0.383 &0.57 & 0.675 
&0.313 &0.499 &0.612 
&0.234 &0.414 &0.535 \\ 
&ComplEx
&0.378 &0.578 &0.693 
&0.307 &0.507 &0.603 
&0.232 &0.393 &0.528  \\ 
&DistMult
&0.408 &0.596 &0.681 
&0.317 &0.510 &0.621 
&0.239 &0.412 &0.548\\
&CKRL
&0.450 &0.612 &0.708 
&0.340 &0.524 &0.632 
&0.248 &0.431 &0.560\\
&KGTtm
&0.481 &\underline{0.640} &0.726 
&0.357 &0.527 &0.656 
&0.263 &\underline{0.449} &\underline{0.576}\\
&KGIst
&\underline{0.485} &0.634 &\underline{0.732}
&\underline{0.374} &\underline{0.552} &\underline{0.660} 
&\underline{0.264} &0.442 &0.569\\
&CAGED 
 &\textbf{0.516} &\textbf{0.650} &\textbf{0.753}
 &\textbf{0.400} &\textbf{0.585} &\textbf{0.687}
 &\textbf{0.274} &\textbf{0.486} &\textbf{0.599}\\ \bottomrule
 
\end{tabular}
}
% \\
\footnotesize{$^a$ Note that when $K$ equals anomaly ratio, $Precision@K$ and $Recall@K$  have the same value.}
\label{tab:performance_CAGED}
% \vspace{-2mm}
\end{table*}

\subsection{Effectiveness (\textbf{Q1})}
To address Q1, we evaluate all methods on the three datasets with a noise level of 5\%. Table~\ref{tab:performance_CAGED} summarizes the results (similar trends are observed for other noise levels, as shown in Table~\ref{tab:performance_CAGED}). Our key observations are:
\begin{enumerate}
    \item KG error detection methods (including CAGED, CKRL, KGTtm, and KGIst) outperform standard KG embedding techniques (e.g., TransE, ComplEx, and DistMult) because the latter do not explicitly account for erroneous triples.
    \item CAGED consistently achieves the highest performance in terms of both recall and precision. For instance, when evaluated at $K$ corresponding to 5\% and with a 5\% anomaly ratio, CAGED improves over the second-best method by approximately $1.6\%$, $7.3\%$, and $3.1\%$ on FB15K, WN18RR, and NELL-995, respectively. This improvement is attributed to the multi-view contrastive learning strategy that effectively captures both intra-triple and cross-view features.
    \item The superiority of CAGED is particularly pronounced when focusing on the top-ranked erroneous triples (i.e., for smaller $K$ values).
\end{enumerate}
Furthermore, experiments on NELL-995 with noise levels of 5\%, 10\%, and 15\% (see Table~\ref{tab:ratio}) confirm the robustness and stability of our model under varying conditions.

\begin{table}[t]
%\smalll
  \caption{Four pairs of variants of CAGED.}
  \label{tab:variantsname}
\centering
\vspace{0.2cm}
\resizebox{0.7\textwidth}{!}{
% \resizebox{200pt}{42pt}{
\begin{tabular}{lcc} 
\toprule
 & Original component & Replacement\\ \midrule
Var\_DN &    KG augmentation  & Node dropping \\ 
Var\_EP &  KG augmentation  & Edge perturbation \\  \cmidrule(lr){2-2} \cmidrule(lr){3-3}
Var\_Concat & Bi-LSTM units &  Only concatenation\\
Var\_LSTM &  Bi-LSTM units &  LSTM units  \\ \cmidrule(lr){2-2} \cmidrule(lr){3-3}
Var\_GCN & EaGNN & R-GCN\\ 
Var\_GAT & EaGNN & KGAT\\  \cmidrule(lr){2-2} \cmidrule(lr){3-3}
Var\_Local &Joint optimization& Only negative sampling\\
Var\_Global&Joint optimization& Only contrastive learning\\
\bottomrule
\end{tabular}
\vspace{-0.5cm}
}
\end{table}

\subsection{Ablation Study (\textbf{Q2})}\label{sec:abaltion_study}
To investigate Q2, we conduct an ablation study on the NELL-995 dataset by evaluating several variants of CAGED (see Table~\ref{tab:variantsname}). Due to space limitations, we only report results for NELL-995, as similar trends are observed for FB15K and WN18RR.

\begin{table}[t]
\centering%\smalll
% \vspace{3pt}
\caption{The comparisons of CAGED and its variants on NELL-995 with ratio of errors equals $5\%$.}
\setlength{\tabcolsep}{0.7mm}{}{
% \vspace{-0.2cm}
\begin{tabular}{lcccccccccc}
\toprule
&\multicolumn{5}{c}{$Precision@K$} &\multicolumn{5}{c}{$Recall@K$}  \\ \cmidrule(lr){2-6} \cmidrule(lr){7-11}
Top@K  &1\%  &2\%  &3\%  &4\%  &5\% &1\%  &2\%  &3\%  &4\%  &5\%\\  \cmidrule(lr){2-6} \cmidrule(lr){7-11}
CAGED 
&\textbf{$.850$} &\textbf{$.736$} &\textbf{$.644$} &\textbf{$.573$} &\textbf{$.516$} 
&$.170$ &$.294$ &$.386$ &$.458$ &$.516$\\  
\cmidrule(lr){2-6} \cmidrule(lr){7-11}
Var\_DN
&$.613$ &$.490$ &$.412$ &$.358$ &$.325$
&$.122$ &$.196$ &$.247$ &$.286$ &$.325$\\
Var\_EP  
&$.593$ &$.500$ &$.416$ &$.354$ &$.313$
&$.118$ &$.200$ &$.249$ &$.283$ &$.313$\\   \cmidrule(lr){2-6} \cmidrule(lr){7-11}
Var\_Concat
&$.767$ &$.648$ &$.532$ &$.468$ &$.440$
&$.153$ &$.259$ &$.319$ &$.374$ &$.440$\\
Var\_LSTM
&$.797$ &$.662$ &$.564$ &$.500$ &$.458$
&$.159$ &$.264$ &$.338$ &$.400$ &$.458$\\ 
\cmidrule(lr){2-6} \cmidrule(lr){7-11}
Var\_GCN 
&$.760$ &$.623$ &$.529$ &$.464$ &$.414$
&$.152$ &$.249$ &$.317$ &$.371$ &$.414$\\ 
Var\_GAT
&$.782$ &$.629$ &$.551$ &$.471$ &$.426$
&$.156$ &$.251$ &$.330$ &$.377$ &$.426$\\  \cmidrule(lr){2-6} \cmidrule(lr){7-11}
Var\_Local
&$.724$ &$.603$ &$.504$ &$.433$ &$.383$
&$.144$ &$.241$ &$.302$ &$.346$ &$.383$\\
Var\_Global
&$.783$ &$.652$ &$.560$ &$.492$ &$.443$
&$.156$ &$.261$ &$.336$ &$.393$ &$.443$\\ 
\bottomrule 
\end{tabular}}
\label{tab:variants}
% \vspace{-0.5cm}
\end{table}
\paragraph{Component 1: KG Augmentation.}
We compare our tailored KG augmentation strategy with two generic augmentation methods: node dropping (Var\_DN) and edge perturbation (Var\_EP). As illustrated in Table~\ref{tab:variants}, both Var\_DN and Var\_EP underperform, even relative to some KG embedding baselines like TransE. This suggests that applying standard augmentations can disrupt the triple structure and introduce additional noise, thereby impairing error detection. The significant performance gap highlights the importance of our specialized augmentation approach, which preserves the distribution of potential errors while generating diverse views.

% \captionsetup[figure]{}
\begin{figure*}[tbp]
\centering
  \setlength{\abovecaptionskip}{1pt}
  \subfigcapskip=-3pt
  \subfigure[Impact of hyper-parameter $\mu$]{
    \label{fig:mu}
    \includegraphics[width=0.3\textwidth]{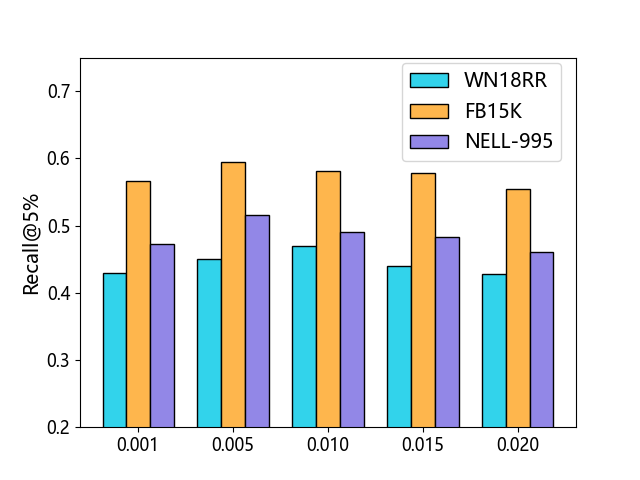}}
  % \hspace{2mm}
  \subfigure[Hyper-parameter $\lambda$]{
    \label{fig:lambda}
    \includegraphics[width=0.3\textwidth]{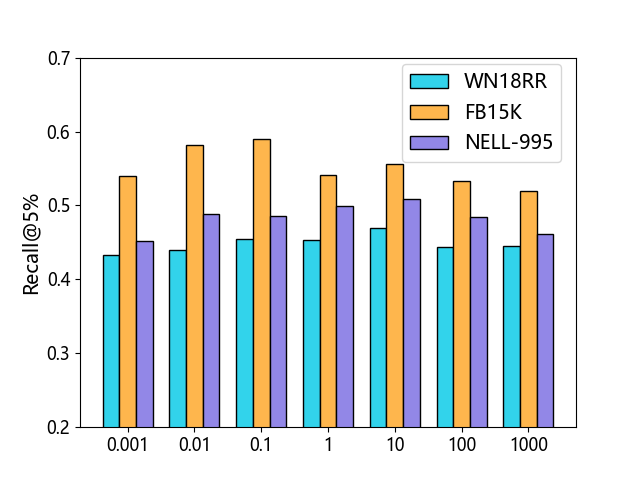}}
  % \hspace{2mm}
  \subfigure[Hyper-parameter $\gamma$]{
    \label{fig:gamma}
    \includegraphics[width=0.3\textwidth]{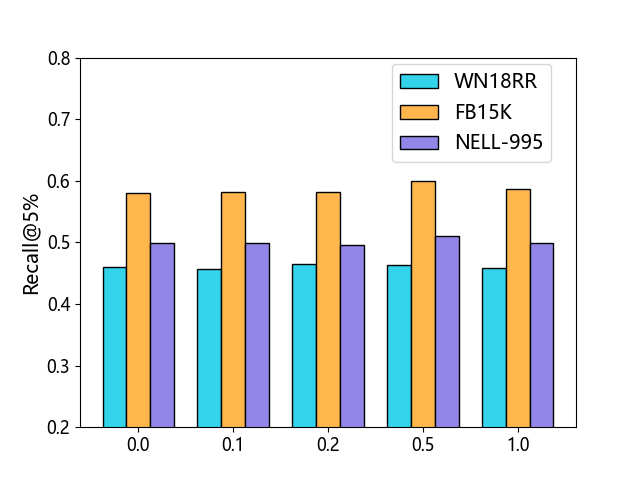}}
    %\vspace{-0.2cm}
  \caption{Impact of hyperparameters on the three datasets.}
\label{fig:parameter}
\vspace{-0.3cm}
\end{figure*}

\paragraph{Component 2: EaGNN Encoder.}
To validate the effectiveness of our EaGNN encoder, we substitute it with leading KG neural networks such as R-GCN~\cite{schlichtkrull2018modeling} (Var\_GCN) and KGAT~\cite{nathani-etal-2019-learning} (Var\_GAT). Results in Table~\ref{tab:variants} indicate that, although Var\_GAT performs better than Var\_GCN, both still lag behind the full CAGED model. This outcome underscores that traditional KGNNs, which assume all triples are correct, are less capable of filtering out noise. Additionally, we test variants that modify the local structure modeling: Var\_LSTM (using LSTM instead of Bi-LSTM) and Var\_Concat (direct concatenation of entity, relation, and tail embeddings). The drop in performance for Var\_Concat—and to a lesser extent for Var\_LSTM—demonstrates the critical role of capturing local sequential structure via Bi-LSTM units.

\paragraph{Component 3: Joint Estimation.}
We also assess the impact of our joint optimization strategy by comparing two variants: Var\_Local, which relies solely on the translation-based KG embedding loss (i.e., using negative sampling), and Var\_Global, which employs only the contrastive learning loss. As shown in Table~\ref{tab:variants}, both variants are outperformed by the integrated model, with Var\_Local performing particularly poorly, even when compared to simple baselines like DistMult. These findings confirm that combining local structural features with cross-view consistency is essential for robust error detection.

\subsection{Parameter Analysis (\textbf{Q3})}\label{sec:parameter}
We further explore the influence of key hyperparameters on model performance, with the results summarized in Figure~\ref{fig:parameter}.

The attention threshold $\mu$, regulates the selectivity of the attention mechanism. A larger $\mu$ limits attention to a small set of highly related neighbors, whereas a smaller $\mu$ allows a broader neighborhood to influence the representation. As shown in Figure~\ref{fig:mu}, optimal performance is achieved when $\mu$ is approximately $0.005$ for FB15K and NELL-995, and $0.01$ for WN18RR. Decreasing $\mu$ further introduces excessive noise, leading to performance degradation.

The coefficient $\lambda$, which balances the contribution of cross-view inconsistency and intra-triple coherence, is varied from $10^{-3}$ to $10^{3}$. Figure~\ref{fig:lambda} shows that WN18RR and NELL-995 obtain their best results around $\lambda=10$, while FB15K performs optimally when $\lambda$ is below $1$ (around $0.1$). This difference is likely due to the denser structure of FB15K, which allows the cross-view signal to play a more significant role.

Finally, the margin parameter $\gamma$ is varied from $0.1$ to $1.0$. As illustrated in Figure~\ref{fig:gamma}, the detection performance remains relatively stable across this range, suggesting that the joint training strategy helps prevent the model from converging to suboptimal solutions.

\section{Summary}
Traditional KG error detection methods typically rely on synthetically generated false triples—created by randomly replacing head or tail entities—which fail to capture the nuanced errors found in practice. In contrast, our CAGED framework leverages contrastive learning to generate multi-view representations centered on triples, allowing the model to discern subtle inconsistencies that indicate errors. Experimental results on three real-world knowledge graphs demonstrate that CAGED outperforms existing state-of-the-art methods across various evaluation metrics and noise levels.

\chapter{Integrating Entity Attributes for Error-Aware Reasoning}
\section{Introduction}
Knowledge graphs (KGs) consolidate vast amounts of relational data as triples, i.e., \emph{(head entity, relation, tail entity)}. They appear in both general-purpose settings (e.g., YAGO~\cite{Mahdisoltani-etal15YAGO}, DBpedia~\cite{Lehmann-etal15DBpedia}) and in specialized domains such as biomedicine or agriculture. KGs serve as the backbone for many AI applications, including recommender systems enhanced by KG information~\cite{wang2019knowledge} and conversational agents powered by structured knowledge~\cite{huang2019knowledge}.

A prominent research direction involves KG embedding, where entities are represented as continuous vectors and relations as operations (e.g., translations or projections) in a shared latent space. This approach enables efficient inference via simple numerical computations~\cite{TransE,rotate,guo2018knowledge}. Despite the extensive development of embedding techniques, the adverse impact of noisy or erroneous triples is frequently neglected. Manual curation is infeasible at scale, and most real-world KGs are automatically extracted from web corpora using heuristic methods~\cite{bollacker2008freebase, Lehmann-etal15DBpedia, Heindorf-etal16Vandalism}. Consequently, a significant fraction of the triples are noisy. For example, the NELL KG~\cite{mitchell2018never} contains about 2.4 million triples with an estimated accuracy of 74\%, indicating roughly 0.6 million erroneous triples~\cite{jia2019triple}. Such inaccuracies can severely degrade downstream application performance, underscoring the need for error-aware KG embedding methods.

Developing robust, error-aware embeddings is challenging due to the unknown and diverse nature of KG errors. Recent efforts have explored mitigating noise by assigning reliability scores to triples~\cite{shan2018confidence, nayyeri2021pattern}. Early works such as Vault~\cite{dong2014knowledge} estimated triple reliability via prior models, while methods like CKRL~\cite{xie2018does} and NoiGAN~\cite{cheng2019noigan} further refined this idea by leveraging internal structural cues. However, solely relying on graph topology may not suffice; for example, a triple like \{London, is\_larger\_than, Washington\} might be ambiguous without additional context to resolve which “London” is intended.
\begin{figure}[t!]
\centering
\begin{center}
	\includegraphics[scale=1.]{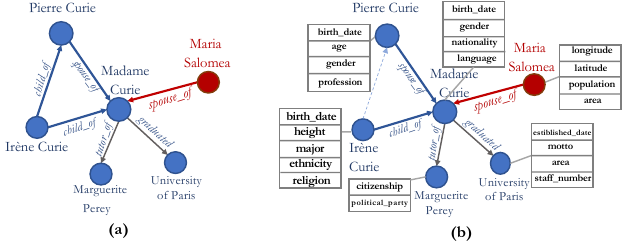}
	% \vspace{-0.4cm}
\end{center}
    \caption{Running examples of complex errors in real-world KGs. (a) presents a KG error with mismatched head/tail entities and relations. This error can be detected by reasoning over neighboring triples. But, real-world KGs are often incomplete and noisy. (b) demonstrates a further difficult case with some key links missing. In this case, the rich semantics in entity attribute types can facilitate the detection of errors.}
% 	\caption{\blue{Running examples of complex errors in real-world KGs. (a) presents a common linking error with mismatched head/tail entities and relations. (b) shows a complex scenario where the error can only be detected by reasoning over neighboring triples. However, real-world KGs are often incomplete and sparse, (c) demonstrates a further difficult error with some key links missing. In such a situation, the rich semantics in entity attribute types facilitate the detection of errors.}}
% % 	{\red Comments 2\_345.}
	\label{fig:figure0}
	\vspace{-0.5cm}
\end{figure}

Many KGs are also accompanied by rich attribute data that describe entity properties. For instance, as illustrated in Figure~\ref{fig:figure0}, an entity such as \emph{MadameCurie} may be associated with attributes like \{birth\_date, gender, nationality, language\}, highlighting its identity as a person, whereas an entity like \emph{MariaSalomea} might include \{latitude, longitude, population, area\}, reflecting its geographical nature. The interplay between an entity’s attributes and the KG structure provides valuable signals for error detection. Typically, entities with similar attribute profiles are connected via specific relations; for example, those with geographical attributes are often linked by \emph{live\_in} or \emph{born\_in}, while entities with literary characteristics (e.g., \{pen\_name, writing\_style\}) tend to be connected by \emph{author\_of}. Thus, if a triple connects entities whose attributes are inconsistent with the stated relation (e.g., \{MadameCurie, spouse\_of, MariaSalomea\} where a person is linked to a location), it likely indicates an error.

Nonetheless, integrating attribute information into error-aware KG embedding poses two key challenges. First, entity attributes are highly heterogeneous—varying in type, number, and context—so that a uniform encoder is needed to fuse diverse attribute information. An entity may exhibit different attribute sets in different roles (e.g., a scientist versus a writer). Second, aligning the semantics extracted from attributes with the KG’s structural information is nontrivial, as these components inherently differ in characteristics. While some approaches directly merge attribute-derived features with entity embeddings, they often fail to capture the nuanced correlations required for effectively filtering out erroneous triples.

This work seeks to answer the following research questions:  \ding{182} How can entities, relations, and attributes be jointly embedded into a unified vector space? \ding{183} How can we design an effective detector to compute a confidence score for each triple based on the learned features? \ding{184} How can these confidence scores be utilized to improve error-aware KG embeddings?
To address these challenges, we propose a novel framework called \textbf{AEKE} (Attributed Error-aware Knowledge Embedding)~\cite{zhang2023integrating}. The core idea of AEKE is to exploit the correlation between KG structure and entity attributes to emphasize reliable triples during embedding. Specifically, we construct two triple-level hypergraphs—the \emph{relational hypergraph} and the \emph{attribute hypergraph}—to capture, respectively, the structural topology of the KG and the semantics conveyed by entity attributes. A contrastive learning framework is then employed to learn complementary representations from these two views. By jointly analyzing three types of anomaly signals—(i) intra-triple self-contradiction, (ii) cross-triple consistency, and (iii) the alignment between attribute information and graph structure—our method computes a confidence score for each triple. These scores are further used to adaptively adjust both the aggregation weights in the contrastive learning process and the margin loss in KG embedding, thereby reducing the influence of noisy triples.

The primary contributions of this work are as follows:
\begin{itemize}
    \item We introduce AEKE, a novel KG embedding framework that leverages entity attributes to enable error-aware learning.
    \item We design a multi-view contrastive learning strategy that uses attribute information as a complementary view to guide KG representation learning.
    \item We propose a mechanism for computing triple confidence scores by integrating signals from intra-triple, cross-triple, and attribute-structure consistency.
    \item We employ these confidence scores to adaptively update the aggregation in multi-view learning and the margin loss in KG embedding, thereby mitigating the impact of erroneous triples.
\end{itemize}

\begin{figure*}[t]
\begin{center}	\includegraphics[scale=0.67]{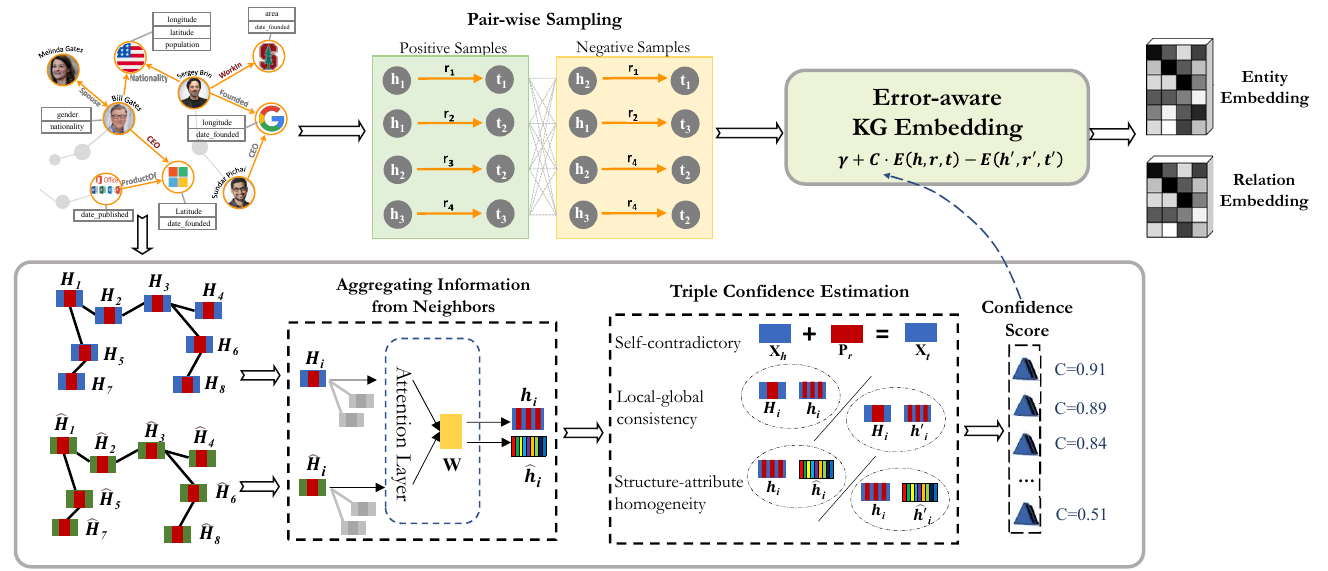}
	\caption{We perform a relation-induced construction process to build the \emph{relational hypergraph}, and construct the \emph{attribute hypergraph} based on entity attributes. These two hypergraphs can be regarded as congruent views of the target KG.  A contrastive learning framework is used to learn the representation of each instance from these two different views. Meanwhile, a triple confidence estimation module is designed to calculate the confidence score of each triple by considering: self-contradictory within the triple; local-global consistency in graph structure; and structure-attribute homogeneity. Under the joint adaptive training scheme, we leverage confidence scores to adaptively update the weighted aggregation in contrastive learning and margin loss in KG embedding, such that potential errors would contribute little to KG learning.}
\label{fig:figure1_AEKE}
\end{center}
\end{figure*}
\section{Error-Aware Knowledge Graph Modeling}
% \section{Error-Aware Knowledge Graph Embedding}
We present AEKE, a framework designed to learn error-aware knowledge graph (KG) embeddings by incorporating entity attributes. As shown in Fig.~\ref{fig:figure1_AEKE}, AEKE is comprised of three main components: \ding{182} the KG representation learning model, which integrates a confidence score $C(h, r, t)$ into traditional KG embedding models to mitigate the impact of noise on embedding vectors; \ding{183} the triple confidence learning module, which calculates a confidence score to assess the correctness and significance of each triple based on both internal structural information and external heterogeneous attribute data; \ding{184} a joint adaptive training scheme, which optimizes both the KG representation learning and the confidence learning models in an end-to-end manner.

\subsection{Knowledge Graph Representation Learning}
Knowledge graphs are effective tools for storing and handling structured data on real-world entities and facts, making them essential in many knowledge-driven applications. However, real-world KGs are often riddled with errors due to noisy sources and imperfect extraction methods during construction. Most existing KG embedding techniques assume all triples are correct, which results in the overfitting of noisy information into embeddings, causing significant performance degradation in downstream tasks.

To address this issue, we introduce the concept of a confidence score to distinguish between noisy and accurate triples. The confidence value ranges from $[0, 1]$, where values closer to 0 indicate a higher likelihood of error. By incorporating confidence, we propose a new error-aware objective function to filter out noisy triples from the embedding model's learning process:
\begin{equation} \label{eq:margin_loss}
\begin{aligned}
    L_{emb}=\sum_{(h, r, t) \in \mathcal{G}} \sum_{\left(h^{\prime}, r^{\prime}, t^{\prime}\right) \in \mathcal{G}^{\prime}} \max (0, \gamma + C(h, r, t) \cdot E(h, r, t) \\- E(h^{\prime}, r^{\prime}, t^{\prime}) ),
\end{aligned}
\end{equation}
where $E(h, r, t)=\left\|e_{h}+e_{r}-e_{t}\right\|_{2}$ is the energy score based on the translational assumption. $\gamma > 0$ is the margin hyperparameter, and $\mathcal{G}$ represents the set of positive triples. The triple confidence $C(h, r, t)$ helps the model focus more on convincing triples, thus improving embedding learning. For negative sample generation, since explicit negative triples are not available, we follow a corruption strategy:
\begin{equation}
\begin{aligned}
\mathcal{G}^{\prime}=&\left\{\left(h^{\prime}, r, t\right) \mid h^{\prime} \in \mathcal{E}\right\} \cup\left\{\left(h, r, t^{\prime}\right) \mid t^{\prime} \in \mathcal{E}\right\} \\
& \cup\left\{\left(h, r^{\prime}, t\right) \mid r^{\prime} \in \mathcal{R}\right\}, \quad(h, r, t) \in \mathcal{G} .
\end{aligned}
\end{equation}
This procedure replaces one entity or relation in a positive triple with another randomly chosen entity or relation from the entire set, and any resulting triples already in $\mathcal{G}$ are discarded to ensure the correctness of the negative samples.

\subsection{Triple Confidence Learning with Entity Attributes} \label{sec:triple_confidence_learning}
To enhance the robustness of our model, we learn a confidence score $C(h, r, t)$ for each triple, which quantifies its correctness by leveraging both graph structure and entity attributes. Real-world KGs are often enriched with entity attributes that can provide valuable semantic information. These attributes are highly correlated with the relations in the KG and can help identify potential errors. We observe three key relationships:
\begin{itemize}
    \item In the triple, relations can be viewed as translations acting on the low-dimensional embeddings of the entities. The better a triple aligns with the translation assumption, i.e., $\mathbf{h} + \mathbf{r} \simeq \mathbf{t}$, the more likely it is to be correct.
    \item In the graph, connected triples that share the same entity are typically semantically relevant. Intuitively, a KG can be viewed as a social group, where each triple is an individual. The acknowledgment from neighboring triples reflects how well a triple fits within the broader structure.
    \item Entity attributes are often closely related to the graph structure. Entities with relevant semantic attributes are usually connected by specific relations. A mismatch between an entity's attributes and its related triples suggests a higher likelihood of error.
\end{itemize}

To incorporate these insights, we propose a multi-view learning framework that models the structural information of the KG and its attributes. We use two triple-level hypergraphs—\emph{relational hypergraph} and \emph{attribute hypergraph}—to capture, respectively, the topological structure of the KG and the semantic information embedded in entity attributes. These hypergraphs are processed in parallel through a unified contrastive learning framework, enabling us to measure the confidence of each triple by considering three types of anomaly signals: self-contradiction in the relational structure, global consistency across triples, and attribute-structure dependency.

\paragraph{Multi-view Construction.}
Traditional KG embedding models typically only consider entities and relations within triples, overlooking the global correlations among triples. To capture these correlations, we introduce a novel approach that transforms the KG into hyper-views at the triple level, treating each relational triple as a node in the graph. Specifically, we construct a \emph{relational hypergraph} by linking relational triples that share the same entity in the original KG, ensuring that the transformation preserves the potential for error detection within individual triples. 

\begin{definition}\label{def:relational_graph}
	\textbf{Relational Hypergraph.} Given a knowledge graph $\mathcal{G}=\{ (h,r,t) | h,t \in \mathcal{E}, r \in \mathcal{R}\}$, the corresponding relational hypergraph is the triple-level graph $\mathcal{G}_r = (\widetilde{\mathcal{V}}, \widetilde{\mathcal{A}_{r}}, \widetilde{\mathcal{X}_{r}})$, where $\widetilde{\mathcal{V}}$ and $\widetilde{\mathcal{A}_{r}}$ are the sets of nodes and adjacency matrix, respectively. $\widetilde{\mathcal{X}_{r}} = \{ \Gamma(h,r,t) \ |\ (h,r,t) \in \mathcal{G} \}$ represents the feature matrix of $\widetilde{\mathcal{V}}$, where $\Gamma(\cdot)$ is a concatenation function. $\widetilde{\mathcal{A}_{r}}(v, u|u,v \in \widetilde{\mathcal{V}}) = 1$, if $u$ and $v$ share the same entity in the original KG, i.e., $\mathcal{G}$.
\end{definition}

Next, we build an \emph{attribute hypergraph} to capture the relationship between entities and their attributes. This hypergraph uses the attributes of the head and tail entities in a triple to reconstruct the semantics of the triple from the attribute perspective.

\begin{definition}\label{def:attribute_graph}
	\textbf{Attribute Hypergraph.} Given the same knowledge graph $\mathcal{G}=\{ (h,r,t) | h,t \in \mathcal{E}, r \in \mathcal{R}\}$, with entity attribute set  $\mathcal{A}=\{\left\{a_{h, 1}, a_{h, 2}, \ldots, a_{h,\left|A_{h}\right|}\}|h\in \mathcal{E}\right\}$ where $a_{h,i}$ is the $i^{th}$ attribute type for entity $h$, the attribute hypergraph can be denoted as $\mathcal{G}_{a}=(\mathcal{V}_{a},\mathcal{A}_{a}, \mathcal{X}_{a})$, where $\mathcal{V}_{a}$ and $\mathcal{A}_{a}$ are the set of nodes and adjacency matrix, respectively, which have the same structure as $\mathcal{G}_{r}$. $\mathcal{X}_{a}$ represents the feature matrices of each node $v_{i} \in \mathcal{V}_{a}$ learned from corresponding entity attributes.
\end{definition}

These two hypergraphs, \emph{relational hypergraph} and \emph{attribute hypergraph}, can be seen as complementary views of the same KG, providing richer information for error detection. The \emph{relational hypergraph} captures the global correlations between relational triples, while the \emph{attribute hypergraph} highlights the semantic meaning derived from entity attributes. By measuring the consistency between these two views, we can more accurately assess the trustworthiness of each triple in the original KG.

\paragraph{Learning from View I: Relational Hypergraph.} \label{sec:Relational_encoder}
Traditional KG embedding methods typically focus on the local relationships within individual triples, disregarding the broader context. To address this, we propose a new dual-view encoder that simultaneously learns both local and global context for relational triples.

\noindent{\bf Local View of Relational Structure Modeling.} 
The \emph{relational hypergraph} transforms triples into a structure that can lose some local relational details, such as the head-relation-tail structure inherent within triples. To preserve this local information, we initialize the embedding of entities and relations in the original KG, then use a BiLSTM layer to encode the local structure of each triple. The local representation $p_i$ for the $i^{th}$ triple $(h, r, t)$ is defined as:
\begin{equation}  
p_{i}=\mathcal{G}_{local}(h, r, t) = f_{concat}(f_{BiLSTM}\left(e_h, e_{r}, e_{t}\right)).
\label{equ:pi}
\end{equation}

\noindent{\bf Global View of Neighbor Information Aggregation.} 
Beyond the local structure, the global context of neighboring triples also contains valuable information for detecting anomalies. To model this, we employ a graph attention network (GAT) to selectively aggregate features from neighboring triples. Given an anchor triple $p_{i} \in \mathbb{R}^{d}$, its embedding is updated by attending over neighboring triples, $\left\{p_{1}, p_{2}, \ldots, p_{m}\right\}$, through a two-layer attention mechanism:
\begin{equation}  
att_{i j}=f_{att}\left(\mathbf{W} p_{i}, \mathbf{W} p_{j}\right)
\label{attention_coefficient}.
\end{equation}
The attention values are normalized using the softmax function:
\begin{equation}  
\alpha_{i j}=\frac{\exp \left(att_{i j}\right)}{\sum_{k=1}^{m} \exp \left(att_{i k}\right)}.
\label{eq:eq3}
\end{equation}
Finally, the updated embedding is computed as:
\begin{equation}  
x_{i}=\sigma\left(\sum_{j=1}^{m} \alpha_{i j} \mathbf{W} p_{j}\right).
\label{eq:xi}
\end{equation}

\paragraph{Learning from View II: Attribute Hypergraph.} \label{sec:Attribute_encoder}
In addition to relational structure, entity attributes contain valuable semantic information that enhances KG embedding quality. To effectively incorporate this information, we design a relation-specific encoder, $g_{attr}$, which learns attribute-based representations from the \emph{attribute hypergraph}.

\noindent{\bf Attribute Hypergraph Embedding.} 
Entity attributes are diverse and unevenly distributed across different entities. An entity can have distinct attribute sets depending on its role in different triples. For instance, a person entity might have attributes related to \emph{research area} when playing the role of a scientist and \emph{pen name} when considered as a writer. To handle these variations, we introduce a relation-specific mechanism that selects relevant attributes based on the relation in the triple.

Given a triple $(h, r, t)$, the attributes of the head and tail entities are aggregated using relation-specific attention mechanisms. The attribute-based entity representations $\hat{e}_{h}$ and $\hat{e}_{t}$ are computed as:
\begin{equation}
    a t t_{h, i} =f_{att} \left(f_{emb} \left(e_r\right), f_{emb} \left(a_{h, i}\right)\right),
\end{equation}
\begin{equation}
    \alpha_{h, i} =\frac{\exp \left(a t t_{h, i} \right)}{\sum_{j=1}^{\left|A_{h}\right|} \exp \left(a t t_{h, j} \right)},
\end{equation}
and the final aggregated representation for the entity $h$ is:
\begin{equation}
\hat{e}_{h} =\sum_{i=1}^{\left|A_{h}\right|} \alpha_{h, i}  * f_{emb} \left(a_{h, i}\right).
\end{equation}

\paragraph{Model Learning.} \label{sec:constrstive_learning}
To learn robust features from both \emph{relational hypergraph} and \emph{attribute hypergraph}, we utilize a contrastive loss function that maximizes the mutual information between the embeddings derived from these two views.

\noindent{\bf Contrasting Between Structure and Attribute Views.} 
The \emph{relational hypergraph} models the graph structure, while the \emph{attribute hypergraph} captures the entity attributes. These two views are complementary, and we apply the normalized temperature-scaled cross entropy loss to maximize their mutual information:
\begin{equation} 
\mathcal{L}_{\text{sa}}(x_i, z_i)=-\log \frac{\exp \left(\operatorname{sim}\left(x_i, z_i\right) / \tau\right)}{\sum_{j\in{\{1,2, \dots , N\} \setminus \{i\} }} \exp \left(\operatorname{sim}\left(x_i, z_j\right) / \tau\right)},
\label{con_loss}
\end{equation}

\noindent{\bf Contrasting Between Local and Global Views.} 
We also contrast the local and global views of each triple, as the relational structure and its global context provide complementary information. The contrastive loss for this objective is defined as:
\begin{equation} 
\mathcal{L}_{\text{lg}}({p}_i, x_i)=-\log \frac{\exp \left(\operatorname{sim}\left({p}_i, x_i\right) / \tau\right)}{\sum_{j\in{\{1,2, \dots , N\} \setminus \{i\} }} \exp \left(\operatorname{sim}\left({p}_i, x_j\right) / \tau\right)}.
\label{con_loss1}
\end{equation}
Finally, combining these two losses, we obtain the overall contrastive loss:
\begin{equation}
\mathcal{L}_{con}=\frac{1}{2 N} \sum_{i=1}^{N}\left(\mathcal{L}_{sa}\left(x_i, z_i\right)+\mathcal{L}_{lg}\left(p_i, x_i\right)\right).
\label{eq:loss_con}
\end{equation}

\paragraph{Triple Confidence Estimation.}
We calculate the triple confidence score based on three key anomaly signals: self-contradiction within the relational structure, global consistency across triples, and attribute-structure dependency. These signals help us assess the reliability of each triple.

\noindent{\bf Self-contradictory Measurement.} In the relational structure, the closer a triple fits the translation assumption $\mathbf{h} + \mathbf{r} \simeq \mathbf{t}$, the more reliable it is. We define the local triple confidence $LT(h, r, t)$ based on the Euclidean distance between the entities:
 \begin{equation}
     LT(h, r, t)=\frac{1}{1+e^{ - \left\|e_{h}+e_{r}-e_{t}\right\|_{2}}}.
 \end{equation}

\noindent{\bf Global Acknowledgment Estimation.} 
The degree of acknowledgment from neighboring triples reflects whether a target triple is valid. We define the global triple confidence $GT(h, r, t)$ as the cosine similarity between the local and global representations of the triple:
\begin{equation}
     GT(h, r, t)=sim(p_i,x_i),
\end{equation}

\noindent{\bf Structure-attribute Dependency Estimation.} 
The consistency between the relational and attribute hypergraph views is an important signal for determining the correctness of a triple. We compute the attribute-structure confidence $AT(h, r, t)$ as:
\begin{equation}
     AT(h, r, t)=sim(x_i,z_i),
\end{equation}

\noindent{\bf Final Triple Confidence.} 
We combine these three signals to calculate the final confidence score for each triple:
\begin{equation}  
\begin{aligned}
C(h, r, t)= \sigma(LT(h,r,t)  + \lambda_1 \cdot  GT(h,r,t) \\
+  \lambda_2 \cdot AT(h,r,t)),
\end{aligned}
\label{confidence_score}
\end{equation}

\subsection{Joint Adaptive Training Scheme}
To integrate the learned confidence information into the KG embedding process, we define a comprehensive objective function:
\begin{equation}
\mathcal{L}=\mathcal{L}_{con} + \beta \mathcal{L}_{emb}.
\label{eq:loss_all}
\end{equation}
This adaptive training scheme ensures that the KG representation learning and confidence learning processes mutually reinforce each other, allowing the model to progressively improve its ability to filter out erroneous triples while learning accurate KG embeddings.

\begin{table*}[t]
\centering%\smalll
% \vspace{2pt}
\caption{ Error detection results of Precision@K and Recall@K based on the three datasets with anomaly ratio $= 5\%$.}
 \resizebox{1\textwidth}{!}{
\vspace{0.2cm}
%\resizebox{490pt}{100pt}{
\begin{tabular}{llccccccccccccccc}
\toprule
& &\multicolumn{5}{c}{FB15K-237} &\multicolumn{5}{c}{DB15K} &\multicolumn{5}{c}{YAGO15K} \cr \cmidrule(lr){3-7} \cmidrule(lr){8-12} \cmidrule(lr){13-17} 
& &$K$=$1\%$  &$K$=$2$\%  &$K$=$3$\%  &$K$=$4$\%  &$K$=$5$\%$^a$ & $K$=$1\%$  &$K$=$2$\%  &$K$=$3$\%  &$K$=$4$\%  &$K$=$5$\%$^a$ & $K$=$1\%$  &$K$=$2$\%  &$K$=$3$\%  &$K$=$4$\%  &$K$=$5$\%$^a$ \\ \cmidrule(lr){3-7} \cmidrule(lr){8-12} \cmidrule(lr){13-17} 
% \multirow{9}{*}{$Precision@K$}
\parbox[t]{5mm}{\multirow{12}{*}{\rotatebox[origin=c]{90}{$Precision@K$}}}
& TransE
&0.756 &0.674 &0.605 &0.546 &0.488 
&0.671 &0.566 &0.535 &0.472 &0.443
&0.534 &0.455 &0.362 &0.319 &0.279 \\ 
& ComplEx
 &0.718 &0.651 &0.590 &0.534 &0.485 
 &0.679&0.612&0.532&0.469&0.424&0.503&0.426&0.369&0.310&0.273  \\ 
& DistMult
 &0.709 &0.646 &0.582 &0.529 &0.483
 &0.638&0.587&0.523&0.499&0.445&0.513&0.423&0.347&0.347&0.302 \\
& SimplE 
&0.744 &0.667 &0.611 &0.556 &0.515 
&0.709 &0.616 &0.554 &0.504 &0.455 
&0.560 &0.459 &0.373 &0.319 &0.281 \\
&TuckER
&0.742 &0.680 &0.614 &0.552 &0.514 
&0.711 &0.630 &0.550 &0.509 &0.452 
&0.549 &0.454 &0.375 &0.331 &0.276 \\ 
& EARL
& 0.762& 0.692& 0.639& 0.582& 0.531
& 0.729& 0.652& 0.573& 0.530 & 0.483 
& 0.578& 0.487& 0.394 & 0.338& 0.303 \\
%  &0.764 &0.692 &0.582 &0.529 &0.483
%  &0.574 &0.451 &0.390 &0.349 &0.322
%  &0.630 &0.553 &0.493 &0.446 &0.408 \\
% & TuckER
%  &0.709 &0.646 &0.582 &0.529 &0.483
%  &0.574 &0.451 &0.390 &0.349 &0.322
%  &0.630 &0.553 &0.493 &0.446 &0.408 \\
% & EARL
%  &0.709 &0.646 &0.582 &0.529 &0.483
%  &0.574 &0.451 &0.390 &0.349 &0.322
%  &0.630 &0.553 &0.493 &0.446 &0.408 \\
& TTMF
 &0.815 &0.767 &0.713 &0.612 &0.579
 &0.744&0.685&0.623&0.557&0.510&0.701&0.569&0.454&0.413&0.346 \\
&CAGED
&0.852 &0.796 &0.735 &0.665 &0.595
&0.802& 0.722& 0.658& 0.596& \underline{0.554}& 
0.724& 0.599& 0.489& 0.429& 0.373\\
& CKRL
 &0.789 &0.736 &0.684 &0.630 &0.574
 &0.787&\underline{0.723}&0.634&0.599&0.526&0.662&0.531&0.438&0.382&0.327 \\
& NoiGAN
 & 0.837 & 0.788 & 0.727 & 0.649 & 0.585
 &\underline{0.823} &0.718 &\underline{0.676} &\underline{0.606} &0.553
 &0.737&0.592&0.477&0.403&0.379 \\
& CrossVal 
 &\underline{0.874} &\underline{0.814} &\underline{0.742} &\underline{0.667} &\underline{0.596} 
 &0.819 &0.707 &0.645 &0.572 &0.548
 &\underline{0.754} &\underline{0.647} &\underline{0.562} &\underline{0.500} &\underline{0.424} \\ 
& AEKE 
&\textbf{0.892 } &\textbf{0.822 } &\textbf{0.753 } &\textbf{0.691 } &\textbf{0.614 } &\textbf{0.853} &\textbf{0.756} &\textbf{0.705} &\textbf{0.625} &\textbf{0.582} &\textbf{0.778} &\textbf{0.676} &\textbf{0.590} &\textbf{0.519} &\textbf{0.440} \\ \cmidrule(lr){3-7} \cmidrule(lr){8-12} \cmidrule(lr){13-17} 
%\multicolumn{16}{c}{Recall@K} \\ \cmidrule(lr){2-6} \cmidrule(lr){7-11} \cmidrule(lr){12-16} 
%Dataset &\multicolumn{5}{c}{FB15K-237} &\multicolumn{5}{c}{DB15K} &\multicolumn{5}{c}{NELL-995} \cr \cmidrule(lr){2-6} \cmidrule(lr){7-11} \cmidrule(lr){12-16} 
% &$K$=$1\%$  &$K$=$2$\%  &$K$=$3$\%  &$K$=$4$\%  &$K$=$5$\% & $K$=$1\%$  &$K$=$2$\%  &$K$=$3$\%  &$K$=$4$\%  &$K$=$5$\% & $K$=$1\%$  &$K$=$2$\%  &$K$=$3$\%  &$K$=$4$\%  &$K$=$5$\% \\ \cmidrule(lr){2-6} \cmidrule(lr){7-11} \cmidrule(lr){12-16} 
% \multirow{9}{*}{$Recall@K$}
\parbox[t]{5mm}{\multirow{12}{*}{\rotatebox[origin=c]{90}{$Recall@K$}}}
&TransE%~\cite{bordes2013translating}
&0.151&0.269&0.363&0.437&0.488&0.134&0.226&0.321&0.377&0.443&0.106&0.182&0.217&0.255&0.279  \\
&ComplEx%~\cite{trouillon2016complex} 
&0.144&0.260&0.354&0.427&0.485&0.135&0.244&0.319&0.375&0.424&0.100&0.170&0.221&0.248&0.273 \\
&DistMult%~\cite{yang2015embedding} 
&0.142&0.258&0.349&0.423&0.483&0.127&0.234&0.313&0.399&0.445&0.102&0.169&0.208&0.277&0.302 \\
&SimplE
&0.149 &0.267 &0.366 &0.445 &0.515 
&0.142 &0.246 &0.332 &0.403 &0.455 
&0.112 &0.184 &0.224 &0.255 &0.281 \\
&TuckER
&0.148 &0.272  &0.368 &0.442 &0.514
&0.142 &0.252  &0.330 &0.407 &0.452 
&0.110 &0.182 &0.225  &0.265 &0.276 \\ 
& EARL
&0.152 &0.277 &0.383 &0.466 &0.531  
&0.146 &0.261 &0.344 &0.424  &0.483 
&0.116 &0.195 &0.236 &0.270 &0.303\\
% &R-GCN%~\cite{yang2015embedding} 
% &0.141 &0.258 &0.349 &0.423 &0.483 
% &0.114 &0.180 &0.234 &0.279 &0.322  
% &0.126 &0.221 &0.295 &0.357 &0.408 \\
% &KGAT%~\cite{yang2015embedding} 
% &0.141 &0.258 &0.349 &0.423 &0.483 
% &0.114 &0.180 &0.234 &0.279 &0.322  
% &0.126 &0.221 &0.295 &0.357 &0.408 \\
&TTMF%~\cite{jia2019triple} 
&0.163&0.306&0.427&0.489&0.579&0.148&0.274&0.373&0.445&0.510&0.140&0.227&0.272&0.330&0.346 \\
&CAGED
&0.171 &0.318 &0.441 &0.532 &0.595
&0.160 &0.288 &0.395 &0.477 &\underline{0.554}
&0.145 &0.240 &0.293 &0.343 &0.374\\
&CKRL%~\cite{xie2018does}  
&0.158&0.294&0.410&0.504&0.574&0.157&\underline{0.289}&0.380&0.479&0.526&0.132&0.212&0.262&0.305&0.327\\ 
&NoiGAN%~\cite{BelthZVK20}
&0.167&0.315&0.436&0.519&0.585
% &0.164&0.287&0.405&0.484&0.553
&\underline{0.164} &0.287 &\underline{0.405} &\underline{0.484} &0.553
&0.147&0.236&0.286&0.322&0.379 \\
% &0.163 &0.307 &0.428 &0.490 &0.579 
% &0.154 &0.251 &0.309 &0.355 &0.396  
% &0.161 &0.276 &0.361 &0.428 &0.481 \\
&CrossVal 
% &0.175&0.325&0.445&0.533&0.596
&\underline{0.175} &\underline{0.325} &\underline{0.445} &\underline{0.533} &\underline{0.596}
&0.163&0.282&0.387&0.457&0.548
&\underline{0.150} &\underline{0.258} &\underline{0.337} &\underline{0.400} &\underline{0.424} \\
&AEKE 
&\textbf{0.178} &\textbf{0.328} &\textbf{0.452} &\textbf{0.552} &\textbf{0.614} &\textbf{0.170} &\textbf{0.302} &\textbf{0.423} &\textbf{0.500} &\textbf{0.582} &\textbf{0.155} &\textbf{0.270} &\textbf{0.354} &\textbf{0.415} &\textbf{0.440} \\\bottomrule
\end{tabular}}
% \\
% \tabnote{$^{\rm a}$This footnote shows what footnote symbols to use.}
% \footnotesize{$^a$ Please note that according to Eqs.\eqref{eq:precision}-\eqref{eq:recall}, when $K$ equals  $5\%$, $Precision@K$ and $Recall@K$ are equal.}
% }
\label{table:results}
\end{table*}
\section{Experiments}

In this section, we present comprehensive experiments to evaluate the performance of the proposed AEKE framework across various real-world KGs. The primary goals are to answer the following research questions:
\begin{itemize}
    \item \textbf{Q1}: How effectively does AEKE identify noisy triples in knowledge graphs?
    \item \textbf{Q2}: How do the KG embeddings learned by AEKE compare to the state-of-the-art KG embedding models in terms of quality?
    \item \textbf{Q3}: What contributions do the individual components of AEKE make to its overall performance?
    % \item \textit{\textbf{Q4}}~(\cref{sec:parameters}): How do different hyperparameters affect AEKE’s performance?
\end{itemize}

\subsection{Datasets}
We evaluate AEKE on three real-world benchmark datasets: FB15K-237, YAGO15K, and DB15K. These datasets are known for their high reliability due to extensive human curation. Following prior works~\cite{xie2018does,cheng2019noigan}, we introduce 5\% and 15\% noisy triples into each dataset to simulate real-world errors. The noisy triples are generated by replacing the head or tail entity in a given triple $(h, r, t)$ with an entity that has appeared in the same position with the same relation in the dataset, producing a more challenging and realistic form of error.

\noindent{\bf FB15K-237} is a widely-used subset of Freebase containing 114 relations and 10,054 entities, offering a large-scale knowledge base with over 1 billion triples. FB15K-237 improves upon the FB15K dataset by addressing issues with inverse relations and handling symmetric, asymmetrical, and combinatorial relationships, as well as attributes.

\noindent{\bf DB15K} is a subset of Wikidata, designed to address the weaknesses of Wikipedia-based datasets. It excludes inverse relations to prevent test leakage, similar to the process used for FB15K-237.

\noindent{\bf YAGO15K} is built by augmenting WordNet with over 1 million entities and aligning it with Freebase to form a knowledge graph.

\subsection{Capability of AEKE in Distinguishing KG Errors (Q1)} \label{sec:error_detection}
In this section, we evaluate AEKE’s ability to distinguish errors within KGs. Specifically, we rank all triples in the target KG by their confidence scores in ascending order. The top-ranked triples are considered potential errors. We conduct experiments on FB15K-237, DB15K, and YAGO15K with 5\% and 15\% noisy triples. Below, we discuss the experiment setup, baseline models, and evaluation metrics.

\paragraph{Baselines.}
We compare AEKE to several baseline models from three categories: error-aware embedding methods, error detection alternatives, and KG embedding methods. The baselines are as follows:

\noindent{\bf TransE}: Assumes entities and relations are embedded in the same space, where the tail entity can be predicted by the head entity and relation.

\noindent{\bf DistMult}: A bilinear model that calculates the confidence of potential semantics for entities and relations.

\noindent{\bf ComplEx}: An extension of DistMult that uses complex numbers to model both symmetric and asymmetric relations.

\noindent{\bf SimplE}: An interpretable embedding model that employs weight tying to integrate specific background knowledge.

\noindent{\bf TuckER}: A linear model for link prediction based on Tucker decomposition of a binary tensor containing known facts.

\noindent{\bf EARL}: Focuses on learning embeddings for reserved entities, considering connected relations and neighbors.

\noindent{\bf TTMF}: Uses semantic information to calculate the trustworthiness of triples, distinguishing between normal and anomalous triples.

\noindent{\bf CAGED}: Combines KG embedding and contrastive learning to assess the trustworthiness of each triple based on multi-view consistency.

\noindent{\bf NoiGAN}: Uses Generative Adversarial Networks (GANs) for noise-aware knowledge graph embeddings.

\noindent{\bf CKRL}: A confidence-aware representation learning method that assigns confidence scores to triples to identify potential noise.

\noindent{\bf CrossVal}: Uses external human-curated knowledge graphs as auxiliary information to improve error detection within the target KG.

\paragraph{Evaluation Protocol.}
We implement all models using PyTorch, and the experiments are conducted on an Nvidia 3090 GPU. The Adam optimizer is used with a batch size of 256, a learning rate of 0.01, and Xavier initialization for model parameters. The embedding size is set to 100 for all models.

We use ranking measures to evaluate the performance of each model. A triple's confidence score is used to rank it, with lower scores indicating higher likelihood of errors. The evaluation metrics are:

\noindent{\bf Precision@K}: Measures the percentage of true anomalies in the top K rankings.
\begin{equation}
    \text{Precision@K} = \frac{\mid \text{Errors Found in Top K} \mid}{K}.
    \label{eq:precision}
\end{equation}

\noindent{\bf Recall@K}: Measures the percentage of true anomalies found within the total errors in the KG.
\begin{equation} 
    \text{Recall@K} = \frac{\mid \text{Errors Found in Top K} \mid}{\mid \text{Total Errors in KG} \mid}.
    \label{eq:recall}
\end{equation}

\paragraph{Experimental Results.}
The results for Q1 are summarized in Table~\ref{table:results}. We observe the following:

\noindent\textbf{Obs. 1.} AEKE outperforms both embedding methods and state-of-the-art error detection baselines. For example, with a 5\% anomaly ratio and $K=5\%$, AEKE achieves a 1.8\% improvement over the second-best method in terms of recall and precision.

\noindent\textbf{Obs. 2.} While KG embedding methods like TransE, ComplEx, and DistMult show decent performance, they lag behind tailored error detection methods. This is because these embedding methods do not account for errors in the KG, resulting in less discriminative representations for normal and noisy triples.

\noindent\textbf{Obs. 3.} Incorporating auxiliary human-curated information, as seen in CrossVal, improves error detection. However, AEKE surpasses it by leveraging the relational structure within triples, the global context across triples, and the semantics from entity attributes, making AEKE more effective at error detection.

\subsection{Quality of KG Embeddings Learned by AEKE (Q2)} \label{sec:kg_completion}
Next, we evaluate the quality of the KG embeddings learned by AEKE using the KG completion task. In this task, the goal is to predict missing entities in incomplete triples. We compare AEKE with strong baselines to assess the quality of its embeddings.

\begin{table*}[t]
% \small
\centering
%%\vspace{2pt}
\caption{ Results of knowledge graph completion.}
 \resizebox{1\textwidth}{!}{
\begin{tabular}{ccccccccccccccccccc}
\toprule
% \multicolumn{16}{c}{Precision@K}\\ \midrule

\multicolumn{2}{c}{Dataset} &\multicolumn{4}{c}{FB15K-237} &\multicolumn{4}{c}{DB15K} &\multicolumn{4}{c}{YAGO15K}\cr \cmidrule(lr){3-6} \cmidrule(lr){7-10} \cmidrule(lr){11-14} 
\multicolumn{2}{c}{Metrics} &MRR  &Hit@1  &Hit@3  &Hit@10   &MRR  &Hit@1  &Hit@3  &Hit@10 &MRR  &Hit@1  &Hit@3  &Hit@10\\ \midrule
\multirow{7}{*}{0\%} &
TransE 
&0.302 &0.211 &0.344 &0.468 
&0.420 &0.385 &0.431 &0.509
&0.311 &0.199 &0.369 &0.523\\
&SimplE 
&0.288 &0.202 &0.314 &0.455 &0.437 &0.392 &0.445 &0.505 &0.296 &0.188 &0.344 &0.487
% &0.252 &0.167 &0.275 &0.423 
% &0.441 &0.395 &0.448 &0.502
% &0.292 &0.204 &0.327 &0.475
\\
&EARL  
&0.308 &0.209 &0.338 &0.474 &0.447 &0.389 &0.459 &0.508 &0.317 &0.199 &0.372 &0.509\\
% &0.255 &0.169 &0.284 &0.437 
% &0.448 &0.401 &0.452 &0.515
% &0.263 &0.189 &0.285 &0.432\\
&RGCN 
&0.363 &0.306 &0.387 &0.512
&0.474 &0.410 &0.487 &0.576
&0.381 &0.269 &0.424 &0.588\\
&RGHAT 
&0.422 &0.362 &\underline{0.446} &\textbf{0.535}
&\underline{0.483} &\underline{0.422} &\underline{0.499} &\underline{0.588}
&0.412 &0.295 &\underline{0.467} &\underline{0.618}\\
&NoiGAN 
&\underline{0.424} &\underline{0.371} &0.445 &0.526 
&0.480 &0.421 &0.483 &0.581
&\underline{0.413} &\underline{0.310} &0.465 &0.605\\
&CKRL
&0.410&0.362&0.438&0.536&0.482&0.416&0.490&0.586&0.408&0.298&0.461&0.610\\
&Ours
&\textbf{0.427} & \textbf{0.375} & \textbf{0.453} & \underline{0.534}
&\textbf{0.486} &\textbf{0.426} &\textbf{0.503} & \textbf{0.587}
&\textbf{0.418} &\textbf{0.316} &\textbf{0.474} &\textbf{0.622}\\ \midrule
\multirow{7}{*}{5\%} &
TransE 
&0.294 &0.201 &0.335 &0.465 
&0.407 &0.358 &0.417 &0.481
&0.288 &0.161 &0.345 &0.493\\
&SimplE 
&0.266 &0.178 &0.281 &0.416 &0.407 &0.354 &0.412 &0.466 &0.277 &0.166 &0.313 &0.455\\
% &0.241 &0.155 &0.265 &0.419 
% &0.425 &0.376 &0.433 &0.482
% &0.269 &0.177 &0.333 &0.454\\
&EARL 
&0.284 &0.194 &0.308 &0.437 &0.407 &0.366 &0.416 &0.466 &0.283 &0.180 &0.348 &0.468\\
% &0.247 &0.162 &0.277 &0.428 
% &0.431 &0.395 &0.450 &0.496
% &0.246 &0.151 &0.275 &0.423\\
&RGCN 
&0.348 &0.261 &0.375 &0.493 
&0.456 &0.408 &0.472 &0.551
&0.365 &0.250 &0.397 &0.562\\
&RGHAT 
&0.401 &0.342 &\underline{0.442} &0.507
&0.460 &0.411 &0.472 &0.565
&0.392 &0.275 &0.442 &0.580\\
&NoiGAN
&\underline{0.418} &\underline{0.360} &0.440 &\underline{0.515}
&\underline{0.474} &\underline{0.415} &\underline{0.479} &\underline{0.578}
&\underline{0.404} &\underline{0.305} &\underline{0.457} &\underline{0.598}\\
&CKRL
&0.400&0.340&0.426&0.512&0.471&0.410&0.483&0.572&0.397&0.292&0.449&0.594\\
% &TTMF
% &0.398&0.336&0.421&0.506&0.45&0.401&0.46&0.552&0.376&0.258&0.432&0.574\\
% &AEKE(hard)
% &0.411&0.345&0.433&0.518&0.461&0.412&0.474&0.566&0.384&0.267&0.443&0.590\\
&Ours
&\textbf{0.424}  &\textbf{0.369}  &\textbf{0.447}  &\textbf{0.519}
&\textbf{0.482}  &\textbf{0.424}  &\textbf{0.499}  &\textbf{0.583}
&\textbf{0.412}  &\textbf{0.308}  &\textbf{0.465}  &\textbf{0.612}\\ \midrule
\multirow{7}{*}{15\%} &
TransE 
&0.267 &0.185 &0.309 &0.451 
&0.386 &0.340 &0.401 &0.469
&0.276 &0.149 &0.334 &0.473\\
&SimplE 
&0.242 &0.154 &0.241 &0.361 &0.368 &0.314 &0.358 &0.416 &0.231 &0.134 &0.288 &0.419\\
&EARL 
&0.259 &0.169 &0.266 &0.381 &0.368 &0.326 &0.362 &0.416 &0.237 &0.147 &0.32 &0.431\\
&RGCN 
&0.325 &0.237 &0.356 &0.476
&0.437 &0.398 &0.455 &0.534
&0.343 &0.238 &0.375 &0.549\\
&RGHAT 
&0.395 &0.332 &0.416 &0.498 
&0.442 &0.398 &0.454 &0.540
&0.371 &0.258 &0.427 &0.565\\
&NoiGAN 
&\underline{0.409} &\underline{0.348} &\underline{0.427} &\underline{0.506} 
&\underline{0.463} &\underline{0.401} &0.462 &\underline{0.566}
&\underline{0.385} &\underline{0.295} &\underline{0.448} &\underline{0.589}\\
&CKRL
&0.396&0.334&0.422&0.501&0.458&0.400&\underline{0.467}&0.559&0.375&0.275&0.437&0.579\\
&Ours
&\textbf{0.417} & \textbf{0.363} & \textbf{0.440} & \textbf{0.512}
&\textbf{0.475} &\textbf{0.411} &\textbf{0.492} & \textbf{0.579}
&\textbf{0.401} &\textbf{0.299} &\textbf{0.451} & \textbf{0.600}\\
\bottomrule 
\end{tabular}}
\label{tab:performance_AEKE}
\end{table*}

\paragraph{Baseline Methods.}
We compare AEKE against the following baselines:
\begin{itemize}
    \item \textit{Embedding-based}: \textbf{TransE}, \textbf{SimplE}, \textbf{EARL}
    \item \textit{State-of-the-art completion models}: \textbf{RGCN}, \textbf{RGHAT}
    \item \textit{Error-aware embedding methods}: \textbf{NoiGAN}, \textbf{CKRL}
\end{itemize}

\paragraph{Evaluation Protocol}
We focus on entity prediction, where the task is to predict the head or tail entity of a given triple $(?, r, t)$ or $(h, r, ?)$, respectively. We use the Mean Rank (MRR) and Hits@K as evaluation metrics.

\paragraph{Experimental Results}
The results on FB15K-237, DB15K, and YAGO15K with 5\% and 15\% noisy triples are presented in Table~\ref{tab:performance_AEKE}. Key observations include:

\noindent\textbf{Obs. 1.} AEKE consistently outperforms embedding-based models and other error-aware methods, demonstrating the effectiveness of the learned KG embeddings.

\noindent\textbf{Obs. 2.} Error-aware embedding methods like NoiGAN and CKRL perform better than traditional embedding methods, but AEKE shows superior performance due to its ability to model both relational structure and entity attributes.

\noindent\textbf{Obs. 3.} As the anomaly rate increases, AEKE's performance gap over other methods grows. For example, with a 5\% anomaly ratio, AEKE shows 0.3\%, 0.6\%, and 0.5\% improvements on FB15K-237, DB15K, and YAGO15K, respectively. At a 15\% anomaly ratio, the improvements are more significant: 0.8\%, 1.2\%, and 1.6\%.

\begin{figure*}[!t]
\centering
\begin{center}
	\includegraphics[scale=0.7]{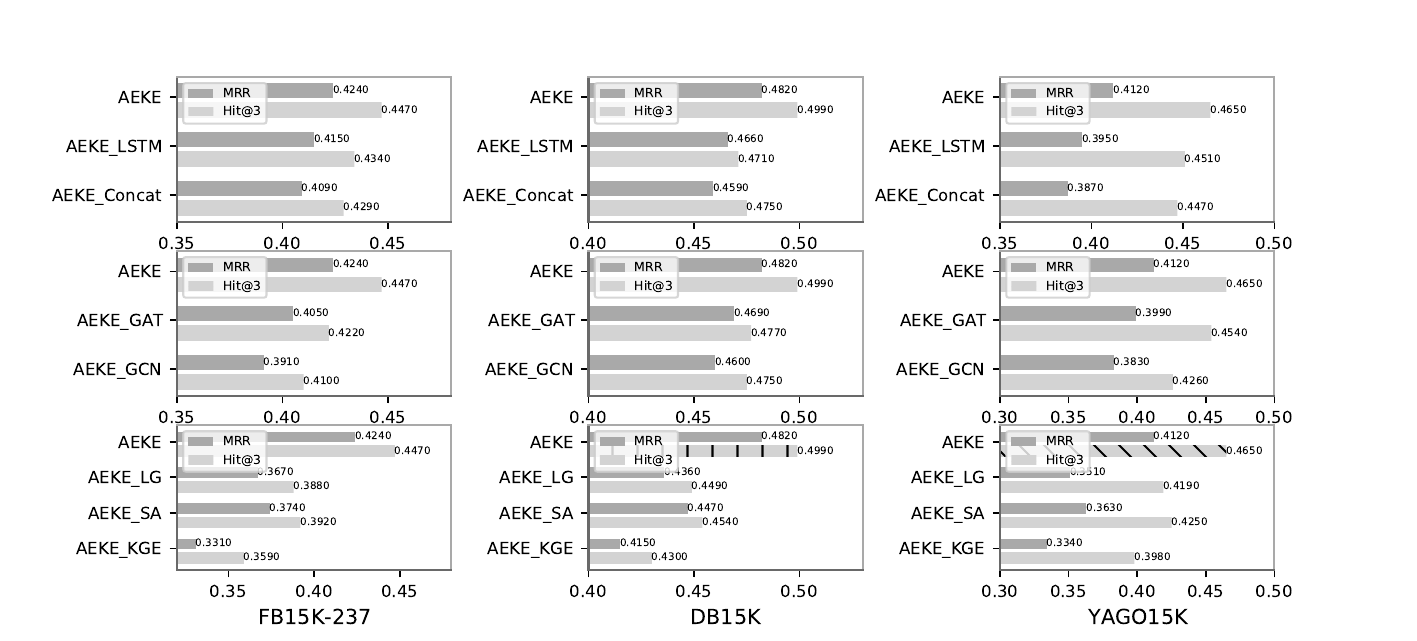}
\end{center}
	\caption{ KG completion results of AEKE variants based on the three datasets with anomaly ratio = $5\%$.}
	\label{fig:figure2}
	\vspace{-0.4cm}
\end{figure*}
\subsection{Ablation Study (Q3)} \label{sec:ablation_study}
We conduct an ablation study to assess the contribution of each component of AEKE. Four variants of AEKE are evaluated, and the effects of different modules are discussed.

\paragraph{Role of Attribute Hypergraph Encoder}
We evaluate the importance of the attribute encoder by comparing AEKE with and without the attribute view. Removing the attribute encoder leads to a significant drop in performance, confirming the importance of entity attributes in error-aware embedding.

\paragraph{Role of Relational Hypergraph Encoder}
To assess the effectiveness of capturing the relational structure, we compare AEKE with variants using different methods for encoding relational triples. AEKE outperforms these variants due to its tailored attention mechanism, which effectively filters out noisy triples.

\section{Summary}
Learning accurate and reliable embeddings for entities and relations in knowledge graphs (KGs) plays a critical role in enabling various downstream applications. While many existing approaches leverage the internal structure of KGs to train models for error detection, they are limited by the fact that the topological information present in KGs alone is insufficient for validation tasks, particularly in real-world scenarios. In this work, we introduce a novel framework for KG representation learning, named AEKE, which integrates semantic data from entity attributes to automatically validate triples in KGs. We conceptualize the original KG, which lacks attribute information, as a \emph{relational hypergraph}, and construct an \emph{attribute hypergraph} using the supplementary entity attribute information. This attribute hypergraph serves as an alternative view of the target KG. The confidence score for each triple is computed by considering multiple factors: inconsistency within the triple itself, alignment between local and global graph structures, and the compatibility between structure and attributes across triple-level views. Experimental results show that AEKE outperforms current state-of-the-art error detection methods for KGs. 
Given that real-world KGs continuously evolve with new data, extending AEKE to handle temporal KG representations would be a highly valuable direction for future work. Furthermore, we aim to explore the use of AEKE's error-aware KG representation learning in practical downstream tasks such as question answering and recommender systems.

\chapter{Logical Reasoning for Inductive Knowledge Graph Completion}
\section{Introduction}

Knowledge graphs (KGs) are structured collections of real-world information, typically represented as triples $(h, r, t)$, where $h$ is the head entity, $r$ is the relation, and $t$ is the tail entity~\cite{Bollacker-etal07Freebase,Lehmann-etal15DBpedia,Mahdisoltani-etal15YAGO,li2022constructing,dong2023hierarchy}. These graphs provide a way to model complex relationships between entities across various domains. However, KGs are often incomplete~\cite{zhang2022contrastive,dong2023active,li2022kcube,li2022constructing}, and manually curating facts is expensive, leading to the need for automated knowledge graph completion (KGC) methods~\cite{zhang2023integrating,ji2021survey}.

Currently, embedding-based techniques~\cite{bordes2013translating,lin2015learning,trouillon2016complex} dominate the field of KGC. These methods embed entities and relations into continuous vector spaces and predict missing relations between entities. While effective, these techniques generally operate under a transductive setting, where they assume all entities are present during both training and inference. This limitation arises when new entities are introduced, as the embeddings need to be retrained, which is impractical due to the constant growth and scale of real-world KGs.

Recent research has turned towards inductive KGC~\cite{wang2019logic,wang2021relational}, where models are designed to predict missing relations for unseen entities, which more accurately reflects real-world scenarios. In dynamic KGs, such as those used in e-commerce or biomedical domains, new entities, like users or products, are constantly added. To address this, several approaches have incorporated external resources, such as entity attributes, textual descriptions, and ontologies, to aid inductive KGC. However, these external data sources are often difficult to obtain, limiting their practical applicability. Alternatively, some methods have attempted to learn entity-independent rules for KGC, either through statistical~\cite{galarraga2013amie} or differentiable~\cite{yang2017differentiable} approaches, treating the problem as rule mining. However, rule-based techniques often struggle with scalability and generalization due to the domain-specific nature of rules across different KGs.

The rise of graph neural networks (GNNs) has led to the development of inductive KGC techniques based on message passing (MP)~\cite{schlichtkrull2018modeling,wang2021relational}. GraIL~\cite{teru2020inductive} is a notable early work that learns logical rules through reasoning over subgraphs surrounding a target triple. It aggregates structural information from neighboring entities within the subgraph to generate entity representations. While GraIL and subsequent methods~\cite{zhang2018link} have shown promising results, they face two main challenges: $(i)$ Data sparsity, especially when new entities lack sufficient connections to the existing graph, and $(ii)$ The limitation of traditional MP approaches that treat messages as entity-specific, which overlooks the importance of relation semantics~\cite{wang2021relational}. This is problematic for inductive KGC, where the reasoning process should be entity-independent.

\begin{figure*}[t]
	\centering
	\includegraphics[width=1\textwidth]{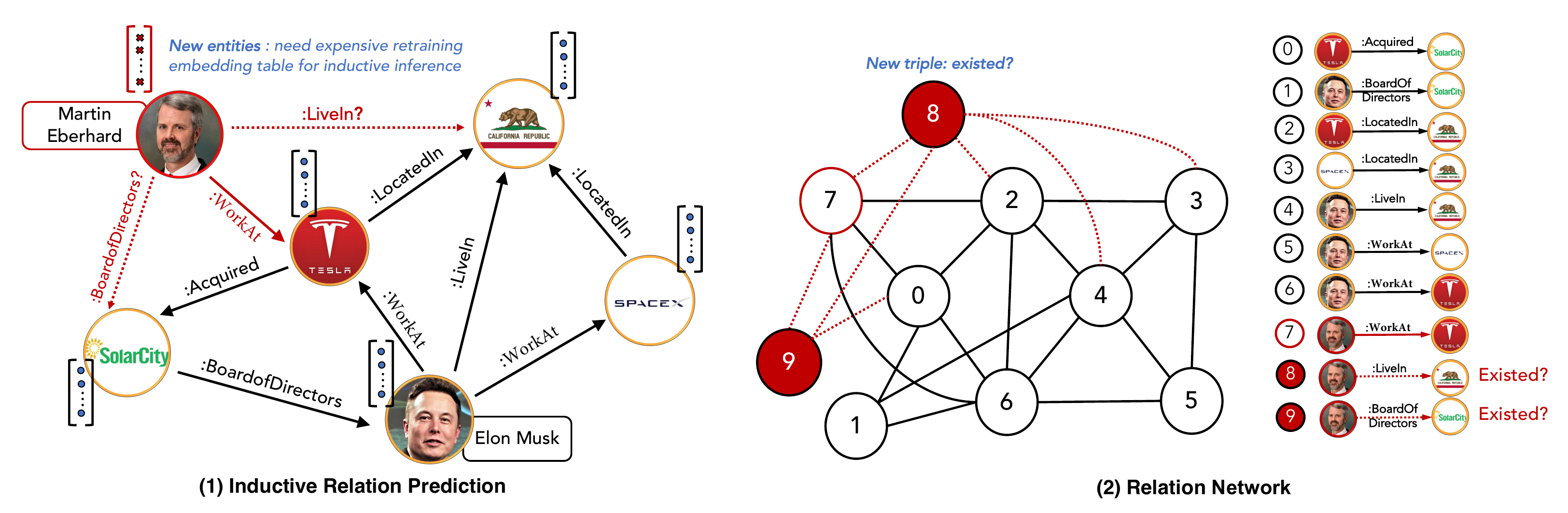}
	\vspace{-5mm}
	\caption{\textit{(i)} A toy example of inductive knowledge graph completion, i.e. predicting the relationship (``?'') for unseen entities, e.g. ``Martin Eberhard'', provided with a few links, e.g. (Martin, :WorkAt, TESLA); \textit{(ii)} Illustration of the corresponding \textit{relation network} for knowledge graph, which regards each triple as a relational node and thus could aggregate context information for inductive inference without expensive retraining look-up embedding tables as embedding-based paradigm.}\label{fig:illustration}
	% \vspace{-5mm}
\end{figure*}

In real-world KGs, the surrounding subgraph of a given triple contains the necessary logical evidence to infer the relationship between the entities. For instance, in Figure~\ref{fig:illustration}, learning the relation pattern between \textit{Elon Musk}, \textit{TESLA}, and \textit{California State} can help predict the relationship \textit{(Martin Eberhard, :LiveIn, California State)} based on the inferred pattern ``\textit{:LiveIn} $\simeq$ \textit{:WorkAt} $\land$ \textit{:LocatedIn}''. To address this, we introduce a novel framework called NORAN, which focuses on mining relation semantics for inductive KGC.

Our approach first redefines the KGC problem to bridge the gap between traditional embedding-based methods and inductive settings. Inspired by the insights gained, we propose the \emph{relation network}, a novel graph constructed by focusing on relations in the original KG. This relation network provides a hypergraph-like structure that represents the distribution and correlation of relations. By modeling this relation network through message-passing, we can capture entity-independent relation patterns, enabling more effective inductive KGC. We formally define the logical evidence extracted from the relation network and demonstrate its effectiveness in inductive KG completion.

Our main contributions are summarized as follows:
\begin{itemize}
   \item We introduce NORAN, a novel framework for inductive KG completion that learns latent relation semantics.
   \item We propose the \emph{relation network}, a hypergraph-based representation of the KG that centers on relations, and formally define inductive KGC as \textit{k-hop logic reasoning} over this hypergraph.
   \item We introduce the \textit{informax} training objective, which enables the effective capture of logical evidence from the relation network for KGC.
   \item We provide a theoretical analysis to identify the most effective message-passing strategies and offer guidelines for selecting models that enable inductive reasoning over the relation network.
   \item Extensive experiments on five real-world KG benchmarks demonstrate the superior performance of NORAN.
\end{itemize}

\section{Inductive Knowledge Graph Completion}
In this section, we introduce the NORAN framework, designed to uncover latent patterns among relational instances for inductive KG completion. As illustrated in Figure~\ref{fig:illustration}, NORAN is composed of three core components: $(i)$ Relation network construction, which redefines KG modeling with a focus on relations. $(ii)$ Relational message passing, a framework that implicitly extracts logical evidence independent of entities through the relation network. $(iii)$ Training scheme and KG inference, which includes a detailed theoretical analysis to identify effective message-passing strategies, offering insights into model selection for inductive reasoning over the relation network.

\subsection{Relation Network Construction}\label{sec:relation_network}
Conventional KG representation methods often represent a KG as a heterogeneous graph centered on entities, with relations serving as semantic edges. However, such approaches face limitations in capturing intricate relational semantics and fail to provide expressive representations for inductive KG completion. To bridge the gap between traditional KG representation learning and the requirements of inductive KGC, we propose a new perspective that focuses on relation correlations. This novel modeling approach introduces a hyper-level view of KGs based on relations, and we formally define inductive KG completion as \textit{k-hop logical reasoning} over an entity-independent network.

\begin{table*}[!ht]
	\caption{The general description framework for MP layers}
	% \vspace{-5mm}
	\centering
	\resizebox{1.0\textwidth}{!}{
		\begin{tabular}{cclcc}
			\toprule
			MP layer                                & Motivation          & Convolutional Matrix $\mathcal{C}$                                                & Feature Transformation $f$                       & Category of $\mathcal{C}$ \\
			\midrule
			GCN                & Spatial Convolution & $\mathcal{C} = \widetilde{\bm{D}}^{1/2}{(\bm{A}+\bm{I})}\widetilde{\bm{D}}^{1/2}$ & $f = \bm{W} \in \mathbb{R}^{d_k \times d_{k+1}}$ & Fixed                     \\
			GraphSAGE & Inductive Learning  & $\mathcal{C} = \widetilde{\bm{D}}^{-1}(\bm{A}+\bm{I})$                            & $f = \bm{W} \in \mathbb{R}^{d_k \times d_{k+1}}$ & Fixed                     \\
			GIN             & WL-Test             & $\mathcal{C} = \bm{A}+\bm{I}$                                                     & $f$ is a two-layer MLP                           & Fixed                     \\
			SGC           & Scalability         & $\mathcal{C} = \widetilde{\bm{D}}^{1/2}{(\bm{A}+\bm{I})}\widetilde{\bm{D}}^{1/2}$ & $f = lambda\ x: x$                                     & Fixed                     \\
			\midrule
			GAT        & Self Attention      & $\begin{cases} \mathcal{C} = (\bm{A}+\bm{I})\cdot \mathcal{T} \\ \mathcal{T}_{(i,j)} = \frac{exp([\Theta \bm{x}_i || \Theta \bm{x}_j]\cdot \bm{a})}{\sum_{k \in \mathcal{N}(i) \cup i} exp([\Theta \bm{x}_i || \Theta \bm{x}_j]\cdot \bm{a})}  \end{cases}$                                                      & $f = \Theta \in \mathbb{R}^{d_k \times d_{k+1}}$ & Learnable                 \\
			\bottomrule
			\multicolumn{5}{l}{ $^a$ $\widetilde{\bm{D}}$ is the diagonal matrix of $(\bm{A}+\bm{I})$}
		\end{tabular}}
		% \vspace{-5mm}
	\label{tab:framework_for_mp_1}
\end{table*}

\paragraph{Relation network.}
To enhance the ability to capture logical evidence for reasoning, we propose a new KG modeling perspective, which transforms the target KG into hyper-level views by representing each triple as a relational node. Specifically, we construct the \emph{relation network} via a relation-driven process. Following a systematic approach, we treat each triple as a node in the relational graph and establish connections between nodes if they involve a shared entity within the original KG.

\begin{definition}\label{def:relational_graph_noran}
	\textbf{Relation network.} Given a knowledge graph $\mathcal{G}=\{ (h,r,t) | h,t \in \mathcal{E}, r \in \mathcal{R}\}$, the corresponding relation network is the triple-level graph $\mathcal{G}_r = (\widetilde{\mathcal{V}}, \widetilde{\mathcal{A}}, \widetilde{\mathcal{X}_{r}})$, where $\widetilde{\mathcal{V}}$ and $\widetilde{\mathcal{A}}$ are the sets of relational nodes and adjacency metrics, respectively. $\widetilde{\mathcal{X}_{r}} = \{ \Gamma(h,r,t) \ |\ (h,r,t) \in \mathcal{G} \}$ represents the feature matrix of $\widetilde{\mathcal{V}}$, where $\Gamma(\cdot)$ is the concatenation function that transforms each relational triple into a node, i.e., $v = \Gamma(h,r,t)$. $\widetilde{\mathcal{A}}(v, u|v,u \in \widetilde{\mathcal{V}}) = 1$, if $v$ and $u$ share the same entity in the original KG.
\end{definition}

\paragraph{Interpretation with logic reasoning.}~\label{sec:def_logicrule}
The topology of the relation network effectively depicts relational distributions, while the connections between nodes convey relation correlations. By modeling the relation network using a $k$-layer message-passing framework, we can capture relational patterns as entity-independent context for inductive KG completion. This entity-independent context is formally defined as \textit{k-hop logical evidence} over the relation network.

\begin{definition}\label{def:logic_rule}
	\textbf{K-hop logic evidence.} Let $\widetilde{\mathcal{G}} = (\widetilde{\mathbf{V}}, \widetilde{\mathcal{A}})$ be the relation network of a knowledge graph $\mathcal{G}=(\mathcal{E}, \mathcal{R})$. 	Given any center node $v_i \in \widetilde{\mathbf{V}}$, the k-hop ego graph $\widetilde{\mathcal{G}_i}=(\widetilde{\mathbf{V}}_i, \widetilde{\mathbf{A}}_i)$ centering at $v_i$ contains the relational contextual information for logical reasoning. In this paper, we model such contextual information as $\Lambda^{k}(v_i)$ via a k-layer message-passing network $\Omega$. Thus, the overall logic evidence of knowledge graph $\mathcal{G}$ can be modeled by traversing all k-hop ego graphs, i.e., $\Lambda(\mathcal{G}) = \Omega_{GNN}(\Lambda^{k}(v)\ |\ v \in \widetilde{\mathcal{G}})$. 
\end{definition}

Taking the logical evidence ``\textit{:LiveIn = :WorkAt $\land$ :LocatedIn}'' in Fig.~\ref{fig:illustration} as an example, the relation pattern among \textit{Elon Musk}, \textit{TESLA}, and \textit{California State} illustrates a concrete case of logical evidence. The triple \textit{(Martin Eberhard, :LiveIn, California State)} can then be inferred with confidence. According to Definition~\ref{def:logic_rule}, such logical evidence corresponds to $k$-hop ego graphs centered on target nodes in the relation network (Fig.~\ref{fig:illustration}~$(ii)$), describing the relational context required for reasoning.

The construction of the relation network offers two main advantages: $(i)$ It introduces a novel perspective for KG modeling, enabling entity-independent logic evidence to be naturally captured via $k$-hop ego graphs sampled from the relation network. This eliminates the need for fine-grained embeddings of unseen entities. $(ii)$ Compared to the original KG, the relation network exhibits a denser structure and is compatible with any GNN model. An incident graph (incidence matrix representation) could also encode relationships between nodes (entities) and edges (relations) in a bipartite structure, where rows represent nodes and columns represent edges. While the incidence matrix captures explicit connections between entities and relations, it lacks the higher-order relational semantics that NORAN’s Relation Network emphasizes.

\subsection{Relational Message Passing}\label{sec:logic_rule_infomax}

To effectively extract entity-independent contextual information as logical evidence, we introduce a novel relational message-passing framework tailored for inductive KG completion.

\paragraph{Feature initialization.}\label{sec:instantiations_of_gamma} 
Since each instance in the relation network corresponds to a triple $(h, r, t)$ from the original KG, we first initialize the embeddings of entities and relations in the original KG randomly. To capture the local relational structures within each triple, we utilize a set of Bi-LSTM units as a local information modeling layer. The embedding of each triple is obtained as follows:
\begin{equation} \label{equ:lstm}
	\bm{x}_i = \Gamma(h,r,t) = concat(\Phi(\bm{e}_h, \bm{e}_r, \bm{e}_t)),
\end{equation}
where $\Phi(\cdot)$ is a Bi-LSTM unit, and $\bm{e}_h, \bm{e}_r, \bm{e}_t$ are the initial embeddings of $h$, $r$, and $t$. The resulting triple embedding $\bm{x}_i$ effectively encodes the relational structure of the input triple, and it is used as the initial node embedding in the relation network. During both training and inference, the entity embeddings are kept \textit{fixed}.

\paragraph{Message passing.}
The message-passing layer is defined as:
\begin{equation} 
	\bm{X}^{(k+1)} = \sigma \big(\mathcal{C}^{(k)}\bm{X}^{(k)} \circ f^{(k)}\big),
\end{equation}
where $\bm{X}^{(k)}$ is the relational embedding at the $k$-th layer; $\mathcal{C}^{(k)}$ is the convolutional matrix for the $k$-th layer; $f^{(k)} : \mathbb{R}^{d_k} \rightarrow \mathbb{R}^{d_{k+1}}$ is the linear transformation matrix; and $\sigma$ is the activation function.  

The reformulated message-passing layers are summarized in Table~\ref{tab:framework_for_mp_1}. By performing message passing over relational nodes, we can naturally capture relation patterns as entity-independent contextual information, which is essential for inductive KG completion.

In this work, we train two message-passing GNNs, $\Omega$ and $\Psi$: $\Omega$ is applied to the $k$-hop ego graph to extract logical evidence, while $\Psi$ is applied to the \textit{relation network} to learn relational embeddings for inductive inference.

\paragraph{Mutual information maximization.}
Reasoning with logical evidence is pivotal for inductive KG completion, as it emulates human-like inference. However, as mentioned earlier, logical evidence is challenging to represent and has typically been used as a regularization mechanism in prior work. To address this, we propose \textit{Logic Evidence Information Maximization} (LEIM), which aims to maximize the mutual information between the logical evidence extracted from $k$-hop ego graphs and the relational semantics of the center node. The optimization objective is defined as:
\begin{equation} \label{equ:LEIM}
	\mathcal{L}_{\omega, \psi} = -\mathcal{I}_{\omega}(\Lambda(\mathcal{G}), \Psi^k(\widetilde{\mathcal{G}})),
\end{equation}
where $\Lambda(\mathcal{G}) = \Omega_{GNN}(\Lambda^{k}(v)\ |\ v \in \widetilde{\mathcal{G}})$ is the logical evidence extracted by $\Omega$, $\mathcal{I}(\cdot, \cdot)$ is the mutual information, $\Psi^k(\cdot)$ is a $k$-layer GNN, and $\omega, \psi$ are the trainable parameters for $\mathcal{I}$ and $\Psi$, respectively. The objective minimizes the loss to find the optimal parameters $\hat{\psi}$ that preserve the mutual information between the logical evidence $\Lambda(\mathcal{G})$ and the relational embedding of the target node $v$ learned by $\Psi^k(\widetilde{\mathcal{G}})$.

In practice, we adopt a Jensen-Shannon MI estimator (JSD) to estimate the mutual information:
\begin{multline}\label{equ:jsd_mi}
	\mathcal{I}^{JSD}_{\omega, \psi}(\Lambda, \Psi) = \mathbb{E}_\mathbb{P}\Big(-sp(-T_{\omega, \psi}(\Lambda^k(v), \Psi^k(v)))\Big) \\ - \mathbb{E}_{\mathbb{P}\times\mathbb{P}'}\Big(-sp(T_{\omega, \psi}(\Lambda^k(v'), \Psi^k(v)))\Big),
\end{multline}
where $v$ is a node sampled from the relation network under distribution $\mathbb{P}$, $v'$ is sampled from a negative distribution $\mathbb{P}'=\mathbb{P}$, and $sp$ represents the softplus activation function. To encourage alignment of positive pairs, e.g., $(\Lambda(v), \Psi(v))$, and distinguish them from negative pairs, e.g., $(\Lambda(v'), \Psi(v))$ where $v \neq v'$, we use $T_{\omega, \psi}(\Lambda, \Psi)$ to predict the correlation between $\Lambda$ and $\Psi$ as follows:
\begin{equation} 
	T_{\omega, \psi}(\Lambda^k(v), \bm{x}_v) = \sum_{(p,q) \in \Lambda^k(v)}\log (f_\omega([\bm{h}_p || \bm{h}_q || \bm{x}_v])),
\end{equation}
where $\bm{h}_p$ and $\bm{h}_q$ are the features of source and target nodes $p$ and $q$ obtained from the $k$-hop logical ego graph $\Lambda^k(\mathcal{G})$, and $\bm{x}_v = \Psi^k(v)$ is the relational embedding of node $v$. 
$f:\mathbb{R}^{ |\bm{h}| + |\bm{h}| + |\bm{x}| } \to \mathbb{R}^{(0,1)}$ is a fully connected discriminator used to differentiate positive and negative pairs.

\subsection{Training Scheme and KG Inference}\label{sec:complete_alg}

\paragraph{Training objective.}
This section outlines the complete algorithm for \textit{NORAN} along with the training procedure. After training, we obtain the model \(\Psi \circ \Gamma\), which is used to infer representations for inductive triples. 
Given the relational embedding learned for an inductive triple via \(\Psi \circ \Gamma\), we perform link prediction using a classifier based on \textit{logistic regression}, defined as \(p(\bm{x}) = Sigmoid(\bm{w}^T\bm{x}+b)\). 

\paragraph{Inductive inference and complexity analysis.}
For inductive inference, given a triple \(t=(h,r,t)\) with an unseen entity $h$ or $t$, we first create a new node \(v=\Gamma(h,r,t)\) and update the original relation network \(\widetilde{\mathcal{G}}\) by adding node \(v\) to it. The probability of triple existence is then predicted using \(p\circ \Psi \circ \Gamma\).

Unlike the \textit{Embedding-based Paradigm}, which requires retraining for inductive inference, our framework only updates the relation network and performs inference with the trained model \(p\circ \Psi \circ \Gamma\). The time complexity of inference in our framework is:
\begin{equation*}
	\mathcal{O}(\underbrace{3bf^2}_{\Gamma} + \underbrace{bd^L f + bLf^2}_{\Psi}),
\end{equation*}
where \(b\) is the number of inductive triples, \(f\) is the hidden dimension (assuming all embeddings have the same dimension), \(d\) is the average node degree in relation network \(\widetilde{\mathcal{G}}\), and \(L\) is the number of layers in \(\Psi\). The time complexity of \(p\) and relation network updates is negligible compared to the rest. Thus, the primary computational cost arises from the message-passing operation in \(\Psi\), which grows exponentially with \(L\). However, empirical results indicate that a two-layer GNN typically suffices, as the local view of GNNs is naturally suited to this task. This aligns with existing research advocating for shallower GNNs.

\section{Experiments}
This section provides an empirical evaluation of the proposed framework and its performance across five KG datasets. The study is guided by the following research questions:
\begin{itemize}
	\item \textit{\textbf{Q1}}~(\ref{sec:main_results}): How does our proposed framework compare against the strongest baselines, including traditional embedding-based, rule-based, and other message-passing (MP) methods?
	\item \textit{\textbf{Q2}}~(\ref{sec:ablation_relation_network}): With respect to Remark~\ref{remark: linking patterns} and Remark~\ref{remark: linking directions}, does our proposed \textit{relation network} effectively model inductive KG completion?
	\item \textit{\textbf{Q3}}~(\ref{sec:ablation_I}). Is our proposed \textit{logic evidence information maximization} a more effective training objective than standard negative sampling for inductive KGC?
\end{itemize}

\subsection{Experimental Setup}

\paragraph{Datasets.}
We evaluate our framework on five widely-used KG benchmarks: $(i)$ \textbf{FB15K-237}, a subset of FB15K with inverse relations removed; $(ii)$ \textbf{WN18RR}, a subset of WN18 with inverse relations excluded; $(iii)$ \textbf{NELL995}, derived from the $995$-th iteration of the NELL system and containing general knowledge; $(iv)$ \textbf{OGBL-WIKIKG2}, extracted from the Wikidata knowledge base; and $(v)$ \textbf{OGBL-BIOKG}, constructed from biomedical repositories. 
\begin{table*}[t]
	\centering
	\caption{Main results of inductive KGC on five benchmark datasets. We underline the best results within each of the three categories and bold NORAN's results that are better than all baselines.}
	% \vspace{-4mm}
	\label{tab:comparison_results}
	\resizebox{1.0\textwidth}{!}{
		\begin{tabular}{clcccccccccccccccc}
			\toprule
			% \multicolumn{16}{c}{Precision@K}\\ \midrule
			\multirow{2}{*}{Categories} & Datasets                                & \multicolumn{3}{c}{FB15K-237} & \multicolumn{3}{c}{WN18RR} & \multicolumn{3}{c}{NELL995} & \multicolumn{3}{c}{	OGBL-WIKIKG2} & \multicolumn{3}{c}{OGBL-BIOKG}                                                                                                                                                                                                                                                                                \\
			\cmidrule(lr){2-2} \cmidrule(lr){3-5} \cmidrule(lr){6-8} \cmidrule(lr){9-11} \cmidrule(lr){12-14} \cmidrule(lr){15-17}
			                            & Metrics                                 & MRR                           & Hit@1                      & Hit@3                       & MRR                              & Hit@1                          & Hit@3                      & MRR                        & Hit@1                      & Hit@3                      & MRR                        & Hit@1                      & Hit@3                      & MRR                        & Hit@1                      & Hit@3   \\
			\midrule
			\multirow{5}{*}{Emb. Based}
			                            & TransE
			                            & 0.289                                   & 0.198                         & 0.324                      & 0.265                       & 0.058                            & 0.445                          & 0.254                      & 0.169                      & 0.271                      & 0.213                      & 0.122                      & 0.229                      & 0.317                      & 0.26                       & 0.345                                \\
			                            & DistMult
			                            & 0.241                                   & 0.155                         & 0.263                      & 0.430                       & 0.390                            & 0.440                          & 0.267                      & 0.174                      & 0.295                      & 0.199                      & 0.115                      & 0.210                      & 0.341                      & 0.278                      & 0.363                                \\
			                            & ComplEx
			                            & 0.247                                   & 0.158                         & 0.275                      & 0.440                       & 0.410                            & 0.460                          & 0.227                      & 0.149                      & 0.249                      & 0.236                      & 0.136                      & 0.254                      & 0.322                      & 0.257                      & 0.360                                \\
			                            & SimplE
			                            & \underline{0.338}                       & \underline{0.241}             & \underline{0.375}          & \underline{0.476}           & \underline{0.428}                & \underline{0.492}              & \underline{0.291}          & 0.198                      & \underline{0.314}          & 0.220                      & 0.131                      & 0.239                      & 0.319                      & 0.246                      & 0.358                                \\
			                            & QuatE
			                            & 0.319                                   & \underline{0.241}             & 0.358                      & 0.446                       & 0.382                            & 0.478                          & 0.285                      & \underline{0.201}          & 0.307                      & \underline{0.248}          & \underline{0.139}          & \underline{0.262}          & \underline{0.363}          & \underline{0.294}          & \underline{0.382}                    \\
			\cmidrule(lr){2-17}
                \multirow{2}{*}{Rule Based}
                                        &RuleN    &0.453       &0.387      &0.491      &0.514      &0.461      &0.532      &0.346      &0.279      &0.366 &- &- &- &- &- &- \\
                                        &DRUM     &0.447       &0.373      &0.478      &0.521      &0.458      &0.549      &0.340      &0.261      &0.363 &- &- &- &- &- &- \\
                \cmidrule(lr){2-17}
			\multirow{7}{*}{MP based}
			                            & RGCN
			                            & 0.427                                   & 0.367                         & 0.451                      & 0.501                       & 0.458                            & 0.519                          & 0.329                      & 0.256                      & 0.348                      & 0.285                      & 0.176                      & 0.324                      & 0.381                      & 0.319                      & 0.399                                \\
			                            & RGHAT
			                            & 0.440                                   & 0.361                         & 0.483                      & 0.518                       & 0.460                            & 0.540                          & 0.337                      & 0.274                      & 0.351                      & 0.301                      & 0.192                      & 0.329                      & 0.395                      & 0.334                      & 0.418                                \\
			                            & GraIL
			                            & 0.465                                   & 0.389                         & 0.482                      & 0.512                       & 0.453                            & 0.539                          & \underline{0.355}          & \underline{0.282}          & 0.367                      & 0.327                      & 0.201                      & 0.336                      & 0.434                      & 0.379                      & 0.451                                \\
                               &ConGLR &0.463& 0.402& 0.483& 
                                0.512& 0.452& 0.541
                               & 0.352& 0.276& 0.366& 0.318& 0.219& 0.338& 0.422& 0.365& 0.431\\

			                            & PATHCON
			                            & \underline{0.483}                       & \underline{0.425}             & \underline{0.499}          & \underline{0.522}           & \underline{0.462}                &0.546           & 0.349                      & 0.276                      & \underline{0.369}          & \underline{0.339}          & \underline{0.243}          & \underline{0.347}          & \underline{0.457}          & \underline{0.395}          & \underline{0.472}                    \\
                               % https://github.com/zjukg/RMPI
                               &RMPI & 0.459& 0.396& 0.480 &
                               0.514& 0.454& 0.544
                               & 0.339& 0.277& 0.365& 0.313& 0.213& 0.336& 0.414& 0.358& 0.421\\
                               
                               & MBE &0.477 &0.410 &0.495 &0.519 &0.451 &\underline{0.549}  &0.344 &0.270 &0.359 &0.331 &0.234 &0.345 &0.452 &0.380 &0.470 \\
			\cmidrule(lr){2-17}
			\multirow{3}{*}{Ours$^a$}
			                            & NORAN(GS)
			                            & \textbf{0.483}                          & \textbf{0.440}                & \textbf{0.504}             & \textbf{0.535}              & \textbf{0.471}                   & \textbf{0.564}                 & \textbf{0.364}             & \textbf{0.298}             & \textbf{0.381}             & \textbf{0.349}             & 0.237                      & \textbf{0.361}             & 0.454                      & \textbf{0.395}             & \textbf{0.479}                       \\
			                            & NORAN(GIN)
			                            & \textbf{\underline{0.499}}              & \textbf{\underline{0.451}}    & \textbf{\underline{0.519}} & \textbf{0.530}              & \textbf{0.467}                   & \textbf{0.560}                 & \textbf{0.370}             & \textbf{0.308}             & \textbf{0.379}             & \textbf{0.353}             & \textbf{0.251}             & \textbf{0.367}             & \textbf{0.469}             & \textbf{0.407}             & \textbf{0.492}                       \\
			                            & NORAN(GAT)
			                            & 0.468                                   & 0.422                         & 0.489                      & \textbf{\underline{0.540}}  & \textbf{\underline{0.499}}       & \textbf{\underline{0.575}}     & \textbf{\underline{0.374}} & \textbf{\underline{0.310}} & \textbf{\underline{0.392}} & \textbf{\underline{0.358}} & \textbf{\underline{0.260}} & \textbf{\underline{0.371}} & \textbf{\underline{0.475}} & \textbf{\underline{0.411}} & \textbf{\underline{0.495}}           \\ \cmidrule(lr){2-17}
			\textit{NORAN - Emb.}$^b$   & +2.2\% (Avg.)                           & +1.6\%                        & +2.6\%                     & +2.0\%                      & +1.8\%                           & +3.7\%                         & +2.9\%                     & +1.9\%                     & +2.8\%                     & +2.3\%                     & +1.9\%                     & +1.7\%                     & +2.4\%                     & +1.8\%                     & +1.6\%                     & +2.3\%  \\
			\textit{NORAN - GNN}$^b$    & +11.2\% (Avg.)                          & +16.1\%                       & +21.0 \%                   & +14.4\%                     & +6.4\%                           & +7.1\%                         & +8.3\%                     & +8.3\%                     & +10.9\%                    & +7.8\%                     & +11.0 \%                   & +12.1\%                    & +10.9\%                    & +11.2\%                    & +11.7\%                    & +11.5\% \\
			\bottomrule
			\multicolumn{14}{l}{$^a$ We test our NORAN with three backbones: GraphSAGE, GIN, and GAT, denoted as NORAN(GS), NORAN(GIN), NORAN(GAT), respectively. }                                                                  \\
			\multicolumn{14}{l}{$^b$ We show the margin between the best results of NORAN and the ones of the two branches of baselines methods. }
		\end{tabular}
	}
 \vspace{-3mm}
	\label{tab:main_results_noran}
\end{table*}

\paragraph{Baselines.}
To validate the effectiveness of NORAN, we compare it against state-of-the-art baselines. These baselines are categorized into three groups: $(i)$ \textit{embedding-based} methods, including \textbf{TransE}, \textbf{DistMult}, \textbf{ComplEx}, \textbf{SimplE}, and \textbf{QuatE}; $(ii)$ \textit{rule-based} methods, such as \textbf{RuleN} and \textbf{DRUM}; and $(iii)$ \textit{MP-based} methods, including \textbf{RGCN}, \textbf{RGHAT}, \textbf{GraIL}, \textbf{ConGLR}, \textbf{PATHCON}, \textbf{RMPI}, and \textbf{MBE}. Notably, the embedding-based methods are not inherently inductive, so we retrain their embedding tables for inductive inference. For MP-based methods, we select the most effective ones for KGC in the same setting. We exclude certain related works from our experiments as their original implementations are tailored for KGC with unseen relations, whereas our model focuses on KGC with unseen entities.

\paragraph{Implementation Details.}
The five benchmark datasets were originally designed for the \textit{transductive setting}, where test entities are subsets of training entities. To adapt them for inductive testing, we create new inductive datasets by sampling disjoint subgraphs from the original KGs. Specifically, we randomly select a subset of entities from the test set, remove them and their associated edges from the training set, and use the remaining training set for model training. The removed edges are reintroduced during evaluation. We use \textbf{MRR} (mean reciprocal rank) and \textbf{Hit@1, 3} as evaluation metrics.

All baseline methods are implemented using their open-source codebases. The models are implemented in PyTorch and trained on an RTX 3090 GPU with 24 GB RAM. For fairness, the embedding size for entities and relations is set to $100$ for the input, latent, and output layers across all models, except for PATHCON, which uses dataset-specific node labeling for its input embedding size. We optimize all models using the Adam optimizer with a batch size of $256$, except for the larger OGBL datasets, where the batch size is reduced to $32$ to avoid memory issues. Model parameters are initialized using the Xavier initializer, with an initial learning rate of $0.005$. Hyperparameters are tuned via grid search, with the learning rate selected from $\{0.05, 0.01, 0.005, 0.001\}$ and the margin parameter for negative sampling ranging from $0$ to $1$. The best configurations are used as defaults for other hyperparameters. To ensure reproducibility, we use a fixed random seed and report the average results of three runs.

\subsection{Main Results: Q1}\label{sec:main_results}
To address Q1, we conduct extensive experiments comparing NORAN with the strongest baselines. The results are summarized in Table~\ref{tab:main_results_noran}, with key observations outlined below.

\noindent \textit{Obs. 1. NORAN significantly outperforms state-of-the-art KGC baselines.} We compare NORAN against $5$ embedding-based, $2$ rule-based, and $7$ MP-based methods on $5$ KG datasets. As shown in Table~\ref{tab:main_results_noran}, NORAN achieves superior performance across all evaluation metrics (MRR, Hit@1, Hit@3). Results for NORAN are \textbf{bolded} when they outperform all baselines. We evaluate NORAN with three backbones, and the bolded results dominate, highlighting the framework's effectiveness. Additionally, we compute the margin between NORAN's best results (\underline{underlined}) and those of the baselines. NORAN outperforms the best embedding-based and MP-based baselines by an average margin of $11.2\%$ and $2.2\%$, respectively.

\begin{table*}[t]
	\centering
	\caption{Results of ablation study for relation network construction rules on three benchmark datasets.}
	\label{tab:ablation_relation_network_app}
 % \vspace{-4mm}
	% 	\setlength{\tabcolsep}{5.5pt}
	\resizebox{.9\textwidth}{!}{
		\begin{tabular}{clcccccccccc}
			\toprule
			% \multicolumn{16}{c}{Precision@K}\\ \midrule
			\multirow{2}{*}{Construction Rule} & Datasets   & \multicolumn{3}{c}{WN18RR} & \multicolumn{3}{c}{NELL995} & \multicolumn{3}{c}{	OGBL-WIKIKG2}                                                 \\
			\cmidrule(lr){2-2} \cmidrule(lr){3-5} \cmidrule(lr){6-8} \cmidrule(lr){9-11}
			                                   & Metrics    & MRR                        & Hit@1                       & Hit@3                            & MRR   & Hit@1 & Hit@3 & MRR   & Hit@1 & Hit@3 \\
			\midrule
			\multirow{1}{*}{Default}
			                                   & NORAN(GAT)
			                                   & 0.540      & 0.499                      & 0.575                       & 0.374                            & 0.310 & 0.392 & 0.358 & 0.260 & 0.371         \\
			\cmidrule(lr){2-11}

			\multirow{1}{*}{w.o. head-head}
			                                   & NORAN(GAT)
			                                   & 0.521      & 0.457                      & 0.558                       & 0.348                            & 0.273 & 0.375 & 0.329 & 0.230 & 0.341         \\
			\cmidrule(lr){2-11}

			\multirow{1}{*}{w.o. tail-tail}
			                                   & NORAN(GAT)
			                                   & 0.517      & 0.454                      & 0.549                       & 0.351                            & 0.288 & 0.364 & 0.327 & 0.224 & 0.342         \\
			\cmidrule(lr){2-11}

			\multirow{1}{*}{w.o. head-tail}
			                                   & NORAN(GAT)
			                                   & 0.501      & 0.446                      & 0.532                       & 0.339                            & 0.275 & 0.359 & 0.314 & 0.212 & 0.319         \\

			\bottomrule
		\end{tabular}}
\vspace{-2mm}
\end{table*}

\noindent \textit{Obs. 2. NORAN with GAT (learnable convolution matrix) generally outperforms MP layers with fixed convolution matrices.} We test NORAN with three backbones: GAT, GraphSAGE, and GIN. GAT, which uses a learnable convolution matrix, achieves the best results (\underline{underlined}) on four datasets, while GraphSAGE and GIN exhibit comparable performance.

\subsection{Ablation Study on Relation Network: Q2}\label{sec:ablation_relation_network}
The relation network's construction involves two key aspects: $(i)$ the semantics of various `\textit{entity sharing}' patterns (Remark~\ref{remark: linking patterns}) and $(ii)$ the semantics of `\textit{linking direction}' (Remark~\ref{remark: linking directions}). We conduct an ablation study to validate these aspects.

\begin{remark}\label{remark: linking patterns}
	\textbf{Linking pattern}. For any two triples sharing entities, i.e., $T_1=(h_1, r_1, t_1) \cap T_2=(h_2, r_2, t_2)$, there are three linking patterns: $(i)$ head-head sharing~($h_1 = h_2)$, $(ii)$ tail-tail sharing~($t_1 = t_2$), and $(iii)$ head-tail sharing~($t_1 = h_2 \oplus t_2=h_1$). Our construction criterion links all three patterns, as they carry distinct semantic meanings.
\end{remark}
\begin{remark}\label{remark: linking directions}
	\textbf{Linking direction}. As per Def.~\ref{def:relational_graph}, edges in the relation network are bidirectional, regardless of the linking pattern. This design is based on the rationale that relation semantics in KGs are not direction-sensitive. For example, the triple \textit{(Elon Musk, :WorkIn, TESLA)} in Figure~\ref{fig:illustration} (i) is semantically equivalent to \textit{(TESLA, :Foundedby, Elon Musk)}.
\end{remark}

\begin{table*}[t]
	\centering
	\caption{Ablation study of training objective, i.e., \textit{Logic Evidence Information Maximization} (LEIM)}
	\label{tab:ablation_I_app}
 % \vspace{-4mm}
	% 	\setlength{\tabcolsep}{5.5pt}
		\resizebox{0.9\textwidth}{!}{
	\begin{tabular}{clcccccccccc}
		\toprule
		% \multicolumn{16}{c}{Precision@K}\\ \midrule
		\multirow{2}{*}{Training Objective} & Datasets            & \multicolumn{3}{c}{WN18RR} & \multicolumn{3}{c}{NELL995} & \multicolumn{3}{c}{	OGBL-WIKIKG2}                                                 \\
		\cmidrule(lr){2-2} \cmidrule(lr){3-5} \cmidrule(lr){6-8} \cmidrule(lr){9-11}
		                                   & Metrics             & MRR                        & Hit@1                       & Hit@3                            & MRR   & Hit@1 & Hit@3 & MRR   & Hit@1 & Hit@3 \\
		\midrule
		\multirow{3}{*}{\shortstack{Naive                                                                                                                                                            \\ Negative Sampling}}
		                                   & NORAN(GS)
		                                   & 0.529               & 0.469                      & 0.554                       & 0.351                            & 0.290 & 0.357 & 0.324 & 0.215 & 0.322         \\
		                                   & NORAN(GIN)
		                                   & 0.537               & 0.490                      & 0.567                       & 0.357                            & 0.291 & 0.374 & 0.335 & 0.231 & 0.349         \\
		                                   & NORAN(GAT)
		                                   & 0.534               & 0.481                      & 0.562                       & 0.361                            & 0.299 & 0.377 & 0.342 & 0.238 & 0.355         \\

		\cmidrule(lr){2-11}

		\multirow{3}{*}{LEIM (JSD)}
		                                   & NORAN(GS)
		                                   & 0.535               & 0.471                      & 0.564                       & 0.364                            & 0.298 & 0.381 & 0.349 & 0.237 & 0.361         \\
		                                   & NORAN(GIN)
		                                   & 0.530               & 0.467                      & 0.560                       & 0.370                            & 0.308 & 0.379 & 0.353 & 0.251 & 0.367         \\
		                                   & NORAN(GAT)
		                                   & 0.540               & 0.499                      & 0.575                       & 0.374                            & 0.310 & 0.392 & 0.358 & 0.260 & 0.371         \\
		\cmidrule(lr){2-11}

		\multirow{3}{*}{LEIM (InfoNCE)}
		                                   & NORAN(GS)
		                                   & 0.532               & 0.476                      & 0.558                       & 0.358                            & 0.294 & 0.369 & 0.331 & 0.220 & 0.343         \\
		                                   & NORAN(GIN)
		                                   & 0.541               & 0.495                      & 0.583                       & 0.365                            & 0.301 & 0.376 & 0.339 & 0.231 & 0.359         \\
		                                   & NORAN(GAT)
		                                   & 0.544               & 0.506                      & 0.581                       & 0.367                            & 0.306 & 0.383 & 0.337 & 0.232 & 0.352         \\

		\bottomrule
	\end{tabular}
 \vspace{-5mm}
		}
\end{table*}

To address Q2, we perform an ablation study on the relation network's construction. For Remark~\ref{remark: linking patterns}, we iteratively remove each linking pattern (head-head, head-tail, tail-tail) and evaluate the resulting network using GAT as the backbone. The results, shown in Table~\ref{tab:ablation_relation_network_app}, demonstrate that removing any pattern significantly degrades NORAN's performance, validating our construction approach.

\noindent \textit{Obs. 3. The head-tail pattern is the most critical for relation network construction.} Ablating the head-tail pattern results in a larger performance drop compared to the other patterns. This pattern aligns with translation-based logic rules, which are central to multi-hop KG reasoning. In contrast, head-head and tail-tail patterns correspond to union and intersection operations, respectively, as defined in complex KG reasoning. While these operations also contribute to KGC, they are often overlooked by translation-based embedding methods, explaining their weaker inductive performance in Table~\ref{tab:main_results_noran}.

\subsection{Ablation Study for Training Objective: Q3}\label{sec:ablation_I}
To address Q3, we evaluate the effectiveness of LEIM by comparing it with an InfoNCE estimator and naive negative sampling (NS). The InfoNCE estimator is defined as:
\begin{multline}\label{equ:infonce_mi}
	\mathcal{I}^{InfoNCE}_{\omega, \psi}(\Lambda, \Psi) = \mathbb{E}_\mathbb{P}\Bigg(T_{\omega, \psi}(\Lambda^k(v), \Psi^k(v)) \\ - \mathbb{E}_{\mathbb{P}'}\Big(\log \sum_{v'\sim\mathbb{P}'}e^{T_{\omega, \psi}(\Lambda^k(v'), \Psi^k(v))}\Big)\Bigg).
\end{multline}
We evaluate LEIM using both JSD (default) and InfoNCE estimators and compare them to naive NS. Results are shown in Table~\ref{tab:ablation_I_app}, with the best results bolded by column. LEIM consistently outperforms naive NS across all backbones, demonstrating its effectiveness. The key difference between JSD and InfoNCE lies in their negative sampling strategies. JSD uses a $1:1$ negative sampling ratio within a batch, while InfoNCE employs a contrastive learning approach, using all other instances in the batch as negative samples. Given the computational efficiency and empirical results, we adopt JSD as the default. Additionally, we observe:

\noindent \textit{Obs. 4. GAT benefits the most from LEIM.} On three datasets, GAT achieves an average performance improvement of $1.2\%$ in MRR and $1.5\%$ in Hit@3 when trained with LEIM, outperforming other backbones. Notably, on WN18RR, GAT transitions from the worst-performing backbone under naive NS to the best-performing one under LEIM. In contrast, GIN, which uses a fixed convolution matrix, shows minimal or negative improvement with LEIM.

\section{Summary}
Inductive KG completion, which infers relations for newly introduced entities, aligns with real-world KGs that continuously evolve. In this work, we propose NORAN, a novel message-passing framework that leverages latent relation semantics for inductive KG completion. Our framework introduces a unique perspective on KG modeling through the \textit{relation network}, which treats relations as indicators of logical rules. Additionally, we propose \textit{logic evidence information maximization} (LEIM), an innovative training objective that preserves logical semantics in KGs. NORAN combines the relation network and LEIM to enable effective inductive KGC. Extensive experiments on five KG benchmarks demonstrate NORAN's superiority over state-of-the-art methods.

\chapter{Knowledge Graph Prompting for Large Language Models}
\section{Introduction}
% \section{LLM Hallucination}
% Large language models (LLMs) have redefined the boundaries of artificial intelligence, enabling machines to compose poetry, summarize scientific literature, and engage in nuanced dialogue. Yet, their prowess is tempered by a critical flaw: their reliance on parametric knowledge—static information encoded during training—leads to hallucinations, where models generate plausible but factually incorrect statements. For instance, an LLM might assert that "Marie Curie discovered penicillin" or invent fictitious historical events, errors that stem from their inability to dynamically access or verify external knowledge. These limitations are particularly problematic in domains where accuracy is non-negotiable, such as healthcare diagnostics, legal analysis, and financial forecasting, where a single factual error can have cascading consequences.

Large language models (LLMs), such as the Clude~\cite{anthropic2024claude} and  GPT series~\cite{openai2023gpt4}, have demonstrated exceptional capabilities across diverse tasks~\cite{yang2024evaluating,xiao2025reliable,dong2024modality,hong2024knowledge,zhang2024structure}, achieving breakthroughs in text comprehension~\cite{brown2020language}, question answering~\cite{khashabi2020unifiedqa}, and content generation~\cite{chowdhery2023palm,hong2024next,yuan2025knapsack}. Despite their success, LLMs face persistent criticism for their limitations in knowledge-intensive tasks, particularly those requiring domain expertise~\cite{Zhang-etal24KnowGPT,xiang2025use,zhang2025faithfulrag}. Their application in specialized domains remains challenging due to three key factors: (i) Knowledge limitations: LLMs possess broad but superficial knowledge in specialized fields, as their training data primarily consists of general-domain content, leading to gaps in domain-specific depth and alignment with current professional standards. (ii) Reasoning complexity: Specialized domains demand precise, multi-step reasoning governed by domain-specific rules and constraints. LLMs often struggle to maintain logical consistency and accuracy throughout extended reasoning chains, especially in technical or highly regulated contexts. (iii) Context sensitivity: Professional fields rely on nuanced, context-dependent interpretations where identical terms or concepts may carry different meanings based on specific scenarios. LLMs frequently fail to grasp these subtleties, resulting in misinterpretations or overly generalized responses.

To adapt LLMs for specific domains, researchers have explored various approaches, which can be broadly classified into two categories: (i) Fine-tuning LLMs with Domain-specific Data. Fine-tuning pre-trained LLMs on specialized datasets enables them to better capture domain-specific vocabulary, terminology, and patterns~\cite{gururangan2020don,lee2020patent,lee2020biobert,ge2024openagi}. This approach has been successfully applied in areas such as recommendation~\cite{LAGCN,zhou2023adaptive} and node classification~\cite{NEGCN,ALDI}, enhancing the relevance and accuracy of generated responses~\cite{huang2024large}. Fine-tuned LLMs have demonstrated effectiveness across various domains. In healthcare, they have been leveraged for clinical note analysis~\cite{alsentzer2019publicly}, biomedical text mining~\cite{lee2020biobert}, and medical dialogue~\cite{valizadeh2022ai}. Similarly, in the legal domain, they have proven useful for legal document classification~\cite{chalkidis2019neural}, contract analysis~\cite{chalkidis2021lexglue}, and legal judgment prediction~\cite{zhong2020does}. (ii) Retrieval-augmented generation (RAG). RAG provides an effective way to tailor LLMs for specialized domains without modifying the model architecture or parameters~\cite{lewis2020retrieval}. Instead of embedding new knowledge through retraining, RAG dynamically retrieves relevant domain-specific information from external sources, enhancing response accuracy and reliability. A typical RAG system operates in three stages: knowledge preparation, retrieval, and integration. First, external textual data is segmented into manageable chunks and transformed into vector representations for efficient indexing. During retrieval, relevant chunks are identified based on keyword matching or vector similarity when a query is submitted. Finally, the retrieved information is combined with the original query to generate well-informed responses.

Despite their success, RAG systems face significant challenges in practical applications due to the inconsistent quality of accessible data. Domain knowledge is frequently distributed across diverse sources—ranging from textbooks and research articles to technical manuals and industry reports~\cite{li2022survey}—which may vary in quality, accuracy, and completeness, potentially leading to discrepancies in the retrieved information~\cite{zhu2021retrieving}. A promising strategy to mitigate these issues is to integrate Knowledge Graphs (KGs) with LLMs. KGs offer a structured representation of domain knowledge, built on well-defined ontologies that specify specialized terminologies, acronyms, and their interrelations within a field~\cite{li2022constructing,shengyuan2024differentiable,zhang2022contrastive,zhang2023integrating,zhang2024logical}. The extensive factual content contained in KGs can help anchor model responses in established facts and principles~\cite{hu2023survey,yang2024give,pan2024unifying}.

\section{Introduction}
Earlier research explored heuristic methods to integrate knowledge from KGs into LLMs during pre-training or fine-tuning. For instance, ERNIE~\cite{sun2021ernie} aligns entity embeddings with word embeddings during pre-training to enhance entity understanding and reasoning. Similarly, KnowBERT~\cite{peters2019knowledge} integrates entity linkers with BERT to inject entity knowledge during fine-tuning for knowledge-intensive tasks. Another approach involves retrieving relevant knowledge from KGs at inference time to augment the LLM's context. For example, K-BERT~\cite{liu2020k} uses an attention mechanism to select relevant triples from KGs based on the query context, appending them to the input sequence. KEPLER~\cite{wang2021kepler} learns joint embeddings of text and KG entities to improve model predictions. Subsequent works have combined graph neural networks with LLMs for joint reasoning~\cite{QA-GNN,GreaseLM,dong2023hierarchy} and introduced interactions between text tokens and KG entities within LLM layers~\cite{sun2021jointlk,taunk2023grapeqa}.

However, as LLMs continue to evolve, most state-of-the-art models remain \emph{closed-source}. For example, GPT-4~\cite{openai2023gpt4} and Claude 3~\cite{anthropic2024claude} are accessible only through APIs, limiting access to model internals. Consequently, research has shifted toward KG prompting, which enhances fixed LLMs with KG-based hard prompts~\cite{liu2023pre,pan2024unifying}. KG prompting has emerged as a new paradigm in natural language processing. For instance, CoK~\cite{wang2023boosting} introduces Chain-of-Knowledge prompting, decomposing LLM-generated reasoning chains into evidence triples and verifying their accuracy using external KGs. Mindmap~\cite{wen2023mindmap} enhances transparency by enabling LLMs to reason over structured KG inputs. RoG~\cite{luo2023reasoning} proposes a planning-retrieval-reasoning framework that synergizes LLMs and KGs for interpretable reasoning. KGR~\cite{guan2024mitigating} autonomously refines LLM-generated responses by leveraging KGs to extract, verify, and refine factual information, mitigating hallucinations in knowledge-intensive tasks.

Despite the promising results of existing KG prompting methods, three critical challenges hinder their practical application. \textbf{\ding{182}} Vast search space. Real-world KGs often contain millions of triples, creating a large search space when retrieving relevant knowledge for prompting. \textbf{\ding{183}} High API costs. Accessing closed-source LLMs like GPT-4 and Claude 3 through proprietary APIs can be prohibitively expensive at scale~\cite{dong2024cost}, necessitating careful selection of the most informative knowledge. \textbf{\ding{184}} Labor-intensive prompt design. LLMs are highly sensitive to prompts, with minor variations potentially yielding drastically different responses. Existing methods rely on manually designed or rule-based prompts, which are inflexible and lack adaptability to varying question semantics and KG structures.

To address these challenges, we propose \underline{\textbf{Know}}ledge \underline{\textbf{G}}raph-based \underline{\textbf{P}}romp\underline{\textbf{T}}ing, or {\bf {KnowGPT}}, a framework that leverages factual knowledge from KGs to ground LLM responses in established facts. This paper addresses two key research questions: \textbf{\ding{182}} Given a query and a large-scale KG, how can we efficiently retrieve relevant factual knowledge? \textbf{\ding{183}} Given the extracted knowledge, how can we construct an effective prompt for LLMs?

We address these questions with a novel prompt learning method that is effective, generalizable, and cost-efficient. For question \textbf{\ding{182}}, we employ deep reinforcement learning (RL) to extract the most informative knowledge from KGs. A tailored reward scheme encourages the agent to discover concise, context-relevant knowledge chains. For question \textbf{\ding{183}}, we introduce a prompt construction strategy based on Multi-Armed Bandit (MAB). Given multiple knowledge extraction strategies and prompt templates, MAB selects the most effective combination for each question by balancing exploration and exploitation.

Our contributions are summarized as follows:
\begin{itemize}
\item We formally define the problem of KG-based prompting, which leverages structured knowledge from KGs to ground LLM responses in established facts.
\item We propose \textit{KnowGPT}, a novel prompting framework that combines deep RL and MAB to generate effective prompts for domain-specific queries.
\item We implement \textit{KnowGPT} on GPT-3.5. Experiments on three QA datasets demonstrate that {KnowGPT} outperforms state-of-the-art baselines by a significant margin. Notably, \textit{KnowGPT} achieves an average improvement of 23.7\% over GPT-3.5 and 2.9\% over GPT-4, with 92.6\% accuracy on the OpenbookQA leaderboard, comparable to human performance.
\end{itemize}

\section{KnowGPT Framework}
\begin{figure*}[t]
\centering
    \includegraphics[width=14.cm]{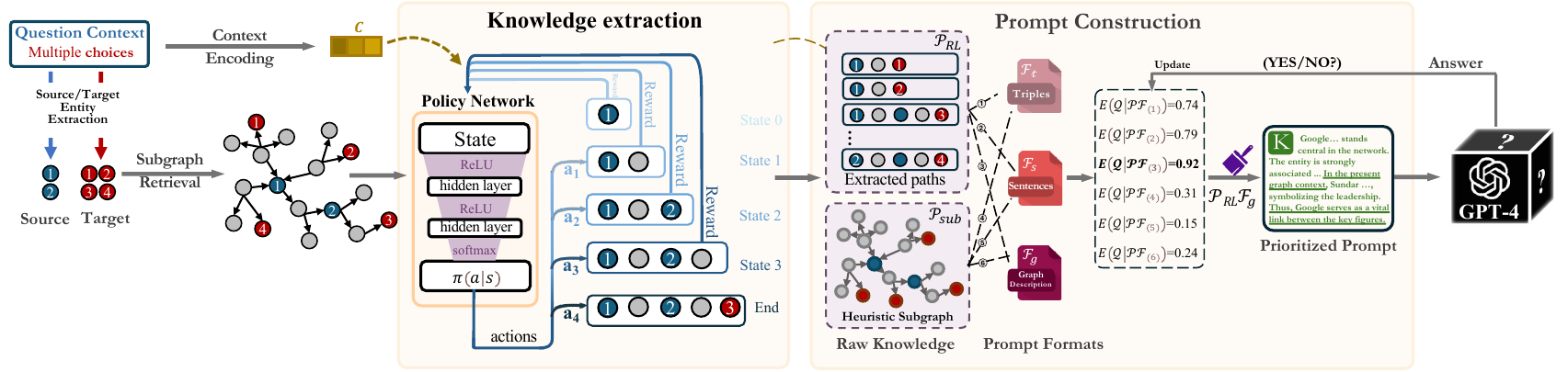}
    \vspace{-2mm}
    \caption{\footnotesize The overall architecture of our proposed knowledge graph prompting framework, i.e., {KnowGPT}.  Given the question context with multiple choices, we first retrieve a question-specific subgraph from the real-world KG. \textit{Knowledge Extraction} is first dedicated to searching for the most informative and concise reasoning background. Then the \textit{Prompt Construction} module is optimized to prioritize the combination of knowledge and formats subject to the given question.}
    \label{fig:fig1}
    % \vspace{-5mm}
\end{figure*}

Learning the prompting function $f_\text{prompt}(\mathcal{Q}, \mathcal{G})$ involves two challenges: determining what knowledge to use from $\mathcal{G}$ and how to construct the prompt. {KnowGPT} addresses these challenges by extracting knowledge with deep RL and constructing prompts with MAB. An overview of the framework is shown in Figure~\ref{fig:fig1}.

\subsection{Knowledge Extraction with Reinforcement Learning}
Intuitively, the relevant reasoning background lies in a question-specific subgraph $\mathcal{G}_{\text{sub}}$ containing all \textit{source} entities $\mathcal{Q}_s$, \textit{target} entities $\mathcal{Q}_t$, and their neighbors. An ideal subgraph $\mathcal{G}_{\text{sub}}$ should: \((i)\) include as many source and target entities as possible, \((ii)\) exhibit strong relevance to the question context, and \((iii)\) be concise to fit within LLM input length constraints.

However, extracting such a subgraph is NP-hard. To address this, we develop $\mathcal{P}_\text{RL}$, a deep RL-based method that samples reasoning chains in a trial-and-error fashion. We assume $\mathcal{G}_{\text{sub}}$ is constructed from a set of reasoning chains $\mathcal{P} = \{\mathcal{P}_1, \mathcal{P}_2, ..., \mathcal{P}_m\}$, where each chain $\mathcal{P}_i = \{(e_i, r_1, t_1), (t_1, r_2, t_2), ..., (t_{|\mathcal{P}_i|-1}, r_{|\mathcal{P}_i|}, t_{|\mathcal{P}_i|})\}$ is a path in $\mathcal{G}$ starting from the $i$-$th$ source entity in $\mathcal{Q}_s$, and $|\mathcal{P}_i|$ is the path length. $\mathcal{G}_{\text{sub}}$ includes all entities and relations in $\mathcal{P}$.

\begin{itemize}
    \item \textbf{State:} A state represents the current entity in the KG. Specifically, it captures the spatial change from entity $h$ to $t$. Following prior work, we define the state vector $\boldsymbol{s}$ as:
    \begin{equation}
        \boldsymbol{s}_{t} = (\boldsymbol{e}_{t}, \boldsymbol{e}_{target} - \boldsymbol{e}_{t}), 
    \end{equation}
    where $\boldsymbol{e}_{t}$ and $\boldsymbol{e}_{target}$ are embeddings of the current and target entities. Initial node embeddings are obtained by transforming KG triples into sentences and feeding them into a pre-trained LM.
    
    \item \textbf{Action:} The action space includes all neighboring entities of the current entity, enabling flexible exploration of the KG.
    
    \item \textbf{Transition:} The transition model $\mathrm{P}$ measures the probability of moving to a new state ($s'$) given the current state ($s$) and action ($a$). In KGs, $\mathrm{P}(s'|s,a) = 1$ if $s$ is connected to $s'$ via $a$; otherwise, $\mathrm{P}(s'|s,a) = 0$.
    
    \item \textbf{Reward:} The reward is based on reachability:
\begin{equation}
r_{reach} =
\begin{cases}
    +1 , & \textit{if } target; \\
    -1 , & \textit{otherwise},
\end{cases}
\end{equation} 
indicating whether the path reaches the target within $K$ steps.
\end{itemize}

To promote context-relatedness and conciseness, we design two auxiliary rewards.

\paragraph{Context-relatedness Auxiliary Reward.}
This reward encourages paths closely related to the question context. We evaluate the semantic relevance of a path $\mathcal{P}_i$ to the context $\mathcal{Q}$ using a fixed matrix $\boldsymbol{W}$ to map the path embedding $\boldsymbol{\mathcal{P}}$ to the same semantic space as the context embedding $\boldsymbol{c}$. The reward is:
\vspace{-2mm}
\begin{equation}
    r_{\text{cr}} = \frac{1}{|i|} \sum_{source}^{i}cos(\boldsymbol{W\times\mathcal{P}}_{i}, \boldsymbol{c}),
\end{equation}
where $\boldsymbol{c}$ is the context embedding from a pre-trained LM~\cite{devlin2018bert}, and $\boldsymbol{\mathcal{P}}_{i}$ is the average embedding of entities and relations in the path.

\paragraph{Conciseness Auxiliary Reward.}
This reward encourages concise paths to fit within LLM input constraints and reduce API costs. The reward for path $\mathcal{P}_{i}$ is:
\begin{equation}
    r_{\text{cs}} = \frac{1}{|\mathcal{P}_{i}|}.
\end{equation}

\paragraph{Training Policy Network.}
We train a policy network $\pi_{\theta}(s,a) = p(a|s;\theta)$ using policy gradient~\cite{xiong2017deeppath} to maximize accumulated rewards. The optimal policy navigates from the source to the target entity while maximizing rewards. Additional training details are provided in the Appendix.

\subsection{Prompt Construction with Multi-armed Bandit}
We design a prompt construction strategy using Multi-Armed Bandit (MAB) to select the best combination of knowledge extraction strategies and prompt templates. Suppose we have knowledge extraction strategies $\{\mathcal{P}_1, \mathcal{P}_2, ..., \mathcal{P}_m\}$ and prompt templates $\mathcal{F} = \{\mathcal{F}_1, \mathcal{F}_2, ..., \mathcal{F}_n\}$. The goal is to identify the best combination for a given question.

We define the selection process as a reward maximization problem:
\begin{equation}
\sigma( f(\mathcal{P}\mathcal{F}_{(i)})) =
\begin{cases}
    1 & \textit{if } accurate; \\
    0 & \textit{otherwise}.
\end{cases}
\end{equation} 
Here, $\mathcal{P}\mathcal{F}_{(i)}$ is a combination of extraction strategy and prompt template, and $r_{pf}\in \{0,1\}$ indicates the LLM's performance.

We formulate the selection mechanism with an expectation function $E(\cdot)$:
\begin{equation}%\small
    E(\mathcal{Q}|\mathcal{P}\mathcal{F}_{(i)}) = \boldsymbol{c} \times \bm{\alpha}_{(i)} + \beta_{(i)}.
    \vspace{1mm}
\end{equation}
Here, $\boldsymbol{c}$ is the context embedding, $\bm{\alpha}{(i)}$ are learned parameters, and $\beta_{(i)}$ introduces noise for exploration.

We update $\bm{\alpha}_{(i)}$ using:
\begin{equation}%\small
\begin{aligned}
     & J({\bf C}_{(i)}^{(k)},\textbf{r}_{pf}^{(i)(k)}) =  \sum\limits_{k=1}^K(\textbf{r}_{pf}^{(i)(k)} - {\bf C}_{(i)}^{(k)} \bm{\alpha}^{(i)})^{2} + {\lambda}^{i}  \parallel  \bm{\alpha}^{(i)} \parallel _2^2. \\
     & \to \bm{\alpha}^{(i)} = \left( ({\bf C}_{(i)}^{(k)})^{\top} {\bf C}_{(i)}^{(k)} + {\lambda}^{i} \textbf{I} \right)^{-1} ({\bf C}_{(i)}^{(k)})^{\top} \textbf{r}_{pf}^{(i)(k)}.
\end{aligned}
\end{equation}

The exploration term $\beta^{(i)}$ is:
\begin{equation}%\small
\begin{aligned}
\beta^{(i)} & = \
&\gamma \times \sqrt{{\bf c}_{i}\left(({\bf C}_{(i)}^{(k)})^{\top} {\bf C}_{(i)}^{(k)} + {\lambda}^{i} \textbf{I}\right)^{-1} ({\bf c}_{(i)})^{\top} },
\end{aligned}
\end{equation}
where $\gamma$ is a fixed constant.

\paragraph{Implementation.}
We implement the aforementioned Multi-Armed Bandit (MAB) strategies using two distinct knowledge extraction approaches and three prompt templates. The MAB framework is flexible, enabling integration with additional knowledge extraction methods and prompt templates to optimize performance. The two knowledge extraction strategies we employ are:

\begin{itemize}
\item $\mathcal{P}_\text{RL}$: This strategy utilizes reinforcement learning (RL)-based extraction, as detailed in the previous subsection.
\item $\mathcal{P}_\text{sub}$: This heuristic sub-graph extraction method focuses on gathering a 2-hop subgraph around both the source and target entities.  Given the inherent instability of RL approaches, we introduce $\mathcal{P}_\text{sub}$ as a backup strategy in the MAB selection process, ensuring robustness if the RL-based method underperforms.
\end{itemize}

For the prompt templates, we incorporate the following:

\begin{itemize}
    \item \textbf{Triples}, represented as $\mathcal{F}_{t}$, correspond to the extracted knowledge in its original triple form, such as (\textit{Sergey\_Brin}, \textit{founder\_of}, \textit{Google}), (\textit{Sundar\_Pichai}, \textit{ceo\_of}, \textit{Google}), (\textit{Google}, \textit{is\_a}, \textit{High-tech Company}). This representation has been empirically shown to be comprehensible by black-box LLMs.
    \item \textbf{Sentences} serve as a transformation method to reframe the knowledge into natural language, $\mathcal{F}_{s}$, such as: “\textit{Sergey Brin, the founder of Google, a high-tech company, has handed over the reins to Sundar Pichai, the current CEO of the company.}”
    \item \textbf{Graph Description}, $\mathcal{F}_{g}$ prompts the LLM by treating the knowledge as a structured graph. This preprocessing is done via the LLM itself to generate descriptions that emphasize key entities, such as: “\textit{Google, a leading high-tech company, stands central in the network. The entity has strong ties to significant figures in the tech industry. Sergey Brin, one of the founders, created Google, marking its historical foundation. Today, Sundar Pichai is the CEO, symbolizing its present leadership. Consequently, Google serves as a pivotal link among these key figures.}”
\end{itemize}

Given the two knowledge extraction methods: $\mathcal{P}_\text{sub}$ and $\mathcal{P}_\text{RL}$, along with the three prompt translation methods: $\mathcal{F}_{t}$, $\mathcal{F}_{s}$, and $\mathcal{F}_{g}$, the MAB is trained to prioritize the most suitable combination of extraction method and prompt format based on feedback from the LLMs. The goal is to select the best pairing for different real-world question scenarios, i.e., 
$\mathcal{P}\mathcal{F} = \{(\mathcal{P}_{sub}\mathcal{F}_{t}),
(\mathcal{P}_{sub}\mathcal{F}_{s}),
(\mathcal{P}_{sub}\mathcal{F}_{g}),
(\mathcal{P}_{RL}\mathcal{F}_{t}),
(\mathcal{P}_{RL}\mathcal{F}_{s}),
(\mathcal{P}_{RL}\mathcal{F}_{g})\}$.

\section{Experiments}
In this section, we perform comprehensive experiments to assess the effectiveness of \texttt{KnowGPT} on three widely used question-answering datasets, spanning both commonsense reasoning and domain-specific tasks. The implementation of \texttt{KnowGPT} is based on GPT-3.5. Our experiments are designed to address the following research questions:

\begin{itemize}
    \item \textbf{RQ1 (Main results)}: How does \texttt{KnowGPT} compare to the current state-of-the-art large language models (LLMs) and knowledge graph-enhanced question answering (QA) baselines? 
    \item \textbf{RQ2 (Ablation Study)}: What contribution does each key component of \texttt{KnowGPT} make to the overall performance?
    \item \textbf{RQ3 (Case study)}: How does knowledge graph integration enhance the handling of complex reasoning tasks? 
\end{itemize}

\begin{table}[hpt]
	\centering
 % \vspace{-2mm}
	\caption{Performance comparison among baseline models and KnowGPT.}
	% \vspace{-3mm}
	\label{tab:main_results}
     \resizebox{1\textwidth}{!}{
	\begin{tabular}{clcccccc} 
	\toprule
 \multirow{2}{*}{\textbf{Catagory}} & \multirow{2}{*}{\textbf{Model}}  
	
    &\multicolumn{2}{c}{\textbf{CommonsenseQA}}
	& \multicolumn{2}{c}{\textbf{OpenBookQA}} 
	&\multicolumn{2}{c}{\textbf{MedQA}}\\
	\cmidrule(lr){3-4} \cmidrule(lr){5-6} \cmidrule(lr){7-8}
	& &IHdev-Acc. &IHtest-Acc. &Dev-Acc. &Test-Acc. &Dev-Acc. &Test-Acc.\\ \midrule[0.6pt]
        \multirow{3}{*}{LM + Fine-tuning}
        &Bert-base &0.573 &0.535 &0.588 &0.566 &0.359 &0.344 \\
        &Bert-large &0.611 &0.554 &0.626 &0.602 &0.373 &0.367 \\
        &RoBerta-large &0.731 &0.687 &0.668 &0.648 &0.369 &0.361 \\ \midrule
        % \cmidrule(lr){1-1} \cmidrule(lr){2-8}
        \multirow{6}{*}{KG-enhanced LM}
        &MHGRN &0.745 &0.713 &0.786 &0.806 &- &- \\
        &QA-GNN &0.765 &0.733 &0.836 &0.828 &0.394 &0.381 \\
        &HamQA &0.769 &0.739 &0.858 &0.846 &0.396 &0.385 \\
        &JointLK &0.777 &0.744 &0.864 &0.856 &0.411 &0.403 \\
        &GreaseLM &0.785 &0.742 &0.857 &0.848 &0.400 &0.385 \\ 
        &GrapeQA &0.782 &0.749 &0.849 &0.824 &0.401 &0.395 \\ 
        \midrule
       
        % \cmidrule(lr){2-8}
        
        % Black-Box LLMs
        \multirow{9}{*}{LLM + Zero-shot}
        &ChatGLM &0.473 &0.469&0.352 &0.360 &0.346 &0.366\\
        % &LLaMA &  &  & & & &\\
        % ~\cite{du2022glm}
        &ChatGLM2 &0.440 &0.425 &0.392 &0.386 &0.432 &0.422 \\
        &Baichuan-7B & 0.491 & 0.476 &0.411 &0.395 &0.334 &0.319\\
        &InternLM &0.477  &0.454  &0.376 &0.406 &0.325 & 0.348\\
        &Llama2 (7b) &0.564 &0.546 &0.524 & 0.467&0.338 &0.340 \\
        &Llama3 (8b) &0.745 &0.723 &0.771 &0.730 &0.639 &0.697 \\
        % \\\midrule
        % 0.406
        &GPT-3 &0.539 &0.520 &0.420 &0.482 &0.312 &0.289 \\
        &GPT-3.5 &0.735 &0.710 &0.598 &0.600 &0.484 &0.487 \\
        &GPT-4 &0.776 &0.786 &0.878 &0.910 &0.739 &0.763 \\ 
        % \cmidrule(lr){2-8}
        \midrule
        \multirow{3}{*}{LLM + KG Prompting }
        &CoK &0.759 &0.739 &0.835 &0.869 &0.706 &0.722 \\
         &RoG &0.750&0.734 &0.823 &0.861 &0.713 &0.726 \\
        & Mindmap & 0.789& 0.784 &0.851 &0.882 &0.747 &0.751\\
        \midrule
        Ours
        &\texttt{KnowGPT} &\textbf{0.827} &\textbf{0.818} &\textbf{0.900} &\textbf{0.924} &\textbf{0.776} &\textbf{0.781} \\\cmidrule(lr){1-8}
        \texttt{KnowGPT} vs. GPT-3.5
        &+ 23.7\% (Avg.) &+ 9.2\% &+ 10.8\% &+ 31.2\% &+ 32.4\% &+ 29.2\% &+ 29.4\% \\
        \texttt{KnowGPT} vs. GPT-4
        &+2.9\% (Avg.) & + 5.1\%&+ 3.3\% &+ 2.2\% &+ 1.4\% &+ 3.7\% &+ 1.8\% \\   
    \bottomrule
	\end{tabular}	}
 \vspace{1mm}\\
 \footnotesize{*We used `text-davinci-002' and `gpt-3.5-turbo' provided by OpenAI as the implementation of GPT models.\\}
 % \vspace{-5mm}
 % We used `text-davinci-002' provided by OpenAI as the implementation of GPT-3, and `gpt-3.5-turbo' for GPT-3.5.
\end{table}

\subsection{Experimental Setup}
{\bf Datasets}.
We evaluate \texttt{KnowGPT} on three question-answering datasets from two distinct fields: CommonsenseQA~\cite{CommonsenseQA} and OpenBookQA~\cite{mihaylov2018can} serve as benchmarks for commonsense reasoning tasks, while MedQA-USMLE~\cite{app11146421} is used for domain-specific question answering. For detailed statistics of these datasets, please refer to Table~\ref{tab:DatasetStatistics} in the Appendix.

\textbf{Background Knowledge Graph}
To facilitate common-sense reasoning, we employ \texttt{ConceptNet}~\cite{speer2017conceptnet}, an extensive commonsense knowledge graph comprising more than 8 million interconnected entities through 34 concise relationships. For tasks specific to the medical domain, we leverage \texttt{USMLE}~\cite{QA-GNN} as our foundational knowledge source. USMLE is a biomedical knowledge graph that amalgamates the Disease Database segment of the Uniﬁed Medical Language System (UMLS)~\cite{gkh061} and DrugBank~\cite{gkx1037}. This repository encompasses 9,958 nodes and 44,561 edges.

{\bf Baselines}.
We select baseline models from four different categories for a thorough comparison.
\noindent {\textit{LM + Fine-tuning}}.
We compare \texttt{KnowGPT} with standard fine-tuned language models (LMs). In particular, we use Bert-base, Bert-large~\cite{devlin2018bert}, and RoBerta-large~\cite{liu2019roberta} as representative fine-tuned LMs. For both commonsense and biomedical QA tasks, we fine-tune these models by adding linear layers.

\noindent \textit{KG-enhanced LM}.
We also compare with several recent models that integrate knowledge graphs into QA tasks, including MHGRN~\cite{MHGRN}, QA-GNN~\cite{QA-GNN}, HamQA~\cite{dong2023hierarchy}, JointLK~\cite{sun2021jointlk}, GreaseLM~\cite{zhang2022greaselm}, and GrapeQA~\cite{taunk2023grapeqa}. 
For a fair comparison, we implement these methods using advanced language models optimized for the respective datasets. RoBerta-large~\cite{liu2019roberta} is used for CommonsenseQA, while AristoRoBERTa~\cite{clark2020f} is employed for OpenBookQA, and for MedQA, we use the top biomedical model, SapBERT~\cite{liu2020self}. These methods, being white-box models, are computationally expensive and therefore cannot be used with modern LLMs like GPT-3.5 or GPT-4.

\noindent \textit{LLM + Zero-shot}.
We include several well-known generative LLMs, including ChatGLM, ChatGLM2, Baichuan-7B, InternLM, GPT-3, GPT-3.5, and GPT-4, as knowledge-agnostic models. We utilize 'text-davinci-002' for GPT-3, 'gpt-3.5-turbo' for GPT-3.5, and 'gpt-4' for GPT-4 (additional details on the implementations of these models are provided in Appendix A.4). These models are evaluated in a zero-shot setting using the test set questions.

\noindent \textit{LLM + KG Prompting}. To test the efficacy of our prompting strategy, we also include state-of-the-art knowledge graph prompting methods, such as CoK~\cite{wang2023boosting}, RoG~\cite{luo2023reasoning}, and Mindmap~\cite{wen2023mindmap}. Notably, KGR~\cite{guan2024mitigating} is excluded from the comparison as the authors have not released their code.

\subsection{Main Results (RQ1)}
To address \textbf{RQ1}, we compare the performance of \texttt{KnowGPT} against state-of-the-art models across three benchmark datasets. We use accuracy as our performance metric, which measures the percentage of questions that the model answers correctly from the total test set questions. The following observations are made:

\begin{itemize}
    \item \texttt{KnowGPT} outperforms all categories of baseline models, including 16 different methods, across all datasets and architectures. This indicates that \texttt{KnowGPT} is highly effective in leveraging knowledge from KGs to enhance LLMs.
    \item \texttt{KnowGPT} outperforms GPT-3.5 and even GPT-4 in terms of accuracy. On average, \texttt{KnowGPT} achieves 23.7\% higher accuracy than GPT-3.5. Specifically, it outperforms GPT-3.5 by 10.8\%, 32.4\%, and 29.4\% on the CommonsenseQA, OpenBookQA, and MedQA datasets, respectively. Even more impressively, despite being based on GPT-3.5, \texttt{KnowGPT} surpasses GPT-4 by 3.3\%, 1.4\%, and 1.8\% on these datasets, respectively. These results emphasize that integrating black-box knowledge can significantly enhance LLM capabilities.
    \item \texttt{KnowGPT} also outperforms all KG-enhanced LMs, demonstrating the effectiveness of our black-box knowledge injection method. This approach offers the advantage of being adaptable to GPT-3.5 through the model API, which is not possible with white-box methods.
\end{itemize}

\paragraph{Leaderboard Ranking.}
We submitted our results to the official OpenBookQA leaderboard, maintained by the dataset authors. The full leaderboard records are available on the official site\footnote{\url{https://leaderboard.allenai.org/open_book_qa/submissions/public}}, and our submission can be accessed at\footnote{\url{https://leaderboard.allenai.org/open_book_qa/submission/cp743buq4uo7qe4e9750}}.

In Table~\ref{tab:csqa_leaderboard}, we summarize related submissions, which fall into three categories: traditional KG-enhanced LMs, fine-tuning of LLMs (e.g., T5-11B used in UnifiedQA), and ensemble methods. \texttt{KnowGPT} outperforms traditional KG-enhanced LMs by a significant 5.2\% improvement over the best baseline. Additionally, although ensemble methods dominate the leaderboard, \texttt{KnowGPT} achieves competitive performance without ensembling, outperforming GenMC Ensemble~\cite{huang2022clues} by 0.6\%. Remarkably, \texttt{KnowGPT} also approaches human-level performance.

\begin{table*}[t]

    \caption{OpenBookQA Leaderboard records of three groups of related models. 
    % \daochen{Why ranked the second but not the first? Which single model is better than us in this table?}).
    % \daochen{How about just removing the ensemble methods in the table? And just say we exclude them in the caption.}
    % Sure, we add the explanation in the caption.
    }
    % \vspace{-3mm}
    \centering
    \resizebox{0.6\textwidth}{!}{
    \begin{tabular}{lcclc}
    \toprule
    \multicolumn{5}{c}{\textbf{Human Performance}  (\textbf{0.917})}\\
    \midrule
    w/o KG & 0.778 &&UnifiedQA~\cite{unifiedqa} & 0.872\\
    MHGRN~\cite{MHGRN} & 0.806 &&DRAGON~\cite{yasunaga2022deep}& 0.878\\  
    QA-GNN~\cite{QA-GNN} & 0.828 && GenMC~\cite{huang2022clues} & 0.898 \\ \cmidrule(lr){4-5}
    GreaseLM~\cite{zhang2022greaselm} & 0.848& &Human Performance  &  0.917
   \\
    HamQA~\cite{dong2023hierarchy}  & 0.850 &&GenMC Ensemble~\cite{huang2022clues}& 0.920\\
    JointLK~\cite{sun2021jointlk} & 0.856 && X-Reasoner~\cite{huang2023mvp}& 0.952 \\\cmidrule(lr){4-5}
    GSC~\cite{wang2021gnn} & 0.874 && \texttt{KnowGPT} & \textbf{0.926}\\
    \bottomrule
    \end{tabular}}
    \label{tab:csqa_leaderboard}
    % \vspace{-5mm}
\end{table*}

\subsection{Ablation Studies (RQ2)}
To answer \textbf{RQ2}, we conduct two ablation studies. \textbf{First}, in Table~\ref{tab:ablation1}, we evaluate the impact of the tailored reinforcement learning-based knowledge extraction module, $\mathcal{P}_\text{RL}$, by comparing it with the heuristic sub-graph extraction strategy, $\mathcal{P}_\text{sub}$. We assess the performance by feeding the extracted knowledge into GPT-3.5 using the 'Sentence' prompt format, $\mathcal{F}_{s}$. The 'w/o KG' baseline is also included, where GPT-3.5 answers the questions without any external reasoning context. The results clearly highlight the significance of our proposed knowledge extraction strategies. \textbf{Second}, we examine the effect of the three prompt formats using the same extracted knowledge. The results, shown in Table~\ref{tab:ablation2}, reveal that although the performance differences between formats are small (2.2\% - 3.3\%), each format is better suited to certain types of questions. We further explain these findings in the case study section. These ablation studies underscore the importance of each module in \texttt{KnowGPT}, demonstrating that combining deep RL-based knowledge extraction with context-aware prompt translation leads to the best performance across all datasets. The MAB-based prompt selection module used in KnowGPT dynamically selects the most effective prompt templates and knowledge combinations for a given query. This adapts to incompleteness and noises in KGs since the MAB balances exploration (trying new strategies) and exploitation (using known effective ones), optimizing for robustness even with imperfect KG data.

\begin{table}[ht]
\caption{Ablation study on the effectiveness of two knowledge extraction methods. 
 %The overall performance is evaluated by directly prompting the extracted knowledge to parameter-agnostic LLMs, including GPT-3, GPT-3.5 and GPT-4. 
 % \daochen{Can we not use notations in the table? Use text to describe[done]}
 }
	% \vspace{-3mm}
	\label{tab:ablation1}
\resizebox{1.0\columnwidth}{!}{
\begin{tabular}{cccccc} 
	\toprule
	\multirow{2}{*}{\textbf{Knowledge Extraction}} & \multirow{2}{*}{\textbf{Model}} 
    &\multicolumn{2}{c}{\textbf{CSQA}}
	& \textbf{OBQA}
	&\textbf{MedQA}\\
	\cmidrule(lr){3-4} \cmidrule(lr){5-5} \cmidrule(lr){6-6}
	& &IHdev &IHtest  &Test  &Test\\ \midrule[0.6pt]
         \multirow{3}{*}{w/o KG}
        &GPT-3 &0.539 &0.520 &0.482 &0.289 \\
        &GPT-3.5 &0.735 &0.710 &0.598 &0.487 \\
        &GPT-4 &0.776 &0.786 &0.910 &0.763 \\ \cmidrule(lr){1-6}
        \multirow{1}{*}{$\mathcal{P}_\text{sub}$}
        &GPT-3.5 &0.750 &0.739 &0.865 &0.695 \\ \cmidrule(lr){1-6}
        \makecell[c]{$\mathcal{P}_\text{RL}$}
        &GPT-3.5 &0.815 &0.800 &0.889 &0.755 \\ \cmidrule(lr){1-6}
        Ours
        &\texttt{KnowGPT} &\textbf{0.827} &\textbf{0.818} &\textbf{0.924} &\textbf{0.781} \\
    \bottomrule
	\end{tabular}	
    }
% \vspace{-5mm}
\end{table}

\begin{table}[ht]

\caption{Ablation study on the effectiveness of two knowledge extraction methods. 
 %The overall performance is evaluated by directly prompting the extracted knowledge to parameter-agnostic LLMs, including GPT-3, GPT-3.5 and GPT-4. 
 % \daochen{Can we not use notations in the table? Use text to describe[done]}
 }
	% \vspace{-3mm}
\centering
	% \vspace{-3mm}
 \label{tab:ablation2}
	
\resizebox{1.0\columnwidth}{!}{
	\begin{tabular}{cccccc} 
	\toprule
	\multirow{2}{*}{\textbf{Knowledge Extraction}} & \multirow{2}{*}{\textbf{Prompts}}  
    &\multicolumn{2}{c}{\textbf{CSQA}}
	& \textbf{OBQA}
	&\textbf{MedQA}\\
	\cmidrule(lr){3-4} \cmidrule(lr){5-5} \cmidrule(lr){6-6}
	& &IHdev &IHtest  &Test  &Test\\ \midrule[0.6pt]
         \multirow{3}{*}{$\mathcal{P}_\text{sub}$}
        &$\mathcal{F}_{t}$ &0.728 &0.701 &0.832 &0.589 \\
        &$\mathcal{F}_{s}$ &0.750 &0.739 &0.865 &0.695 \\
        &$\mathcal{F}_{g}$ &0.737 &0.715 &0.871 &0.680 \\ \midrule[0.6pt]
        \multirow{3}{*}{$\mathcal{P}_\text{RL}$}
        &$\mathcal{F}_{t}$ &0.782 &0.769&0.853 &0.739 \\
        &$\mathcal{F}_{s}$ &0.815 &0.800 &0.889 &0.755 \\
        &$\mathcal{F}_{g}$ &0.806 &0.793 &0.906 &0.762 \\ 
        \midrule[0.6pt]
        \multicolumn{2}{c}{Full \texttt{KnowGPT}}
        &\textbf{0.827} &\textbf{0.818} &\textbf{0.924} &\textbf{0.781} \\
    \bottomrule
	\end{tabular}	}
	
% \vspace{-5mm}
\end{table}

\begin{figure}[thp]
    \centering
    \includegraphics[width=13.6cm]{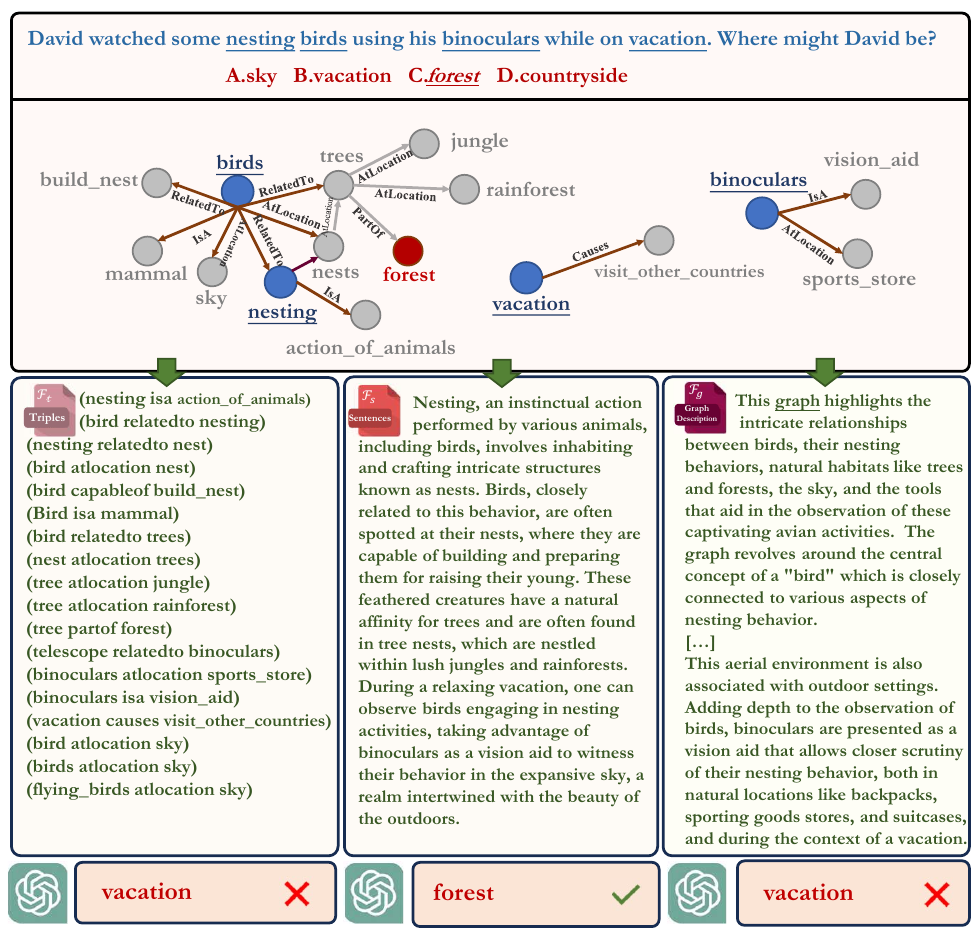}
    \vspace{-2mm}
    \caption{A case study on exploring the effectiveness of different prompt formats for particular questions.
    % ., based on the same extracted knowledge. 
    The extracted knowledge is shown in the middle of this figure in the form of a graph, where the nodes in blue are the key topic entities and the red is the target answer. The text boxes at the bottom are the final prompts generated based on three different formats.  }
    \label{fig:case1}
    \vspace{-6mm}
\end{figure}

\subsection{Case Studies (RQ3)}\label{sec:rq3}
To address \textbf{RQ3}, we provide a detailed case study from CommonsenseQA. In Figure~\ref{fig:case1}, we visualize the extracted knowledge and the corresponding textual input for GPT-3.5. In this case, GPT-3.5 correctly answers the question when the sentence-based prompt is used, but struggles when the question is framed using the triple or graph description formats. This comparison highlights the superior ability of \texttt{KnowGPT} in automatically generating contextually appropriate prompts. The following observations are made: \((i)\) The triple format $\mathcal{F}_{t}$ is most suitable for simple questions that rely on one-hop knowledge. \((ii)\) The graph description format may introduce noise by overemphasizing certain terms, leading to misleading predictions. \((iii)\) \texttt{KnowGPT} excels in constructing suitable prompts based on the specific nature of each question.

\section{Summary}
This paper addresses the challenge of hallucination in large language models (LLMs), particularly in specialized domains. While LLMs have impressive reasoning capabilities, they often struggle with professional questions in specialized fields due to insufficient relevant knowledge in their pre-training datasets. To resolve this issue, we propose \texttt{KnowGPT}, a model that augments LLMs with domain-specific knowledge from knowledge graphs (KGs) to improve the accuracy of answering professional queries. \texttt{KnowGPT} integrates KGs into LLMs using model APIs, requiring no modifications to the underlying LLM architecture. Our approach utilizes a deep reinforcement learning policy to extract concise reasoning background from KGs and employs a Multi-Armed Bandit (MAB) strategy to select the most effective knowledge extraction method and prompt template for each query. Extensive experiments across both general and domain-specific QA tasks demonstrate that \texttt{KnowGPT} outperforms all baseline models, providing strong evidence for the effectiveness of KG-enhanced LLMs.

\chapter{Conclusion and Future Work}
\section{Conclusion}
This thesis has presented a systematic framework for enhancing LLMs with reliable KGs, addressing critical challenges in KG refinement, completion, and dynamic alignment with LLMs. The four interconnected papers form a lifecycle solution that ensures LLMs are grounded in accurate and reliable knowledge:
\begin{itemize}
    \item Contrastive Knowledge Graph Error Detection (Paper 1): By leveraging contrastive learning to identify semantic contradictions in KGs, this work establishes a robust foundation for KG reliability.
    \item Attribute-Aware KG Embedding (Paper 2): Recognizing that structural patterns alone cannot resolve ambiguities in entities with shared names, this work integrates entity attributes (e.g., geospatial coordinates, temporal metadata) into error detection.
    \item Inductive Knowledge Graph Completion (Paper 3): Addressing the incompleteness of even refined KGs, this work introduces a graph neural network (GNN) model that combines structural patterns with logical reasoning to infer missing relationships. The model achieves state-of-the-art performance on inductive benchmarks, enabling applications to dynamic and emerging domains.
    \item Inductive Knowledge Graph Completion (Paper 3): Addressing the incompleteness of even refined KGs, this work introduces a graph neural network (GNN) model that combines structural patterns with logical reasoning to infer missing relationships. The model achieves state-of-the-art performance on inductive benchmarks, enabling applications to dynamic and emerging domains.    
\end{itemize}

Collectively, these contributions advance the field of AI reliability by leveraging structural information from reliable KGs, enabling LLMs to generate responses that are not only fluent but also factually grounded.
\section{Future Work}

While this thesis makes significant strides in enhancing LLMs with reliable KGs, several challenges remain unaddressed, offering fertile ground for future research:

\begin{itemize}
    \item Scalability: Real-time retrieval from billion-scale KGs, such as Wikidata or enterprise knowledge bases, remains computationally intensive. Future work could explore distributed graph processing techniques or approximate nearest neighbor search algorithms to improve retrieval efficiency without sacrificing accuracy.
    \item Multimodal Knowledge Integration: Most existing systems focus on textual KGs, ignoring non-textual data such as images, time series, or geospatial information. Integrating multimodal KGs with LLMs could enable applications in domains like medical imaging (e.g., combining MRI scans with patient records) or urban planning (e.g., linking geospatial data with textual reports).

    \item Temporal Knowledge Integration: Real-world knowledge is dynamic, with facts evolving over time. Extending the framework to handle temporal KGs, where relationships are timestamped, could enable LLMs to generate contextually accurate outputs based on the most up-to-date information.  

    \item Domain-Specific Customization: While the proposed framework is general-purpose, tailoring it to more specific domains (e.g., law, medicine, or education) could yield significant performance gains. For instance, legal KGs could incorporate case law hierarchies, while medical KGs could integrate ontologies like SNOMED-CT~\cite{donnelly2006snomed}. Future research could investigate domain-specific adaptations of the error detection, completion, and alignment modules.
    
    \end{itemize}

\bibliographystyle{plain}
\bibliography{reference}

\end{document}